\documentclass[prodmode,acmtecs]{acmsmall} 

\usepackage{amsmath}
\usepackage{graphicx}
\usepackage{stmaryrd}
\usepackage{amsfonts}

\usepackage[ruled]{algorithm2e}

\SetAlFnt{\small}
\SetAlCapFnt{\small}
\SetAlCapNameFnt{\small}
\SetAlCapHSkip{0pt}
\IncMargin{-\parindent}

\acmVolume{0}
\acmNumber{0}
\acmArticle{0}
\acmYear{0000}
\acmMonth{0}

\begin{document}

\markboth{D.G. Schwartz}{Dynamic Reasoning Systems}

\title{Dynamic Reasoning Systems}
\author{DANIEL G. SCHWARTZ
\affil{Florida State University}}

\begin{abstract}
A {\it dynamic reasoning system} (DRS) is an adaptation of a conventional formal logical system that explicitly portrays reasoning as a temporal activity, with each extralogical input to the system and each inference rule application being viewed as occurring at a distinct time step.   Every DRS incorporates some well-defined logic together with a controller that serves to guide the reasoning process in response to user inputs.  Logics are generic, whereas controllers are application-specific.  Every controller does, nonetheless, provide an algorithm for nonmonotonic belief revision.  The general notion of a DRS comprises a framework within which one can formulate the logic and algorithms for a given application and prove that the algorithms are correct, i.e., that they serve to (i) derive all salient information and (ii) preserve the consistency of the belief set.  This paper illustrates the idea with ordinary first-order predicate calculus, suitably modified for the present purpose, and two examples.  The latter example revisits some classic nonmonotonic reasoning puzzles (Opus the Penguin, Nixon Diamond) and shows how these can be resolved in the context of a DRS, using an expanded version of first-order logic that incorporates typed predicate symbols.   All concepts are rigorously defined and effectively computable, thereby providing the foundation for a future software implementation.
\end{abstract}




\acmformat{Daniel G. Schwartz. 2014. Dynamic reasoning systems.}

\begin{bottomstuff}
Author's address: Daniel G. Schwartz, Department of Computer Science, Mail Code 4530, Florida State University, Tallahassee, FL 32306. 
\end{bottomstuff}

\maketitle

\section{Introduction}

The notion of a {\it dynamic reasoning system} (DRS) was introduced in \cite{schwartz97} for purposes of formulating reasoning involving a logic of `qualified syllogisms'.  The idea arose in an effort to devise some rules for evidence combination.  The logic under study included a multivalent semantics where propositions $P$ were assigned a probabilistic `likelihood value' $l(P)$ in the interval $[0,1]$, so that the likelihood value plays the role of a surrogate truth value.  The situation being modeled is where, based on some evidence, $P$ is assigned a likelihood value $l_1$, and then later, based on other evidence, is assigned a value $l_2$, and it subsequently is desired to combine these values based on some rule into a resulting value $l_3$.  This type of reasoning cannot be represented in a conventional formal logical system with the usual Tarski semantics, since such systems do not allow that a proposition may have more than one truth value; otherwise the semantics would not be mathematically well-defined.  Thus the idea arose to speak more explicitly about different occurrences of the propositions $P$ where the occurrences are separated in time.  In this manner one can construct a well-defined semantics by mapping the different time-stamped occurrences of $P$ to different likelihood/truth values.

In turn, this led to viewing a `derivation path' as it evolves over time as representing the knowledge base, or belief set, of a reasoning agent that is progressively building and modifying its knowledge/beliefs through ongoing interaction with its environment (including inputs from human users or other agents).  It also presented a framework within which one can formulate a Doyle-like procedure for nonmonotonic `reason maintenance' \cite{doyle79,smith88}.  Briefly, if the knowledge base harbors inconsistencies due to contradictory inputs from the environment, then in time a contradiction may appear in the reasoning path (knowledge base, belief set), triggering a back-tracking procedure aimed at uncovering the `culprit' propositions that gave rise to the contradiction and disabling (disbelieving) one or more of them so as to remove the inconsistency. 

Reasoning is nonmonotonic when the discovery and introduction of new information causes one to retract previously held assumptions or conclusions.  This is to be contrasted with classical formal logical systems, which are monotonic in that the introduction of new information (nonlogical axioms) always increases the collection of conclusions (theorems).  \cite{schwartz97} contains an extensive bibliography and survey of the works related to nonmonotonic reasoning as of 1997.  In particular, this includes a discussion of (i) the classic paper by McCarthy and Hayes \cite{mccarthy69} defining the `frame problem' and describing the `situation calculus', (ii) Doyle' s `truth maintenance system' \cite{doyle79} and  subsequent `reason maintenance system' \cite{smith88}, (iii) McCarthy's `circumscription' \cite{mccarthy80}, (iv) Reiter's `default logic' \cite{reiter80}, and (v) McDermott and Doyle's `nonmonotonic logic' \cite{mcdermott80}.  With regard to temporal aspects, there also are discussed works by Shoham and Perlis.  \cite{shoham86,shoham88} explores the idea of making time an explicit feature of the logical formalism for reasoning `about' change, and \cite{shoham93} describes a vision of `agent-oriented programming' that is along the same lines of the present DRS, portraying reasoning itself as a temporal activity.  In \cite{elgot-drapkin88,elgot-drapkin87,elgot-drapkin91,elgot-drapkin90,miller93,perlis90} Perlis and his students introduce and study the notion of `step logic', which studies reasoning as `situated' in time, and in this respect also has elements in common with the notion of a DRS.  Additionally mentioned but not elaborated upon in \cite{schwartz97} is the so-called AGM framework \cite{agm85,gardenfors88,gardenfors92}, named after its originators.  Nonmonotonic reasoning and belief revision are related in that the former may be viewed as a variety of the latter.

These cited works are nowadays regarded as the classic approaches to nonmonotonic reasoning and belief revision.  Since 1997 the AGM approach has risen in prominence, due in large part to the publication \cite{hansson99}, which builds upon and substantially advances the AGM framework.  AGM defines a belief set as a collection of propositions that is closed with respect to the classical consequence operator, and operations of `contraction', `expansion' and `revision' are defined on belief sets.   \cite{hansson99} made the important observation that a belief set can conveniently be represented as the consequential closure of a finite `belief base', and these same AGM operations can be defined in terms of operations performed on belief bases.  Since that publication, AGM has enjoyed a steadily growing population of adherents.  A recent publication \cite{ferme11} overviews the first 25 years of research in this area.

Another research thread that has risen to prominence is the logic-programming approach to nonmonotonic reasoning known as Answer Set Prolog (AnsProlog).  A major work on AnsProlog is the treatise \cite{baral03}.   This line of research suggests that an effective approach to nonmonotonic reasoning can be formulated in an extension of the well-known Prolog programming language.  Interest in this topic has spawned a series of eleven conferences on Logic Programming and Nonmonotonic Reasoning, the most recent of which is \cite{delgrande11}. 

The DRS framework discussed in the present work has elements in common with both AGM and AnsProlog, but also differs from these in several respects.  Most importantly, the present focus is on the creation of computational algorithms that are sufficiently articulated that they can effectively be implemented in software and thereby lead to concrete applications.  This element is still lacking in AGM, despite Hansson's contribution regarding finite belief bases.  The AGM operations continue to be given only as set-theoretic abstractions and have not yet been translated into computable algorithms.  Regarding AnsProlog, this research thread holds promise of a new extension of Prolog, but siimliarly with AGM the necessary algorithms have yet to be formulated.

A way in which the present approach varies from the original AGM approach, but happens to agree with the views expressed by \cite{hansson99} (cf. pp. 15-16), in that it dispenses with two of the original 'rationality postulates', namely, the requirements that the underlying belief set be at all times (i) consistent, and (ii) closed with respect to logical entailment.  The latter is sometimes called the `omniscience' postulate, inasmuch as the modeled agent is thus characterized as knowing all possible logical consequences of its beliefs. 

These postulates are intuitively appealing, but they have the drawback that they lead to infinitary systems and thus cannot be directly implemented on a finite computer.  To wit, the logical consequences of even a fairly simple set of beliefs will be infinite in number; and assurance of consistency effectively requires omniscience since one must know whether the logical consequences of the given beliefs include any contradictions.  Dropping these postulates does have anthropomorphic rationale, however, since humans themselves cannot be omniscient in the sense described, and, because of this, often harbor inconsistent beliefs without being aware of it.  Thus it is not unreasonable that our agent-oriented reasoning models should have these same characteristics.  Similar remarks may be found in the cited pages of \cite{hansson99}. 

The present work differs from the AGM approach in several other respects.  First, what is here taken as a `belief set' is neither a belief set in the sense of AGM and Hansson nor a Hansson-style belief base.  Rather it consists of the set of statements that have been input by an external agent as of some time $t$, together with the consequences of those statements that have been derived in accordance with the algorithms provided in a given 'controller'.  Second, by labeling the statements with the time step when they are entered into the belief set (either by an external agent or derived by means of an inference rule), one can use the labels as a basis for defining the associated algorithms. Third, whereas G\"ardenfors,  Hansson, and virtually all others that have worked with the AGM framework, have confined their language to be only propositional, the present work takes the next step to full first-order predicate logic.  This is significant inasmuch as the consistency of a finite set of propositions with respect to the classical consequence operation can be determined by truth-table methods, whereas the consistency of a finite set of statements in first-order predicate logic is undecidable (the famous result due to G\"odel).  For this reason the present work develops a well-defined semantics for the chosen logic and establishes a soundness theorem, which in turn can be used to establish consistency.  Last, the present use of a controller is itself new, and leads to a new efficacy for applications.   

The notion a controller was not present in the previous work \cite{schwartz97}.  Its introduction here thus fills an important gap in that treatment.  The original conception of a DRS provided a framework for modeling the reasoning processes of an artificial agent to the extent that those processes follow a well-defined logic, but it offered no mechanism for deciding what inference rules to apply at any given time.  What was missing was a means to provide the agent with a sense of purpose, i.e., mechanisms for pursuing goals. This deficiency is remedied in the present treatment.  The controller responds to inputs from the agent's environment, expressed as propositions in the agent's language.  Inputs are classified as being of various `types', and, depending on the input type, a reasoning algorithm is applied.  Some of these algorithms may cause new propositions to be entered into the belief set, which in turn may invoke other algorithms.  These algorithms thus embody the agent's purpose and are domain-specific, tailored to a particular application.  But in general their role is to ensure that (i) all salient propositions are derived and entered into to the belief set, and (ii) the belief set remains consistent.  The latter is achieved by invoking a Doyle-like reason maintenance algorithm whenever a contradiction, i.e., a proposition of the form $P\land\lnot P$, is entered into the belief set. 

This work accordingly represents a rethinking, refinement, and extension of the earlier work, aimed at (1) providing mathematical clarity to some relevant concepts that previously were not explicitly defined, (ii) introducing the notion of a controller and spelling out its properties, and (iii) illustrating these ideas with a small collection of example applications.  The present effort may be viewed as laying the groundwork for a future project to produce a software implementation of the DRS framework, this being a domain-independent software framework into which can be plugged domain-specific modules as required for any given application.  Note that the present mathematical work is a necessary prerequisite for the software implementation inasmuch as this provides the needed formal basis for an unambiguous set of requirements specifications. 

The following Section 2 provides a fully detailed definition of the notion of a DRS.  Section 3 presents the syntax and semantics for first-order predicate logic, suitably adapted for the present purpose, and proves a series of needed results including a Soundness Theorem.  This section also introduces some derived inference rules for use in the ensuing example applications.  Section 4 illustrates the core ideas in an application to a simple document classification system.  Section 5 extends this to an application for multiple-inheritance reasoning, a form of default reasoning underlying frame-based expert systems.  This provides new resolutions for some well-known puzzles from the literature on nonmonotonic reasoning.  Section 6 consists of concluding remarks.  

Regarding the examples, it may be noted that a Subsumption inference rule plays a central role, giving the present work elements in common also with the work on Description Logic (DL), c.f. \cite{baader03}.  The DL notion of a `role' is not employed here, however, inasmuch as the concept of an object having certain properties is modeled in Section 5 through the use of typed predicate symbols.     

An earlier, condensed, version of Sections 2, 3 and 4 has been published as \cite{schwartz10}.  The works \cite{ustymenko08-c,ustymenko10-a} contain precursors to the present notion of a DRS controller, and a DRS application using a `logic of belief and trust' has been described in \cite{ustymenko08-a,ustymenko08-b,ustymenko10-a,ustymenko10-b}.  While the present work employs classical first-order predicate calculus, the DRS framework can accommodate any logic for which the exists a well-defined syntax and semantics.

All proofs of propositions and theorems have been placed in the electronic appendix. 

\section {Dynamic Reasoning Systems}

A {\it dynamic reasoning system} (DRS) comprises a model of an artificial agent's reasoning processes to the extent that those processes adhere to the principles of some well-defined logic.  Formally it is comprised of a `path logic', which provides all the elements necessary for reasoning, and a `controller', which guides the reasoning process. 

\begin{figure}[h]
\centerline{\includegraphics[height=1.75in]{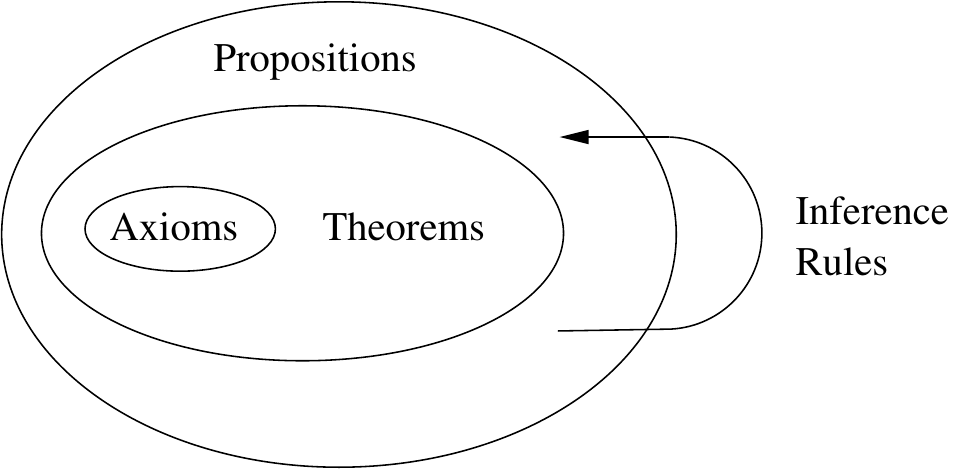}}
\medskip
\centerline{Figure 2.1. Classical formal logical system.}
\end{figure}

\begin{figure}[h]
\centerline{\includegraphics[height=2.25in]{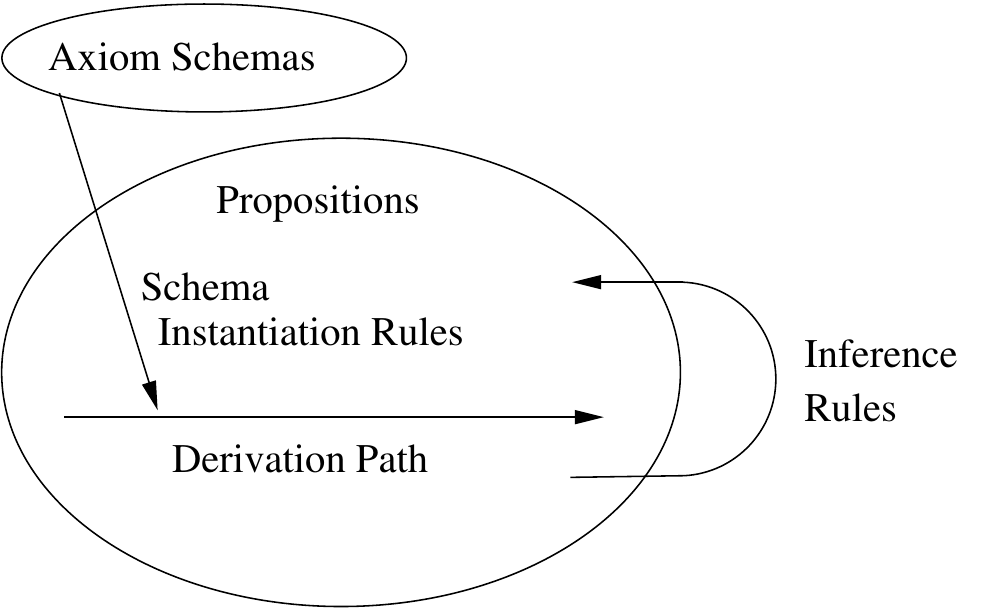}}
\medskip
\centerline{Figure 2.2. Dynamic reasoning system.}
\end{figure}  

For contrast, and by way of introductory overview, the basic structure of a classical formal logic system is portrayed in Figure 2.1 and that of a DRS in Figure 2.2.  A classical system is defined by providing a language consisting of a set of propositions, selecting certain propositions to serve as axioms, and specifying a set of inference rules saying how, from certain premises one can derive certain conclusions.  The theorems then amount to all the propositions that can be derived from the axioms by means of the rules.  Such systems are monotonic in that adding new axioms always serves to increase the set of theorems.  Axioms are of two kinds: logical and extralogical (or proper, or nonlogical).  The logical axioms together with the inference rules comprise the `logic'. The extralogical axioms comprise information about the application domain.  A DRS begins similarly with specifying a language consisting of a set of propositions.  But here the `logic' is given in terms of a set of axioms schemas, some inference rules as above, and some rules for instantiating the schemas. The indicated derivation path serves as the belief set.  Logical axioms may be entered into the derivation path by applying instantiation rules.  Extralogical axioms are entered from an external source (human user, another agent, a mechanical sensor, etc.).  Thus the derivation path evolves over time, with propositions being entered into the path either as extralogical axioms or derived by means of inference rules in accordance with the algorithms provided in the controller. Whenever a new proposition is entered into the path it is marked as `believed'.  In the event that a contradiction arises in the derivation path, a nonmonotonic belief revision process is invoked which leads to certain previously believed propositions becoming disbelieved, thereby removing the contradiction.  The full details for the two components of a DRS are given in Sections 2.1 and 2.2.

\subsection {Path Logic}

A {\it path logic} consists of a language, axiom schemas, inference rules, and a derivation path, as follows.

{\bf Language}: Here denoted $\cal L$, this consists of all {\it expressions} (or {\it formulas}) that can be generated from a given set $\sigma$ of {\it symbols} in accordance with a collection of production rules (or an inductive definition, or some similar manner of definition).  As symbols typically are of different types (e.g., individual variables, constants, predicate symbols, etc.) it is assumed that there is an unlimited supply (uncountably many if necessary) of each type.  Moreover, as is customary, some symbols will be {\it logical symbols} (e.g., logical connectives, quantifiers, and individual variables), and some will be {\it extralogical symbols} (e.g., individual constants and predicate symbols).  It is assumed that $\cal L$ contains at least the logical connectives for expressing {\it negation} and {\it conjunction}, herein denoted $\lnot$ and $\land$, or a means for defining these connectives in terms of the given connectives.  For example, in the following we take $\lnot$ and $\to$ as given and use the standard definition of $\land$ in terms of these. 

{\bf Axiom Schemas}:  Expressed in some meta notation, these describe the expressions of $\cal L$ that are to serve as {\it logical axioms}.

{\bf Inference Rules}:  These must include one or more rules that enable instantiation of the axiom schemas.  All other inference rules will be of the usual kind, i.e., stating that, from expressions having certain forms (premise expressions), one may infer an expression of some other form (a conclusion expression).  Of the latter, two kinds are allowed: {\it logical rules}, which are considered to be part of the underlying logic, and {\it extralogical rules}, which are associated with the intended application.  Note that logical axioms are expressions that are derived by applying the axiom schema instantiation rules.  Inference rules may be viewed formally as mappings from $\cal L$ into itself.

The rule set may include derived rules that simplify deductions by encapsulating frequently used argument patterns. Rules derived using only logical axioms and logical rules will also be {\it logical rules}, and derived rules whose derivations employ extralogical rules will be additional {\it extralogical rules}. 

{\bf Derivation Paths}: These consist of a sequences of pairs $(L_0,B_0),(L_1,B_1),\ldots$, where $L_t$ is the sublanguage of $\cal L$ that is in use at time $t$, and $B_t$ is the {\it belief set} in effect at time $t$.  Such a sequence is {\it generated} as follows.  Since languages are determined by the symbols they employ, it is useful to speak more directly in terms of the set $\sigma_t$ comprising the symbols that are in use at time $t$ and then let $L_t$ be the sublanguage of $\cal L$ that is based on the symbols in $\sigma_t$.  With this in mind, let $\sigma_0$ be the logical symbols of $\cal L$, so that $L_0$ is the minimal language employing only logical symbols, and let $B_0=\emptyset$. Then, given $(L_t,B_t)$, the pair $(L_{t+1},B_{t+1})$ is formed in one of the following ways: 

\begin{enumerate}

\item $\sigma_{t+1}=\sigma_t$ (so that $L_{t+1}=L_t$) and $B_{t+1}$ is obtained from $B_t$ by adding an expression that is derived by application of an inference rule that instantiates an axiom schema,

\item $\sigma_{t+1}=\sigma_t$ and $B_{t+1}$ is obtained from $B_t$ by adding an expression that is derived from expressions appearing earlier in the path by application of an inference rule of the kind that infers a conclusion from some premises,

\item $\sigma_{t+1}=\sigma_t$ and an expression employing these symbols is added to $B_t$ to form $B_{t+1}$, 

\item some new extralogical symbols are added to $\sigma_t$ to form $\sigma_{t+1}$, and an expression employing the new symbols is added to $B_t$ to form $B_{t+1}$,

\item $\sigma_{t+1}=\sigma_t$ and $B_{t+1}$ is obtained from $B_t$ by applying a belief revision algorithm as described in the following.

\end{enumerate}

Note that the use of axiom schemas together with schema instantiation rules here replaces the customary approach of defining logical axioms as all formulas having the `forms' described by the schemas and then including these axioms among the set of `theorems'.  The reason for adopting this alternative approach is to ensure that the DRS formalism is finitary, and hence, machine implementable---it is not possible to represent an infinite set of axioms (or theorems) on a computer.  That the two approaches are equivalent should be obvious.  Expressions entered into the belief set in accordance with either (3) or (4) will be {\it extralogical axioms}.   A DRS can generate any number of different derivation paths, depending on the extralogical axioms that are input and the inference rules that are applied.

Whenever an expression is entered into the belief set it is assigned a {\it label} comprised of:

\begin{enumerate}

\item A {\it time stamp}, this being the value of the subscript $t+1$ on the set $B_{t+1}$ formed by entering the expression into the belief set in accordance with any of the above items (1) through (4). The time stamp effectively serves as an {\it index} indicating the expression's position in the belief set.

\item A {\it from-list}, indicating how the expression came to be entered into the belief set.  In case the expression is entered in accordance with the above item (1), i.e., using a schema instantiation rule, this list consists of the name (or other identifier) of the schema and the name (or other identifier) of the inference rule if the system has more than one such rule.  In case the expression is entered in accordance with above item (2), the list consists of the indexes (time stamps) of the premise expressions and the name (or other identifier) of the inference rule. In case the expression is entered in accordance with either of items (3) or (4), i.e., is a extralogical axiom, the list will consist of some code indicating this (e.g., {\it es} standing for `external source') possibly together with some identifier or other information regarding the source. 

\item A {\it to-list}, being a list of indexes of all expressions that have been entered into the belief set as a result of rule applications involving the given expression as a premise.  Thus to-lists may be updated at any future time.

\item A {\it status indicator} having the value {\it bel} or {\it disbel} according as the proposition asserted by the expression currently is believed or disbelieved.  The primary significance of this status is that only expressions that are believed can serve as premises in inference rule applications.  Whenever an expression is first entered into the belief set, it is assigned status {\it bel}.  This value may then be changed during belief revision at a later time.  When an expression's status is changed from {\it bel} to {\it disbel} it is said to have been {\it retracted}.

\item An {\it epistemic entrenchment factor}, this being a numerical value indicating the strength with which the proposition asserted by the expression is held. This terminology is adopted in recognition of the work by G\"ardenfors, who initiated this concept \cite{gardenfors88,gardenfors92}, and is used here for essentially the same purpose, namely, to assist when making decisions regarding belief retractions.  Depending on the application, however, this value might alternatively be interpreted as a degree of belief, as a certainty factor, as a degree of importance, or some other type of value to be used for this purpose.\footnote{G\"ardenfors asserts that the notion of a degree of epistemic entrenchment is distinct from that of a degree (or probability) of belief. `This degree [of entrenchment] is not determined by how probable a belief is judged to be but rather by how important the belief is to inquiry and deliberation.' \cite{gardenfors88}, p. 17. Nonetheless, degrees of belief could be used as a basis for managing belief retraction if this were deemed appropriate for a given application.} In the present treatment, epistemic entrenchment values are assigned only to axioms.  No provision for propagating these factors to expressions derived via rule inferences is provided, although this would be a natural extension of the present treatment. It is agreed that logical axioms always receive the highest possible epistemic entrenchment value, whatever scale or range may be employed.

\item A {\it knowledge category specification}, having one of the values {\it a priori}, {\it a posteriori}, {\it analytic}, and {\it synthetic}.  These terms are employed in recognition of the philosophical tradition initiated by Kant \cite{kant29}.  Logical axioms are designated as a priori; extralogical axioms are designated as a posteriori; expressions whose derivations employ only logical axioms and logical inference rules are designated as analytic; and expressions whose derivations employ any extralogical axioms or extralogical rules are designated as synthetic. The latter is motivated by the intuition that an ability to apply inference rules and thereby carry out logical derivations is itself a priori knowledge, so that, even if the premises in a rule application are all a posteriori and/or the rule itself is extralogical, the rule application entails a combination of a priori and a posteriori knowledge, and the conclusion of the application thus qualifies as synthetic (rather than a posteriori) under most philosophical interpretations of this term.
  
\end{enumerate}

Thus when an expression $P$ is entered into the belief set, it is more exactly entered as an expression-label pair $(P,\lambda)$, where $\lambda$ is the label.  A DRS's language, axiom schemas, and inference rules comprise a {\it logic} in the usual sense. It is required that this logic be {\it consistent}, i.e., for no expression $P$ is it possible to derive both $P$ and $\lnot P$. The belief set may become inconsistent, nonetheless, through the introduction of contradictory extralogical axioms.

In what follows, only expressions representing a posteriori and synthetic knowledge may be retracted; expressions of a priori knowledge are taken as being held unequivocally.  Thus the term `a priori knowledge' is taken as synonymous with `belief held unequivocally', and `a posteriori knowledge' is interpreted as `belief possibly held only tentatively' (some a posteriori beliefs may be held unequivocally).  Thus the distinction between knowledge and belief is somewhat blurred, and what is referred to as a `belief set' might alternatively be called a `knowledge base', as is often the practice in AI systems.

\subsection {Controller}
  
A {\it controller} effectively determines the modeled agent's {\it purpose} or {\it goals} by managing the DRS's interaction with its environment and guiding the reasoning process.  With regard to the latter, the objectives typically include (i) deriving all expressions salient to the given application and entering these into the belief set, and (ii) ensuring that the belief set remains consistent.  To these ends, the business of the controller amounts to performing the following operations.

\begin{enumerate}

\item Receiving input from its environment, e.g., human users, sensors, or other artificial agents, expressing this input as expressions in the given language $\cal L$, and entering these expressions into the belief set in the manner described above (derivation path items (3) and (4)).  During this operation, new symbols are appropriated as needed to express concepts not already represented in the current $L_t$. 

\item Applying inference rules in accordance with some extralogical objective (some plan, purpose, or goal) and entering the derived conclusions into the belief set in the manner described above (derivation path items (1) and (2)). 

\item Performing any actions that may be prescribed as a result of the above reasoning process, e.g., moving a robotic arm, returning a response to a human user, or sending a message to another artificial agent.

\item Whenever necessary, applying a `dialectical belief revision' algorithm for contradiction resolution in the manner described below.

\item Applying any other belief revision algorithm as may be prescribed by the context of a particular application.

\end{enumerate}

\noindent In some systems, the above may include other types of belief revision operations, but such will not be considered in the present work. 

A {\it contradiction} is an expression of the form $P\land\lnot P$.  Sometimes it is convenient to represent the general notion of contradiction by the falsum symbol, $\bot$.  Contradiction resolution is triggered whenever a contradiction or a designated equivalent expression is entered into the belief set.  We may assume that this only occurs as the result of an inference rule application, since it obviously would make no sense to enter a contradiction directly as an extralogical axiom.  The contradiction resolution algorithm entails three steps:

\begin{enumerate}

\item Starting with the from-list in the label on the contradictory expression, backtrack through the belief set following from-lists until one identifies all extralogical axioms that were involved in the contradiction's derivation.  Note that such extralogical axioms must exist, since, by the consistency of the logic, the contradiction cannot constitute analytical knowledge, and hence must be synthetic. 

\item Change the belief status of one or more of these extralogical axioms, as many as necessary to invalidate the derivation of the given contradiction.  The decision as to which axioms to retract may be dictated, or at least guided by, the epistemic entrenchment values.  In effect, those expressions with the lower values would be preferred for retraction.  In some systems, this retraction process may be automated, and in others it may be human assisted.

\item Forward chain through to-lists starting with the extralogical axiom(s) just retracted, and retract all expressions whose derivations were dependent on those axioms.  These retracted expressions should include the contradiction that triggered this round of belief revision (otherwise the correct extralogical axioms were not retracted).

\end{enumerate}

This belief revision algorithm is reminiscent of Hegel's `dialectic', described as a process of `negation of the negation' \cite{hegel31}.  In that treatment, the latter (first occurring) negation is a perceived internal conflict (here a contradiction), and the former (second occurring) one is an act of transcendence aimed at resolving the conflict (here removing the contradiction).  In recognition of Hegel, the belief revision/retraction process formalized in the above algorithm will be called {\it Dialectical Belief Revision}.

\subsection{General Remarks}

Specifying a DRS requires specifying a path logic and a controller.  A path logic is specified by providing (i) a language $\cal L$, (ii) a set of axiom schemas, and (iii) a set of inference rules. A controller is specified by providing (i) the types of expressions that the DRS can receive as inputs from external sources, with each such type typically being described as expressions having a certain form, and (ii) for each such input type, an algorithm that is to be executed when the DRS receives an input of that type.  Such an algorithm typically involves applying inference rules, thereby deriving new formulas to be entered into the belief set, and it might specify other actions as well, such as moving a robotic arm, writing some information to a file, or returning a response to the user.  All controllers are assumed to include a mechanism for dialectical belief revision as described above.

Thus defined a DRS may be viewed as representing the `mind' of an intelligent agent, where this includes both the agent's reasoning processes and its memory. At any time $t$, the belief set $B_t$ represents the agent's conscious awareness as of that time. Since the extralogical axioms can entail inconsistencies, this captures the fact that an agent can `harbor' inconsistencies without being aware of this.  The presence of inconsistencies only becomes evident to the agent when they lead to a contradictory expression being explicitly entered into the belief set, in effect, making the agent consciously aware of a contradiction that was implicit in its beliefs.  This then triggers a belief revision process aimed at removing the inconsistency that gave rise to the contradiction.  

Depending on the application, the controller may be programmed to carry out axiom schema instantiations and perform derivations based on logical axioms.  Such might be the case, for example, if the logical rules were to include a `resolution rule' and the controller incorporated a Prolog-like theorem prover.  In many applications, however, it may be more appropriate to base the controller on a few suitably chosen derived rules.  The objective in this would be to simplify the controller's design by encapsulating frequently used argument patterns.  In such cases, the use of axiom schemas and logical inference rules is implicit, but no logical axioms per se need be entered into the derivation path.  Accordingly, all members of the belief set will be either a posteriori or synthetic and thus subject to belief revision.  This is illustrated in the examples that follow. 

\section{First-Order Logic}

\subsection{Formalism}

This section presents classical first-order logic (FOL) in a form suitable for incorporation into a DRS.  The treatment follows \cite{hamilton88}.  As symbols for the language $\cal L$ we shall have: {\it individual variables}, ${\bf x}_1,{\bf x}_2,\ldots$, denoted generically by $x,y,z$, etc.; {\it individual constants}, ${\bf a}_1,{\bf a}_2,\ldots$, denoted generically by $a,b,c$, etc.; {\it predicate symbols}, infinitely many for each arity (where arity is indicated by superscripts),  ${\bf A}^1_1,{\bf A}^1_2,\ldots; {\bf A}^2_1,{\bf A}^2_2,\ldots;\ldots$, denoted generically by $\alpha,\beta,\gamma$, etc.; {\it punctuation marks}, namely, the comma and left and right parentheses; the {\it logical connectives} $\lnot$ and $\to$; the ({\it universal}) {\it quantifier symbol}, $\forall$; and the {\it falsum symbol}, $\bot$.\footnote{This omits the customary {\it function symbols} as they are not needed for the examples discussed here.}

Here the {\it logical symbols} will be the individual variables, punctuation marks, logical connectives, quantifier symbol, and falsum symbol.  The {\it extralogical symbols} will be the individual constants and the predicate symbols.  Thus, as discussed in Section 2.1, the sublanguages $L_t$ of $\cal L$ will differ only in their choice of individual constants and predicate symbols.

Given a sublanguage $L$ of $\cal L$, the {\it terms} of $L$ will be the individual variables and the individual contants of $L$.  The {\it atomic formulas} of $L$ will be the falsum symbol and all expressions of the form $\alpha(t_1,\ldots,t_n)$ where $\alpha$ is an $n$-ary predicate symbol of $L$ and $t_1,\ldots,t_n$ are terms of $L$. The {\it formulas} of $L$ will be the atomic formulas of $L$ together with all expressions having the forms $(\lnot P)$, $(P\to Q)$, and $(\forall x)P$, where $P$ and $Q$ are formulas of $L$ and $x$ is an individual variable.  

Further logical connectives and the existential quantifier symbol can be introduced as means for abbreviating other formulas: \medskip

\indent\indent $(P\lor Q)$ for $((\lnot P)\to Q))$ \medskip
 
\indent\indent $(P\land Q)$ for $(\lnot(P\to(\lnot Q)))$ \medskip 

\indent\indent $(P\leftrightarrow Q)$ for $((P\to Q)\land(Q\to P))$ \medskip

\indent\indent $(\exists x)P$ for $(\lnot(\forall x)(\lnot P))$ \medskip

\noindent For readability, parentheses may be dropped according to (i) $\lnot$ takes priority over $\lor$ and $\land$, (ii) $\lor$ and $\land$ take priority over $\to$ and $\leftrightarrow$, and (iii) outermost surrounding parentheses are unneeded.    

In a formula of the form $(\forall x)Q$, the expression $(\forall x)$ is a ({\it universal}) {\it quantifier} and $Q$ is the {\it scope} of the quantifier.  If $x$ occurs in a formula $P$ within the scope of an occurrence of $(\forall x)$ in $P$, then that occurrence of $x$ is {\it bound} in $P$ by that occurrence of $(\forall x)$. If $x$ occurs in $P$ and is not bound by any quantifier, then that occurrence of $x$ is {\it free} in $P$. Note that the same variable $x$ can have both free and bound occurrences in the same formula $P$.  A formula that does not contain any free variable occurrences is {\it closed}.

If an occurrence of $x$ is free in $P$, then a different variable $y$ is {\it substitutable} for that occurrence of $x$ if the occurrence is not within the scope of the quantifier $(\forall y)$ (i.e., putting $y$ in place of $x$ does not create a binding of $y$). Note that this implies that $x$ is always substitutable for any of its own free occurrences in any $P$. An individual constant $a$ is {\it substitutable} for any free occurrence of any variable in any $P$. 

Where $P$ is a formula, $x$ is an individual variable, and $t$ is an individual term that is substitutable for $x$ in $P$, $P(t/x)$ denotes the formula obtained from $P$ by replacing all free occurrences of $x$ in $P$ with occurrences of $t$. Note that the above implies that, if $x$ does not occur free in $P$, or does not appear at all in $P$, then $P(t/x)$ is just $P$. Note also that, if $t$ is not substitutable for $x$ in $P$, then the notation $P(t/x)$ is undefined. 

This notation can be extended to arbitrarily many simultaneous replacements as follows.  Where $P$ is a formula, the variables $x_1,\ldots,x_n$ are distinct, and the terms $t_1,\ldots,t_n$ are substitutable for all the free occurrences of the respective variables $x_1,\ldots,x_n$ in $P$, $P(t_1,\ldots,t_n/x_1,\ldots,x_n)$ denotes the formula obtained from $P$ by replacing all occurrences of $x_1,\ldots,x_n$, respectively, with occurrences of $t_1,\ldots,t_n$.  

The {\it axiom schemas} will be the meta-level expressions (observing the same rules as for formulas for dropping parentheses):

\begin{description}

\item[$\bf (S 1)$] ${\cal A}\to({\cal B}\to{\cal A})$

\item[$\bf (S 2)$] $(({\cal A}\to({\cal B}\to{\cal C}))\to(({\cal A}\to{\cal B})\to({\cal A}\to{\cal C}))$

\item[$\bf(S 3)$] $(\lnot{\cal A}\to\lnot{\cal B})\to({\cal B}\to{\cal A})$

\item[$\bf (S 4)$] $\bot\leftrightarrow({\cal A}\land\lnot{\cal A})$

\item[$\bf (S 5)$] $(\forall {\bf x}){\cal A}\to{\cal B}$

\item[$\bf (S 6)$] $(\forall {\bf x})({\cal A}\to{\cal B})\to({\cal A}\to(\forall {\bf x}){\cal B})$

\end{description}

\noindent Let the {\it formula meta symbols} $\cal A, B, C$ be denoted generically by ${\cal A}_1,{\cal A}_2,\ldots$, and let {\bf x} be the {\it individual variable meta symbol}. Where $\bf S$ is a schema, let ${\bf S}(P_1,\ldots,P_n,x/{\cal A}_1,\ldots,{\cal A}_n,{\bf x})$ be the formula obtained from $\bf S$ by replacing all occurrences of ${\cal A}_1,\ldots,{\cal A}_n$, respectively, with occurrences of $P_1,\ldots,P_n$, and replacing each occurrence of {\bf x} with an occurrence of the individual variable $x$.  

The {\it inference rules} will be:

\begin{description}

\item[$\bf (R 1)$ \bf{Schema Instantiation 1}] Where {\bf S} is one of axiom schemas (1) through (4), infer ${\bf S}(P_1,\ldots,P_n/$ ${\cal A}_1,\ldots,{\cal A}_n)$, where ${\cal A}_1,\ldots,{\cal A}_n$ are all the distinct formula meta symbols occurring in $\bf S$, and $P_1,\ldots,P_n$ are formulas.

\item[$\bf (R 2)$ \bf{Schema Instantiation 2}] Where {\bf S} is axiom schema (5), infer ${\bf S}(P,P(t/x),x/{\cal A},{\cal B},{\bf x})$, where $P$ is any formula, $t$ is any individual term, and $x$ is any individual variable.

\item[$\bf (R 3)$ \bf {Schema Instantiation 3}] Where {\bf S} is axiom schema (6), infer ${\bf S}(P,Q,x/{\cal A},{\cal B},{\bf x})$, where $P,Q$ are any formulas and $x$ is any individual variable that does not occur free in $P$.

\item[$\bf (R 4)$ \bf {Modus Ponens}] From $P$ and $P\to Q$ infer $Q$, where $P,Q$ are any formulas.

\item[$\bf (R 5)$ {\bf Generalization}] From $P$ infer $(\forall x)P$, where $P$ is any formula and $x$ is any individual variable.

\end{description}

\noindent Regarding $(R 2)$, note first that, if $x$ occurs free in $P$, since $x$ is substitutable for itself in $P$, by taking $x$ for $t$ this rule allows one to derive $(\forall x)P\to P$. Next note that, if $x$ does not occur free in $P$, then, by definition of the notation $(t/x)$, $P(t/x)$ is just $P$, and the same rule allows one to to derive $(\forall x)P\to P$. It follows that all formulas of the form $(\forall x)P\to P$ are logical axioms.

It is not difficult to establish that, if one leaves out $(S 4)$, this formalism is equivalent to the first-order predicate calculus of Hamilton \cite{hamilton88}. The present $(S 1)$, $(S 2)$, $(S 3)$, and $(S 6)$ are identical to Hamilton's $(K 1)$, $(K 2)$, $(K 3)$, and $(K 6)$, and it can be seen that $(S 5)$ together with $(S 2)$ effectively replaces Hamilton's $(K 4)$ and $(K 5)$ under the present restricted notion of term that does not involve function symbols and the agreement that writing $P(t/x)$ implies that $t$ is substitutable for $x$ in $P$. To wit, the above note that, if $x$ does not occur free in $P$, then one can derive $(\forall x)P\to P$, gives $(K 4)$, and $(K 5)$ is obtained by the fact that, if $t$ is substitutable for $x$ in $P$, then one can derive $(\forall x)P\to P(t/x)$. Thus all of Hamilton's logical axioms are logical axioms here. Moreover, excluding $(S 4)$, the present formalism does not permit the introduction of logical axioms not found in Hamilton's calculus. In other words, with the exception of $(S 4)$, both formalisms have that same logical axioms.  Because of this equivalence, the present formalism can make use of numerous results proved in \cite{hamilton88}. Moreover, Hamilton's system differs from that of Mendelson \cite{mendelson87} only in Hamilton's $(K 3)$, and Exercise 4, Chapter 2 of \cite{hamilton88} shows that Hamilton's and Mendelson's systems are equivalent.  This allows appropriation of numerous results from \cite{mendelson87}. 

A ({\it first-order\/}) {\it theory} $T$ will consist of a sublanguage of the foregoing language $\cal L$, denoted $L_T$, the foregoing axiom schemas 1 through 6, the foregoing inference rules 1 through 6, and a set of formulas of $L_T$ to serve as {\it extralogical axioms} of $T$. By a {\it proof} in $T$ is meant a sequence $P_1,\ldots,P_n$ of formulas of $L_T$ such that each $P_i$ is either (i) a logical axiom, i.e., is derivable by means of one of the Schema Instantiation rules 1 through 3, (ii) an extralogical axiom, or (iii) can be inferred from formulas occurring before $P_i$ in the sequence by means of either Modus Ponens or Generalization. Such a sequence is a {\it proof of} the last member $P_n$. A formula of $L_T$ is a {\it theorem} of $T$ if it has a proof in $T$. The notation $T\vdash P$ is used to indicate that $P$ is a theorem of $T$, and $T{\not}\vdash P$ is used to indicate the contrary.

Consider an entry $(L_t,B_t)$ in the derivation path of a DRS as defined in Section 2.1.  A formula in $B_t$ will be {\it active} if its status is {\it bel}; otherwise it is {\it inactive}.   The {\it theory determined by} $(L_t,B_t)$ will be  first-order theory $T_t$ whose language is $L_t$ and whose extralogical axioms are the active  extralogical axioms in $B_t$. 

The Dialectical Belief Revision algorithm has the effect of changing the status of some formulas in the belief set from {\it bel} to {\it disbel}.  This is true also for other belief revision algorithms. Because of this there is no guarantee that the fact that the active extralogical axioms in $B_t$ are the extralogical axioms of $T_t$ implies that the active formulas in $B_t$ are all theorems of  $T_t$. That this implication can sometimes fail motivates the following.

A belief revision algorithm for a DRS is {\it normal} if, for any entry $(L_t,B_t)$ in a derivation path for the DRS, given that the active formulas in $B_t$ are theorems of the theory $T_t$ determined by $(L_t,B_t)$, the active formulas in the belief set $B_{t'}$ that results from an application of that algorithm will be theorems of the theory $T_{t'}$ determined by $(L_{t'},B_{t'})$.  \medskip

{\bf Proposition 3.1.} For any DRS, Dialectical Belief Revision is normal.  \medskip

A DRS is {\it normal} if all its belief revision algorithms are normal.  \medskip

{\bf Proposition 3.2.}  In a normal DRS, for each theory $T_t$ determined by a pair $(L_t,B_t)$ in a derivation path for the DRS, the active formulas in $B_t$ will be theorems of $T_t$.  \medskip 

Where $\Gamma$ is a set of formulas in $L_T$, let $T(\Gamma)$ be the theory obtained from $T$ by adjoining the members of $\Gamma$ as extralogical axioms. \medskip

{\bf Proposition 3.3.} Let $T_t$ be the theory determined by an entry $(L_t,B_t)$ in the derivation path for a normal DRS, let $T$ be the theory with language $L_t$ and no extralogical axioms, and let $\Gamma$ be the set of active formulas in $B_t$.  Then, for any formula $P$ of $L_t$, $T_t\vdash P$ iff $T(\Gamma)\vdash P$.  \medskip

It is assumed that the reader is familiar with the notion of {\it tautology} from the Propositional Calculus (PC).  Axiom schemas $(S 1)$, $(S 2)$, and $(S 3)$, together with Modus Ponens, are the axiomatization of PC found in  \cite{hamilton88}.  Where $T$ is a theory and $P$ is a formula of $L_T$, let $T_{\rm PC}\vdash P$ indicate that $P$ has a proof in $T$ using only $(S 1)$, $(S 2)$, and $(S 3)$, Instantiation 1, and Modus Ponens. Then, by treating atomic formulas of $\cal L$ as propositions of PC, one has the following.  \medskip

{\bf Theorem 3.1.} ({\it Soundness Theorem for PC\/}) For any theory $T$, if $T_{\rm PC}\vdash P$, then $P$ is a tautology.  \medskip

{\bf Theorem 3.2.} ({\it Adequacy Theorem for PC\/}) For any theory $T$, if $P$ is a tautology, then $T_{\rm PC}\vdash P$. \medskip

These theorems can be used to show that axiom schema $(S4)$ serves merely as a defining axiom for $\bot$ and does not enable proving any additional formulas not involving $\bot$. Let $T_{{\rm PC(}\bot{\rm )}}\vdash P$ indicate that $P$ has a proof in $T$ using only $(S 1)$, $(S 2)$, $(S 3)$, and $(S 4)$, Schema Instantiation 1, and Modus Ponens.  \medskip

{\bf Proposition 3.4.} If $P$ does not contain any occurrences of $\bot$ and $T_{{\rm PC(}\bot{\rm )}}\vdash P$, then $T_{\rm PC}\vdash P$. \medskip

A theorem of $T$ is said to be {\it derivable} in $T$. A formula $P$ is said to be {\it derivable from} $\Gamma$ in $T$ if $T(\Gamma)\vdash P$. Note that this is equivalent to saying that, if $T\vdash Q$, for all $Q\in\Gamma$, then $T\vdash P$.  When $\Gamma$ is given as a list of one or more formulas, e.g., $\{P_1,\ldots,P_n\}$, the surrounding braces will be dropped, i.e., $T(\{P_1,\ldots,P_n\})$ will be shortened to $T(P_1,\ldots,P_n)$. 

An inference rule is a statement of the form `From some {\it premises} having some forms ${\cal A}_1,\ldots,{\cal A}_n$, infer the {\it conclusion} having some form $\cal A$'.  In the context of a theory $T$, an inference rule may be viewed as a mapping from $L_T$ into itself, with an {\it application} of the rule being represented as an $n+1$-tuple of formulas of $L_T$, $(P_1,\ldots,P_n,P)$, where $P_1,\ldots,P_n$ are the premises and $P$ is the conclusion.  An inference rule is {\it valid} (or {\it derivable}) in a theory $T$, if the conclusion is always derivable from the premises in $T$, i.e., if $T(P_1,\ldots,P_n)\vdash P$.  Consider the following inference rules for an arbitrary $T$.

\begin{description}

\item[\kern 2em{\bf Hypothetical Syllogism}] From $P\to Q$ and $Q\to R$ infer $P\to R$, where $P,Q,R$ are any formulas.

\item[\kern 2em {\bf Aristotelian Syllogism}] From $(\forall x)(P\to Q)$ and $P(a/x)$, infer $Q(a/x)$, where $P,Q$ are any formulas, $x$ is any individual variable, and $a$ is any individual constant.

\item[\kern 2em{\bf Subsumption}] From $(\forall x)(\alpha(x)\to\beta(x))$ and $(\forall x)(\beta(x)\to\gamma(x))$, infer $(\forall x)(\alpha(x)\to\gamma(x))$, where $\alpha,\beta,\gamma$ are any unary predicate symbols, and $x$ is any individual variable.

\item[\kern 2em{\bf And-Introduction}] From $P$ and $Q$ infer $P\land Q$. 

\item[\kern 2em{\bf And-Elimination}] From $P\land Q$ infer $P$ and $Q$. 

\item[\kern 2em{\bf Conflict Detection}] From $(\forall x)\lnot(P\land Q)$, $P(a/x)$, and $Q(a/x)$ infer $\bot$, where $P,Q$ are any formulas, $x$ is any individual variable, and $a$ is any individual constant.

\item[\kern 2em{\bf Contradiction Detection}] From $P$ and $\lnot P$ infer $\bot$, where $P$ is any formula.

\end{description}  

\noindent Hypothetical Syllogism is a well-known principle of classical logic. Aristotelian Syllogism captures the reasoning embodied in the famous argument `All men are mortal; Socrates is a man; therefore Socrates is mortal', by taking $P$ for `is a man', $Q$ for `is mortal', and $a$ for `Socrates'.  A concept $A$ {\it subsumes} concept $B$ if the set of objects represented by $A$ contains the set represented by $B$ as a subset.  Thus Subsumption captures the transitivity of this subsumption relationship.  In the context of a DRS, Conflict Detection can be used for triggering Dialectical Belief Revision.  This is an example on one such triggering rule; others surely are possible. 

Aristotelian Syllogism, Subsumption, and Conflict Detection will be used in the Document Management Assistant application developed in Section 4.  In this respect, they may be considered to be application-specific.  They are not domain-specific, however, inasmuch as they happen to be valid in any first-order theory, as demonstrated by the following. \medskip

{\bf Proposition 3.5.}  The above seven inference rules are valid in any theory $T$. \medskip

A theory $T$ is {\it inconsistent} if there is a formula $P$ of $L_T$ such that both $T\vdash P$ and $T\vdash\lnot P$; otherwise $T$ is {\it consistent}. \medskip

{\bf Proposition 3.6.} A theory $T$ is consistent iff there is a formula $P$ of $L_T$ such that $T{\not}\vdash P$. \medskip

\noindent This shows that any inconsistent system with the full strength of first-order logic is trivial in that all formulas are formally derivable.  

A set $\Gamma$ of formulas of a language $L$ is {\it consistent} if the theory $T$ with language $L$ and with the formulas in $\Gamma$ as extralogical axioms is consistent; otherwise $\Gamma$ is inconsistent.  For an entry $(L_t,B_t)$ in the derivation path of a DRS, $B_t$ is {\it consistent} if the set of active formulas in $B_t$ is consistent; otherwise $B_t$ is {\it inconsistent}.  \medskip

{\bf Proposition 3.7.} For an entry $(L_t,B_t)$ in the derivation path of a normal DRS, $B_t$ is consistent if and only if the theory $T_t$ determined by $(L_t,B_t)$ is consistent. \medskip

\subsection{Semantics and Main Results Regarding First-Order Logic}

This formulation of a semantics for first-order logic follows \cite{shoenfield67}. Let $T$ be a theory of the kind described above. An {\it interpretation} $I$ for the language $L_T$ consists of (i) a nonempty set $D_I$ serving as the {\it domain} of $I$, the elements of which are called {\it individuals}, (ii) for each individual constant $a$ of $L_T$, assignment of a unique individual $I(a)\in D_I$, and (iii) for each $n$-ary predicate symbol $\alpha$ of $L_T$, assignment of an $n$-ary relation $I(\alpha)$ on $D_I$. For each individual $d\in D_I$, let $\hat d$ be a new individual constant, i.e., one not among the individual constants of $L_T$, to serve as the {\it name} of $d$. Let $L_T(I)$ be the language obtained from $L_T$ by adjoining the names of the individuals in $D$ as new extralogical symbols.  Where $I$ is an interpretation for $L_T$, let $I(\hat d)=d$ for all $d\in D$. For $P$ a formula of a language $L$ and $I$ and interpretation for $L$, an {\it $I$-instance} of $P$ is a formula of the form $P(\hat d_1,\ldots,\hat d_n/x_1,\ldots,x_n)$ where $x_1,\ldots,x_n$ are the distinct individual variables occurring free in $P$ and $d_1,\ldots,d_n\in D_I$. Note that $I$-instances as defined above are closed. 

Given an interpretation $I$ for a language $L$, a {\it truth valuation} is a mapping  $v_I:{\rm closed\_formulas\_}$ ${\rm of\_}L(I)\to\{{\bf T}, {\bf F}\}$ satisfying:

\begin{enumerate}

\item For $P$ being the atomic formula $\bot$, $v_I(P)={\bf F}$.

\item For $P$ atomic and having the form $\alpha(t_1,\ldots,t_n)$, $v(P)={\bf T}$ iff $(I(t_1),\ldots,I(t_n))\in I(\alpha)$ (the relation $I(\alpha)$ holds for the $n$-tuple $(I(t_1),\ldots,I(t_n))$.  Note that, since $P$ is closed, the $t_i$ must all be individual constants and may be names of individuals in $D_I$.

\item For $P$ of the form $\lnot Q$, $v_I(P)={\bf T}$ iff $v_I(Q)={\bf F}$.

\item For $P$ of the form $Q\to R$, $v_I(P)={\bf T}$ iff either $v_I(Q)={\bf F}$ or $v_I(R)={\bf T}$.

\item For $P$ of the form $(\forall x)Q$, $v_I(P)={\bf T}$ iff $v_I(Q(\hat d/x))={\bf T}$ for every $d\in D_I$.

\end{enumerate}

\noindent Items 3 and 4 encapsulate the usual truth tables for $\lnot$ and $\to$. By definition of the various abbreviated forms, it will follow that, for closed $P$: \medskip

\indent\indent  For $P$ of the form $Q\land R$, $v_I(P)={\bf T}$ iff $v_I(Q)={\bf T}$ and $v_I(R)={\bf T}$.

\indent\indent  For $P$ of the form $Q\lor R$, $v_I(P)={\bf T}$ iff either $v_I(Q)={\bf T}$ or $v_I(R)={\bf T}$.

\indent\indent  For $P$ of the form $Q\leftrightarrow R$, $v_I(P)={\bf T}$ iff $v_I(Q)=v(R)$.

\indent\indent  For $P$ of the form $(\exists x)Q$, $v_I(P)=v((\lnot(\forall x)(\lnot Q)))$. \medskip

A closed formula $P$ is {\it true} or {\it valid} in an interpretation $I$ if $v_I(P)={\bf T}$. An open formula $P$ is {\it valid} in an interpretation $I$ if $v_I(P')={\bf T}$, for all $I$-instances $P'$ of $P$. The notation $I\models P$ is used to denote that $P$ is valid in $I$. A formula $P$ of a language $L$ is {\it logically valid} if it is valid in every interpretation of $L$. \medskip

 \medskip

{\bf Proposition 3.8.} Let $P$ be a formula of $L_T$ in which no individual variable other than $x$ occurs free, let $a$ be an individual constant of $L_T$, let $I$ be an interpretation of $L_T$, and let $d=I(a)$. Then $I\models P(a/x)$ iff $I\models P(\hat d/x)$. \medskip

{\bf Proposition 3.9.} For any language $L$, the logical axioms of $L$, i.e., all formulas derivable by means of the schema instantiation rules $(R 1)$ through $(R 3)$, are logically valid. \medskip

An inference rule is {\it validity preserving} if, for every application $(P_1,\ldots,P_n,P)$ comprised of formulas in a language $L_T$ for a theory $T$, and for every interpretation $I$ for $L_T$, $I\models P_1,\ldots,I\models P_n$ implies $I\models P$. \medskip

{\bf Proposition 3.10.} Modus Ponens and Generalization are validity preserving. \medskip 

{\bf Theorem 3.3.} ({\it Soundness Theorem for FOL\/}) Let $T$ be a theory with no extralogical axioms and let $P$ be a formula of $L_T$.  If $T\vdash P$, then $P$ is logically valid. \medskip

One can establish the converse, referred to in \cite{hamilton88} as the {\it Adequacy Theorem} (Proposition 4.41). This also appears in \cite{mendelson87} as Corollary 2.18. The proof in \cite{hamilton88} can be adapted to the present system because of the equivalence between that system and the formalism studied here.  It is only necessary to verify that the present notion of semantic interpretation is equivalent to that of \cite{hamilton88}. This amounts to observing that the present notion of $I$-instance is equivalent with the notion of `valuation' in \cite{hamilton88}. Details are omitted as this result in not needed in the present work.

An interpretation $I$ for the language of a theory $T$ is a {\it model} of $T$ if all theorems of $T$ are valid in $I$. The notation $I\models T$ expresses that $I$ is a model of $T$. \medskip

{\bf Theorem 3.4.} ({\it Consistency Theorem\/}) If a theory $T$ has a model, then $T$ is consistent. \medskip

{\bf Proposition 3.11.} If $T$ is a theory with no extralogical axioms, then $T$ is consistent.  \medskip

An interpretation for a language $L$ is a {\it model} for a set $\Gamma$ of formulas of $L$ if all the formulas in $\Gamma$ are valid in $I$.  The notation $I\models\Gamma$ expresses that $I$ is a model of $\Gamma$. \medskip

{\bf Proposition 3.12.} Let $\Gamma$ be the extralogical axioms of a theory $T$ and let $I$ be an interpretation for $L_T$. If $I\models\Gamma$, then $I\models T$. \medskip

For an entry $(B_t,L_t)$ in the derivation path for a DRS (Section 2.1), an interpretation $I$ for $L_t$ will be a {\it model} of $B_t$ if $I\models\Gamma$ where $\Gamma$ is the set of active formulas in $B_t$.  Let $I\models B_t$ indicate that $I$ is a model of $B_t$.  \medskip

{\bf Proposition 3.13.} Let $(L_t,B_t)$ be an entry in a derivation path for a normal DRS. If there is an interpretation $I$ of $L_t$ that is a model of $B_t$, then $B_t$ is consistent. \medskip

\section{Example 1: A Document Management Assistant} 

A DRS is used to represent the reasoning processes of an artificial agent interacting with its environment.  For this purpose the belief set should include a model of the agent's environment, with this model evolving over time as the agent acquires more information about the environment. This section illustrates this idea with a simple DRS based on first-order logic representing an agent that assists its human users in creating and managing a taxonomic document classification system.  Thus the environment in this case consists of the document collection together with its users.  The objective in employing such an agent is to build concept taxonomies suitable for browsing.  In this manner the DRS functions as a {\it Document Management Assistant} (DMA).

In the DMA, documents are represented by individual constants, and document classes are represented by unary predicate symbols.  Membership of document $a$ in class $\alpha$ is expressed by the atomic formula $\alpha(a)$; the property of class $\alpha$ being a subset of class $\beta$ is expressed by $(\forall x)(\alpha(x)\to\beta(x))$, where $x$ is any individual variable; and the property of two classes $\alpha$ and $\beta$ being disjoint is expressed by $(\forall x)(\lnot(\alpha(x)\land\beta(x)))$, where $x$ is any individual variable.  A taxonomic classification hierarchy may thus be constructed by entering formulas of these forms into the belief set as extralogical axioms.  It will be assumed that these axioms are input by human users.

In addition to the belief set, the DMS will employ an extralogical graphical structure representing the taxonomy.  A formula of the form $\alpha(a)$ will be represented by an {\it is-an-element-of} link from a node representing $a$ to a node representing $\alpha$, a formula of the form $(\forall x)(\alpha(x)\to\beta(x))$ will be represented by an {\it is-a-subclass-of} link from a node representing $\alpha$ to a node representing $\beta$, and a formula of the form $(\forall x)(\lnot(\alpha(x)\land\beta(x)))$ will be represented by an {\it are disjoint} link between some nodes representing $\alpha$ and $\beta$.  This structure will be organized as a directed acyclic graph (DAG) without redundant links with respect to the is-an-element-of and is-a-subclass-of links (i.e., ignoring are-disjoint links), where by a redundant link is meant a direct link from some node to an ancestor of that node other than the node's immediate ancestors (i.e., other than its parents).  To this end the controller will maintain a data structure that represents the current state of this graph.  Whenever an axiom expressing a document-class membership is entered into the belief set, a corresponding is-an-element-of link will be entered into the graph, unless this would create a redundant path.  Whenever an axiom expressing a subclass-superclass relationship is entered into the belief set, an is-a-subclass-of link will be entered into the graph, unless this would create either a cycle or a redundant path. Whenever an axiom expressing class disjointedness is entered into the belief set, a corresponding link expressing this will be entered into the graph.  To accommodate this activity, the derivation path as a sequence of pairs is augmented to become a sequence of triples $(L_t,B_t,G_t)$, where $G_t$ is the state of the graph at time $t$.

This is illustrated in Figure 4.1, showing a graph that would be defined by entering the following formulas into the belief set, where TheLibrary, Science, Engineering, Humanities, ComputerScience, Philosophy, and ArtificialIntelligence are taken as alternative labels for the predicate letters ${\bf A}^1_1,\ldots,{\bf A}^1_7$, where Doc1, Doc2, Doc3 are alternative labels for the individual constants ${\bf a}_1,{\bf a}_2,{\bf a}_3$, and where $x$ is the individual variable ${\bf x}_1$: \medskip

\indent\indent $(\forall x)({\rm Science}(x)\to{\rm TheLibrary}(x))$

\indent\indent $(\forall x)({\rm Engineering}(x)\to{\rm TheLibrary}(x))$

\indent\indent $(\forall x)({\rm Humanities}(x)\to{\rm TheLibrary}(x))$

\indent\indent $(\forall x)({\rm ComputerScience}(x)\to{\rm Science}(x))$

\indent\indent $(\forall x)({\rm ComputerScience}(x)\to{\rm Engineering}(x))$

\indent\indent $(\forall x)({\rm Philosophy}(x)\to{\rm Humanities}(x))$

\indent\indent $(\forall x)({\rm ArtificialIntelligence}(x)\to {\rm ComputerScience}(x))$

\indent\indent $(\forall x)\lnot({\rm Engineering}(x)\land{\rm Humanities}(x))$

\indent\indent ${\rm Science}({\rm Doc1})$

\indent\indent ${\rm Engineering}({\rm Doc1})$

\indent\indent ${\rm ArtificialIntelligence}({\rm Doc2})$

\indent\indent ${\rm Philosophy}({\rm Doc3})$

\medskip

The overall purpose of the DMA is to support browsing and search by the human users.  The browsing capability is provided by the graph.   For this one would develop tools for `drilling down' through the graph, progressively narrowing in on specific topics of interest.  User queries would consist of keyword searches and would employ the belief set directly, i.e., they do not necessarily require the graph.  In such queries, the keywords are presumed to be the names of classification categories.  The algorithms associated with the DMA's controller are designed to derive all document-category classifications implicit in the graph and enter these into the belief set.  These document-category pairs can be stored in a simple database, and keyword searches, possibly involving multiple keywords connected by `or' or `and' can be implemented as database queries.  For `and' queries, however, one may alternatively use the graph structure to find those categories' common descendants. 

As described in Section 2.1, entering a formula into the belief set also entails possibly expanding the current language by adding any needed new symbols and assigning the new formula an appropriate label.  For the above formulas, the from-list will consist of an indication that the source of the formula was a human user.  Let us use the code $hu$ for this (as an alternative to the aforementioned {\it es}).  As an epistemic entrenchment value, let us arbitrarily assign each formula the value 0.5 on the scale $[0,1]$. Thus each label will have the form \medskip

\indent\indent $\{t,\{{\it hu}\}, \emptyset, {\it bel}, 0.5, {\it a\ posteriori}\}$ \medskip

\noindent Using $0.5$ for the epistemic entrenchment value effectively makes these values nonfunctional with respect to Dialectical Belief Revision.  When a choice must be made regarding which of several formulas to disbelieve, this choice will either be random or made by the human user.  

\begin{figure}[htp]
\centerline{\includegraphics[height=2.95in]{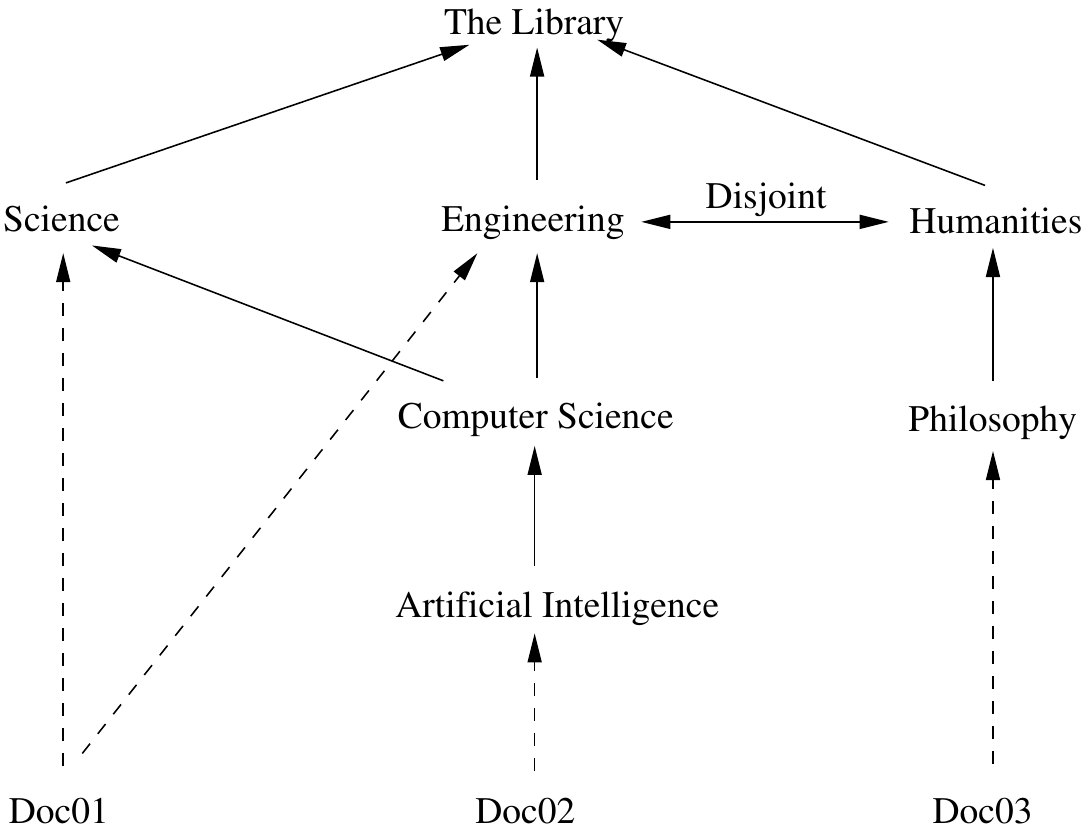}}
\medskip
\centerline{Figure 4.1. A taxonomy fragment.}
\end{figure}  

Whenever the user wishes to enter a new formula into the belief set, the formula is provided to the controller, which executes a reasoning process depending on the type (or form) of the formula. This process may lead to the controller's carrying out inference rule applications, the results of which are provided back to the controller, which may in turn lead to further rule applications and/or belief revision processes.  In this activity the controller additionally modifies the language and graph as appropriate. As discussed in Section 2.2, the general purpose of a controller is two-fold: (i) to derive all salient information for the intended application, and (ii) to ensure that the belief set remains consistent.  In the present case, the salient information is the graph together with the document classifications implicit in the graph, i.e., all formulas of the form $\alpha(a)$ that can be derived from the formulas describing the graph.  In an application, these category-document pairs can be stored in a simple database and used for keyword-based sorting and retrieval (search). 

\subsection{Formal Specification of the DMA}

These considerations motivate the following formal specification for the DMA.  For the path logic, let the language be the language $\cal L$ for first-order logic defined in Section 3.1, let the axiom schemas be $(S 1)$ through $(S 6)$, and let the inference rules be $(R 1)$ through $(R 5)$ together with Aristotelian Syllogism, Subsumption, and Conflict Detection. In accordance with Section 2.1, let $L_0$ be the minimal sublanguage of $\cal L$ consisting of all formulas that can be built up from the atomic formula $\bot$, and let $B_0=\emptyset$.  In addition, let $G_0=\emptyset$. 

For the controller, all inputs by human users must be formulas having one of the forms (i) $\alpha(a)$, (ii) $(\forall x)(\alpha(x)\to\beta(x))$, where $\alpha$ and $\beta$ are distinct, and (iii) $(\forall x)\lnot(\alpha(x)\land\beta(x))$, where $\alpha$ and $\beta$ are distinct.  As mentioned, part of the function of the controller is to maintain the graphs that are represented in the derivation path by the $G_t$.  These graphs have two types of nodes: one type representing documents corresponding to individual constant symbols, and one type representing classification categories corresponding to unary predicate symbols.  There are three kinds of links: is-an-element-of links from documents to categories, is-a-subclass-of links between categories, and are-disjoint links between categories.  It is desired that the controller maintain these graphs, ignoring are-disjoint links, as directed acyclic graphs without redundant links, as described above.

To complete the specification of the controller it is necessary to provide algorithms that will be executed depending on the type of user input. In the present system, it is convenient to distinguish between two kinds of input {\it event}, the first being when a formula is provided to the controller by a human user, and the second being when a formula is provided to the controller as the result of an inference rule application. The following describes algorithms associated with five such event types. Of these, Event Types 1, 4, and 5 correspond to user inputs. These are initiating events, each of which may lead to a sequence of events of some of the other types. 

In all the events it is assumed that, if the formula provided to the controller already exists and is active in the current belief set, its input is immediately rejected.  This prevents unnecessary duplicates. In each of the following, assume that the most recent entry into the derivation path is $(L_t,B_t,G_t)$. \medskip

{\bf Event Type 1:}  A formula of the form $\alpha(a)$ is provided to the controller by a human user.  If either $\alpha$ or $a$ is not in the symbol set for $L_t$, form $L_{t+1}$ by adding the missing ones to the symbol set for $L_t$; otherwise set $L_{t+1}=L_t$.  Form $B_{t+1}$ from $B_t$ by adding the labeled formula $(\alpha(a), \{t+1,\{{\it hu}\}, \emptyset, {\it bel}, 0.5, {\it a\ posteriori}\})$. If there are no nodes representing either $a$ or $\alpha$ in $G_t$, form $G_{t+1}$ by adding such nodes together with an is-an-element-of link from the $a$ node to the $\alpha$ node.  If one of $a$ and $\alpha$ is represented by a node in $G_t$ but the other is not, form $G_{t+1}$ by adding a node representing the one that is missing together with an is-an-element-of link from the $a$ node to the $\alpha$ node.  Note that if a category node is being added, this will become a root node.  If both $a$ and $\alpha$ are represented by nodes in $G_t$, form $G_{t+1}$ by adding an is-an-element-of link from the $a$ node to the $\alpha$ node, unless this would create a redundant path in the graph.

 Search $B_{t+1}$ for active formulas of the forms $(\forall x)\lnot(\alpha(x)\land\beta(x))$ or $(\forall x)\lnot(\beta(x)\land\alpha(x))$, where $\alpha$ is the predicate symbol of the input formula and $\beta$ is some predicate symbol other than $\alpha$, and, if found, search for an active occurrence of $\beta(a)$, where $a$ is the individual constant of the input formula, and for each successful such search, apply Conflict Detection to infer $\bot$ and provide this formula to the controller.  This is an event of Type 2. 

Let $B_{t^*}$ be the most recent belief set, i.e., either it is $B_{t+1}$ or it is the belief set that has resulted from the processes associated with the indicated events of Type 2, if any occurred. If the input formula $\alpha(a)$ is still active, search $B_{t^*}$ for any active formulas having the form $(\forall x)(\alpha(x)\to\beta(x))$, where $\alpha$ is the predicate symbol of the input formula.  For each such formula, apply Aristotelian Syllogism to this and the formula $\alpha(a)$ to infer $\beta(a)$, and provide this formula to the controller. This is an event of Type 3. \medskip

{\bf Event Type 2:} The formula $\bot$ is provided to the controller as the result of an application of Conflict Detection.  Let $L_{t+1}=L_t$.  Form $B_{t+1}$ from $B_t$ by (i) adding the labeled formula $(\bot, \{t+1,F, \emptyset, {\it bel}, 0.5,$ ${\it a\ posteriori}\})$, where the from-list $F$ contains the name of the inference rule (Conflict Detection) that was used to conclude this occurrence of $\bot$, together with the indexes of the formulas that served as premises in the rule application, and (ii) updating the to-lists of all formulas that thus served as premises by including the index $t+1$. Let $G_{t+1}=G_t$. 

Now invoke the Dialectical Belief Revision algorithm on $B_{t+1}$ as described in Section 2.2, starting with the formula $\bot$ just added to the belief set.  As a result of this process, some formulas in the current belief set will have their status changed from {\it bel} to {\it disbel}.  Let $B_{t+2}$ be the belief set obtained from $B_{t+1}$ by making these changes in the relevant formulas' labels.  Let $L_{t+2}=L_{t+1}$.  Obtain $G_{t+2}$ from $G_{t+1}$ by removing any elements representing formulas whose statuses have thus been changed to {\it disbel}.  Specifically, (i) if a formula of the form $\alpha(a)$ is disbelieved, remove the is-an-element-of link connecting the node representing $a$ to the node representing $\alpha$, and remove the node representing $a$, unless it is connected to some node other than the one representing $\alpha$, and (ii) if a formula of the form $(\forall x)(\alpha(x)\to\beta(x))$ is disbelieved, remove the is-a-subclass-of link connecting the node representing $\alpha$ to the node representing $\beta$, and  remove the node representing $\alpha$, unless it is connected to some node other than the one representing $\beta$.  \medskip

{\bf Event Type 3:} A formula of the form $\alpha(a)$ is provided to the controller as a result of an inference rule application (Aristotelian Syllogism).  In this case, both $\alpha$ and $a$ are already in $L_t$, so let $L_{t+1}=L_t$.  Form $B_{t+1}$ from $B_t$ by (i) adding the labelled formula $(\alpha(a), \{t+1,F, \emptyset, {\it bel}, 0.5, {\it a\ posteriori}\})$, where the from-list $F$ contains the name of the inference rule that was used to infer $\alpha(a)$ (Aristotelian Syllogism), together with the indexes of the formulas that served as premises in the rule application, and (ii) updating the to-lists of all formulas that thus served as premises by including the index $t+1$. Let $G_{t+1}=G_t$.  Note that no modification of the graph is warranted, since the membership of the document associated with $a$ in the category associated with $\alpha$ is already implicit in the graph, so that a link between the respective nodes would form a redundant path.

Search $B_{t+1}$ for active formulas of the forms $(\forall x)\lnot(\alpha(x)\land\beta(x))$ or $(\forall x)\lnot(\beta(x)\land\alpha(x))$, where $\alpha$ is the predicate symbol of the input formula and $\beta$ is some predicate symbol other that $\alpha$, and, if found, search for an active occurrence of $\beta(a)$, where $a$ is the individual constant of the input formula, and for each successful such search, apply Conflict Detection to infer $\bot$ and provide this formula to the controller.  This is an event of Type 2.

Let $B_{t^*}$ be the most recent belief set, i.e., either it is $B_{t+1}$ or it is the belief set that has resulted from the processes associated with the indicated events of Type 2, if any occurred. If the input formula $\alpha(a)$ is still active, search $B_{t^*}$ for any active formulas having the form $(\forall x)(\alpha(x)\to\beta(x))$, where $\alpha$ is the predicate symbol of the input formula.  For each such formula, apply Aristotelian Syllogism to this and the formula $\alpha(a)$ to infer $\beta(a)$, and provide this formula to the controller. This is a recursive occurrence of an event of Type 3.  \medskip

{\bf Event Type 4:}  A formula of the form $(\forall x)(\alpha(x)\to\beta(x))$ is provided to the controller by a human user.  If both $\alpha$ and $\beta$ are already in $L_t$, begin by performing as many as possible of the following three actions.  First, search $B_t$ to determine if either of the formulas $(\forall x)\lnot(\alpha(x)\land\beta(x))$ and $(\forall x)\lnot(\beta(x)\land\alpha(x))$ are active, and, if so, reject the input and inform the user that the input is disallowed inasmuch as it contradicts the current belief set.  Second, explore all ancestors of $\alpha$ to see if they include $\beta$, and, if so, reject the input and inform the user that the input is disallowed inasmuch as it would create a redundant path in the subsumption hierarchy. Third, explore all ancestors of $\beta$ as expressed by formulas in $B_t$ to determine whether these include $\alpha$, and, if so, reject the input and inform the user that the input is disallowed inasmuch as it would create a loop in the subsumption hierarchy.   If the input is not rejected for any of these reasons, do the following.
  
If either $\alpha$ or $\beta$ is not in the symbol set for $L_t$, form $L_{t+1}$ by adding the ones that are missing, otherwise let $L_{t+1}=L_t$.  Form $B_{t+1}$ from $B_t$ by adding the labeled formula $((\forall x)(\alpha(x)\to\beta(x)),\{t+1,\{{\it hu}\}, \emptyset, {\it bel}, 0.5, {\it a\ posteriori}\})$. If there are no nodes representing either $\alpha$ or $\beta$ in $G_t$, form $G_{t+1}$ by adding such nodes together with an is-a-subclass-of link from the $\alpha$ node to the $\beta$ node.  If one of $\alpha$ and $\beta$ is represented by a node in $G_t$ but the other is not, form $G_{t+1}$ by adding a node representing the one that is missing together with an is-a-subclass-of link from the $\alpha$ node to the $\beta$ node. If both $\alpha$ and $\beta$ are represented by nodes in $G_t$, form $G_{t+1}$ by adding an is-a-subclass-of link from the $\alpha$ node to the $\beta$ node.

Now search $B_{t+1}$ for any active formulas of the form $\alpha(a)$ where $\alpha$ is the predicate symbol in the input formula, and, for each such formula, apply Aristotelian Syllogism to infer $\beta(a)$, and provide this to the controller. This is an event of Type 3. \medskip

{\bf Event Type 5:} A formula of the form $(\forall x)(\lnot(\alpha(x)\land\beta(x)))$ is provided to the controller by a user. If either $\alpha$ or $\beta$ is not in the symbol set for $L_t$, form $L_{t+1}$ by adding the ones that are missing, otherwise let $L_{t+1}=L_t$.  Form $B_{t+1}$ from $B_t$ by adding the labeled formula $((\forall x)(\lnot(\alpha(x)\land\beta(x))),\{t+1,\{{\it hu}\}, \emptyset, {\it bel}, 0.5, {\it a\ posteriori}\})$. If there are no nodes representing either $\alpha$ or $\beta$ in $G_t$, form $G_{t+1}$ by adding such nodes together with an are-disjoint link between the two nodes.  If one of $\alpha$ and $\beta$ is represented by a node in $G_t$ but the other is not, form $G_{t+1}$ by adding a node representing the one that is missing together with an are-disjoint link between the two nodes. If both $\alpha$ and $\beta$ are represented by nodes in $G_t$, then form $G_{t+1}$ by adding an are-disjoint link between the two nodes.  

Having accomplished this, search $B_{t+1}$ for active formulas of the forms $\alpha(a)$ and $\beta(a)$, and, if found, apply Conflict Detection to infer the formula $\bot$, and provide this to the controller.  This is an event of Type 2. \medskip

It may be noted that the above does not provide for the user's removing links from the graph, or more exactly, changing the status of an extralogical axiom from {\it bel} to {\it disbel}.  An ability to do so would obviously be desirable in any practical application.  In particular, if the user wished to modify the graph by inserting a new category represented by $\gamma$ between two existing categories represented by $\alpha$ and $\beta$, one would need to remove the link between (the categories represented by) $\alpha$ and $\beta$, and then add a link from $\alpha$ to $\gamma$ and a link from $\gamma$ to $\beta$.  It is not difficult to see that removing a link can be handled in a straightforward manner, simply by following to-lists starting with the formula $(\forall x)(\alpha(x)\to\beta(x))$ representing the link between $\alpha$ and $\beta$ and changing all formulas whose derivations relied on that formula to {\it disbel}.  In effect, this undoes the event of adding the link in question, as well as any further additions to the belief set that may have been based on the presence of the link.

Note also that none of the given events employ the Subsumption rule. This is because, in the present example, information regarding subsumption among the categories was not included as part of the `salient information'. Such could be added to a future example, but this would make the associated algorithms more complex.  \medskip

{\bf Proposition 4.1.} The DMA is a normal DRS.  \medskip

\subsection {Illustration}

The application of the algorithms associated with the foregoing events can be illustrated by considering the inputs needed to create the concept taxonomy shown in Figure 4.1.  Let us abbreviate `TheLibrary', `Science', `Engineering', `Humanities', `ComputerScience', `Philosophy', and `ArtificialIntelligence', respectively, by `TL', `S', `E', `H', `CS', `P', and `AI'.  In accordance with the definition of derivation path in Section 2.1, the language $L_0$ will be the language generated by the logical symbols given in Section 3.1, i.e., by $\sigma_0=\{{\bf x}_1,{\bf x}_2,\ldots, \hbox{\rm `,'},\hbox{\rm `('}, \hbox{\rm `)'}, \lnot, \lor, \forall, \bot\}$. This means that the only formula in $L_0$ is $\bot$.  Also in accordance with the Section 2.1, belief set $B_0=\emptyset$.  In accordance with the definition of the DMA, set $G_0=\emptyset$.

Consider an input of the first formula in the foregoing list, namely, $(\forall x)({\rm S}(x)\to{\rm TL}(x))$. This is an event of Type 4.  The language $L_1$ is formed from $L_0$ by adding the symbols S and TL (or, more exactly, the predicate letters ${\bf A}^1_2$ and  ${\bf A}^1_1$), i.e., $\sigma_1=\sigma_0\cup\{{\rm S}, {\rm TL}\}$.  The belief set $B_1$ is formed from $B_0$ by adding the labeled formula $((\forall x)({\rm S}(x)\to{\rm TL}(x)),\{1,\{{\it hu}\}, \emptyset, {\it bel}, 0.5, {\it a\ posteriori}\})$.  The graph $G_1$ is formed from $G_0$ by adding the vertices TL, S and the edge (link) $({\rm S},{\rm TL})$.

The inputs of the next six formulas in the foregoing list are all handled similarly, each comprising an event of Type 4.  This leads to a language $L_7$ generated by symbol set $\sigma_7=\sigma_0\cup\{{\rm TL}, {\rm S}, {\rm E}, {\rm H}, {\rm CS}, {\rm P}, {\rm AI}\}$.  Belief set $B_7$ will consist of the seven indicated labeled formulas with indexes (time stamps) 1 through 7.  Graph $G_7$ will consist of the vertices TL, S, E, H, CS, P, AI and the seven is-a-subclass-of links shown in Figure 4.1.

Input of $(\forall x)\lnot({\rm E}(x)\land{\rm H}(x))$ is an event ot Type 5.  This gives $L_8=L_7$, $B_8$ is formed from $B_7$ by adding the given input formula together with a label having index 8, and $G_8$ is formed from $G_7$ by adding the are-disjoint edge (E, H).

Consider input of the formula ${\rm S}({\rm Doc1})$.  This is an event of Type 1.  This gives symbol set $\sigma_9=\sigma_8\cup\{{\rm Doc1}\}$, the belief set $B_9$ is formed from $B_8$ by adding the input formula with a label having index 9, and graph $G_9$ is obtained from $G_8$ by adding the is-an-element-of edge $({\rm Doc1},{\rm S})$. The algorithm for Event Type 1 then proceeds to apply Aristotelian Syllogism to the input formula and the formula $(\forall x)({\rm S}(x)\to{\rm TH}(x))$ that was input in step 1 (counting $B_0$ as step 0).  This derives the formula ${\rm TL}({\rm Doc1})$ and provides this formula to the controller.  This is an event of Type 3.  The effect of the algorithm for this event type is that $L_{10}=L_9$, $G_{10}=G_9$, and $B_{10}$ is formed from $B_9$ by adding the newly derived formula with a label having index 10.  In addition, the from-list in the label for the derived formula is set to $\{{\rm Aristotelian Syllogism}, 9, 1\}$, and the to-lists in the label for formulas with indexes 9 and 1 are set to $\{10\}$.  For brevity in the following, assume that similar updatings of from-lists and to-lists are performed as appropriate in accordance with the definitions in Section 2.1.  Note that a from-list will refer to at most one inference rule and set of premises, whereas a to-list may contain indexes of any number of derived conclusions.   

Consider input of the formula ${\rm E}({\rm Doc1})$.  This is an event of Type 1.  This gives $L_{11}=L_{10}$, $B_{11}$ is formed from $B_{10}$ by adding the input formula with a label having index 11, and $G_{11}$ is formed from $G_{10}$ by adding the edge $({\rm Doc1}, {\rm E})$.   The algorithm for Event Type 1 then proceeds to apply Aristotelian Syllogism to this formula and the formula $(\forall x)({\rm E}(x)\to{\rm TH}(x))$ that was input in step 2.  This derives the formula ${\rm TL}({\rm Doc1})$ and provides this formula to the controller.  This is an event of Type 3.  Since the formula ${\rm TL}({\rm Doc1})$ is already in the belief set, the rule that duplicates are forbidden is invoked and the algorithm for Event Type 3 is not invoked.

Consider input of the formula ${\rm AI}({\rm Doc2})$.  This is an event of Type 1.  This gives symbol set $\sigma_{12}=\sigma_{11}\cup\{{\rm Doc2}\}$, the belief set $B_{12}$ is formed from $B_{11}$ by adding the input formula with a label having index 12, and graph $G_{12}$ is obtained from $G_{11}$ by adding the is-an-element-of edge $({\rm Doc2},{\rm AI})$.  The algorithm for Event Type 1 then proceeds to apply Aristotelian Syllogism to this formula and the formula $(\forall x)({\rm AI}(x)\to{\rm CS}(x))$ that was input in step 7. This derives the formula ${\rm CS}({\rm Doc2})$ and provides this formula to the controller.  This is an event of Type 3.  The algorithm for Type 3 is invoked, giving $L_{13}=L_{12}$, $B_{13}$ is formed from $B_{12}$ by adding the derived formula with a label having index 13, and $G_{13}=G_{12}$. Then, continuing with the algorithm for Event Type 3, Aristotelian Syllogism is applied to this formula and the formula $(\forall x)({\rm CS}(x)\to{\rm S}(x))$ that was input in step 4. This derives the formula ${\rm S}({\rm Doc2})$ and provides this formula to the controller.  This is a recursive invocation of Event Type 3 leading to $L_{14}=L_{13}$, $B_{14}$ is formed from $B_{13}$ by adding the derived formula with a label having index 14, and $G_{14}=G_{13}$.  Then, continuing with the algorithm for Event Type 3, Aristotelian Syllogism is applied to this formula and the formula $(\forall x)({\rm S}(x)\to{\rm TL}(x))$ that was input in step 4 (as this is the next formula in the belief set to which the inference rule can be applied). This derives the formula ${\rm TL}({\rm Doc2})$ and provides this formula to the controller, which is another event of Type 3.  The algorithm for this event type yields $L_{15}=L_{14}$, $B_{15}$ is formed from $B_{14}$ by adding the derived formula with a label having index 15, and $G_{15}=G_{14}$. Since there are no opportunities to apply Aristotelian Syllogism with this formula, the recursion now backtracks to the first invocation of Event Type 3, since there is another opportunity to apply Aristotelian Syllogism at that point, this time to the derived formula ${\rm CS}({\rm Doc2})$ and the formula $(\forall x)({\rm CS}(x)\to{\rm E}(x))$ that was input in step 5.  The algorithm proceeds similarly with the foregoing, giving $L_{16}=L_{15}$, $B_{16}$ is formed from $B_{15}$ by adding the formula ${\rm E}({\rm Doc2})$  with a label having index 16, $G_{16}=G_{15}$, and then Aristotelian Syllogism is applied deriving ${\rm TL}({\rm Doc2})$, giving an event of Type 3 whose algorithm is not invoked because of the rule forbidding duplicates in the belief set. 

Consider input of the formula ${\rm P}({\rm Doc3})$.  This is an event of Type 1.  Similarly with the foregoing, this gives symbol set $\sigma_{17}=\sigma_{16}\cup\{{\rm Doc3}\}$, $B_{17}$ is formed from $B_{16}$ by adding the input formula with a label having index 17, and $G_{17}$ is formed from $G_{16}$ by adding the edge $({\rm Doc3}, {\rm P})$.  The algorithm for Event Type 1 then proceeds to apply Aristotelian Syllogism to this formula and the formula $(\forall x)({\rm P}(x)\to{\rm H}(x))$ that was input in step 6. This derives the formula ${\rm H}({\rm Doc3})$ and provides this formula to the controller.  This is an event of Type 3.  The algorithm for Type 3 is invoked, giving $L_{18}=L_{17}$, $B_{18}$ is formed from $B_{17}$ by adding the derived formula with a label having index 18, and $G_{18}=G_{17}$. Then, continuing with the algorithm for Event Type 3, Aristotelian Syllogism is applied to this formula and the formula $(\forall x)({\rm H}(x)\to{\rm TL}(x))$ that was input in step 3. This derives the formula ${\rm LT}({\rm Doc2})$ and provides this formula to the controller, which is another event of Type 3.  The algorithm for this event type yields $L_{19}=L_{18}$, $B_{19}$ is formed from $B_{18}$ by adding the derived formula with a label having index 19, and $G_{19}=G_{18}$. Since there are no opportunities to apply Aristotelian Syllogism with this formula, the algorithm terminates.   

This completes the construction of the taxonomy in Figure 4.1.  At this point the language $L_{19}$ is the one generated by symbol set $\sigma_{19}=\sigma_0\cup\{{\rm TL},{\rm  S}, {\rm E}, {\rm H}, {\rm CS}, {\rm P}, {\rm AI}, {\rm Doc1}, {\rm Doc2}, {\rm Doc3}\}$, the belief set $B_{19}$ consists of the labeled formulas described above listed in the order of their indexes 1 through 19, and $G_{19}$ consists of the nodes and edges shown in Figure 4.1.  Note that, at each time step $t$, the belief set $B_t$ contains all formulas of the form $\alpha(a)$ that are implicit in the graph $G_t$.

Now suppose the user inputs ${\rm CS}({\rm Doc3})$.  This is an event of Type 1.  Since both CS and Doc3 are in the current symbol set, this gives $L_{20}=L_{19}$, $B_{20}$ is formed from $B_{19}$ by adding the input formula with a label having index 20, and $G_{20}$ is formed from $G_{19}$ by adding the edge $({\rm Doc3}, {\rm CS})$.  The algorithm for Event Type 1 then proceeds to apply Aristotelian Syllogism to this formula and the formula $(\forall x)({\rm CS}(x)\to{\rm S}(x))$ that was input in step 4. This derives the formula ${\rm S}({\rm Doc3})$ and provides this formula to the controller.  This is an event of Type 3.  The algorithm for Type 3 is invoked, giving $L_{21}=L_{20}$, $B_{21}$ is formed from $B_{20}$ by adding the derived formula with a label having index 21, and $G_{21}=G_{20}$. Then, continuing with the algorithm for Event Type 3, Aristotelian Syllogism is applied to this formula and the formula $(\forall x)({\rm S}(x)\to{\rm TL}(x))$ that was input in step 1. This derives the formula ${\rm TL}({\rm Doc3})$ and provides this formula to the controller.  This is a recursive invocation of Event Type 3 leading to $L_{22}=L_{21}$, $B_{22}$ is formed from $B_{21}$ by adding the derived formula with a label having index 22, and $G_{22}=G_{21}$.  Since there are no further opportunities to apply Aristotelian Syllogism, the recursion backtracks and then continues with the next opportunity to apply Aristotelian syllogism on the previous invocation of Event Type 3, namely, it applies the rule to ${\rm CS}({\rm Doc3})$ and the formula $(\forall x)({\rm CS}(x)\to{\rm E}(x))$ which was input in step 5.  This derives the formula ${\rm E}({\rm Doc3})$ and provides this formula to the controller.  This is an event of Type 3.  The algorithm for Type 3 is invoked, giving $L_{23}=L_{22}$, $B_{23}$ is formed from $B_{22}$ by adding the derived formula with a label having index 23, and $G_{23}=G_{22}$. Then, continuing with the algorithm for Event Type 3, then moves on to the phase where it searches for formulas comprising a conflict.  (Note that this phase was also required in all the above occurrences of Event Type 3, but, for brevity, was not mentioned as there would have been no conflict to detect).  A conflict arises from the formula ${\rm E}({\rm Doc3})$ just derived, the formula ${\rm H}({\rm Doc3})$ entered in step 18, and the formula $(\forall x)\lnot({\rm E}(x)\land{\rm H}(x))$ input in step 8.  Then Conflict Detection is applied to infer $\bot$ and this is provided to the controller.  This is an event of Type 2, giving $L_{24}=L_{23}$, $B_{24}$ is formed from $B_{23}$ by adding the input formula with a label having index 24, and $G_{24}=G_{23}$.  Then Dialectical Belief revision is invoked.  To-lists of the premises in the application of Conflict Detection are explored, leading backwards through the derivations to the extralogical axioms $(\forall x)\lnot({\rm E}(x)\land{\rm H}(x))$, $(\forall x)({\rm CS}(x)\to{\rm E}(x))$, ${\rm CS}({\rm Doc3})$, $(\forall x)({\rm P}(x)\to{\rm H}(x))$, and ${\rm P}({\rm Doc3})$. 

Let us suppose that the user decides to resolve the conflict by changing the status of ${\rm CS}({\rm Doc3})$ to {\it disbel}.  Then, starting with this formula, to-lists are followed and the statuses of the derived formulas are changed to {\it disbel}.  In order, these are ${\rm S}({\rm Doc3})$, ${\rm TL}({\rm Doc3})$, ${\rm E}({\rm Doc3})$, and $\bot$.  This effectively restores the language, belief set, and graph to the state they were in before the formula ${\rm CS}({\rm Doc3})$ was input.  More exactly, $L_{25}=L_{24}$ (which $=L_{19}$), $B_{25}$ is obtained from $B_{24}$ by making the indicated status changes, so that the active formulas are just the ones in $B_{19}$, and $G_{25}=G_{24}$ (which $=G_{19}$).

The user could have alternatively chosen to disbelieve one of the other discovered extralogical axioms.  For example, suppose that $(\forall x)({\rm CS}(x)\to{\rm E}(x))$ is chosen.  In step 16 this formula was used together with ${\rm CS}({\rm Doc2})$ to infer ${\rm E}({\rm Doc2})$, and in step 23 this formula was used together with ${\rm CS}({\rm Doc3})$ to infer ${\rm E}({\rm Doc3})$ (so the to-list for this formula is $\{16, 23\}$).  Accordingly, Dialectical Belief Revision proceeds to change the status of both of these derived formulas to {\it disbel}, and then, as above, changing the status of the derived $\bot$.  Moreover, the is-a-subset-of link from CS to E is removed from the graph.  Thus, in this case, $L_{25}=L_{24}$, $B_{25}$ is obtained from $B_{24}$ by making the indicated status changes, and $G_{25}$ is obtained from $G_{24}$ by removing the indicated edge.  This result is different than the one above, but the belief set is nonetheless consistent.          

As can be seen, the algorithms provided here are completely finitary, and they are sufficiently precisely defined that it is clear they can be implemented on a conventional computer.  To reach this point was the primary goal of the present research.  Moreover, the given example helps to show that the algorithms have the desired effect, namely, (i) the graph is maintained in the proper form as a DAG without redundant links, (ii) all formulas of the form $\alpha(a)$ implicit in the graph are derived and entered into the belief set, and (iii) whenever a triggered algorithm terminates, the belief set is consistent (even though it might not have been during the algorithm's processing).  That this will always be the case is established in the next two subsections.  

\subsection {Saliency}

That the DMA controller produces all relevant salient information as prescribed above can be summarized as a pair of theorems.  \medskip

{\bf Theorem 4.1.} The foregoing algorithms serve to maintain the graph, ignoring are-disjoint links, as a directed acyclic graph without redundant links. \medskip

Thus leaf nodes will be either documents or empty classification categories; if a document is linked to some category, it is not directly linked to any ancestors of that category; and no category is directly linked to any of its ancestors other than its parents.  This makes the graph useful for visualization and browsing.  

In addition, the algorithms ensure that the belief set will contain explicit representation of all and only the document-category classifications implicit in the graph.  This amounts to the following.  \medskip

{\bf Theorem 4.2.} After any process initiated by a user input terminates, the resulting belief set will contain a formula of the form $\alpha(a)$ iff the formula is derivable from the formulas corresponding to links in the graph.  \medskip 

\subsection {Correctness}

From an intuitive standpoint there are good reasons to believe that the given algorithms should serve to preserve the consistency of the belief set for a DMA.  These reasons are (i) it seems evident that the only way it is possible for an inconsistency to arise is if there occurs a formula of the form $(\forall x)\lnot(\alpha(x)\land\beta(x))$ together formulas of the form $\alpha(a)$ and $\beta(a)$, (ii) the foregoing Theorem 6 guarantees that this can be determined simply by scanning the (finite) contents of the belief set, and (iii) the presence of such formulas automatically triggers an application of Dialectical Belief Revision, in which the offending conflict is removed.  By definition, however, consistency requires that there be no formula $P$ such that both $P$ and $\lnot P$ are formally derivable in the first-order theory that has the active members of the belief set its extralogical axioms, and this is not explicitly guaranteed by these reasons.  Moreover, there evidently is no straightforward proof-theoretic argument that will guarantee this.  These considerations motivate the following model-theoretic argument. \medskip  

{\bf Theorem 4.3.} For any derivation path in the DMA, the belief set that results at the conclusion of a process initiated by a user input will be consistent.  \medskip

\section {Example 2: Multiple Inheritance with Exceptions}

The main objective of \cite{schwartz97} was to show how a DRS framework could be used to formulate reasoning about property inheritance with exceptions, where the underlying logic was a probabilistic `logic of qualified syllogisms'.  This work was inspired in part by the frame-based systems due to Minsky \cite{minsky75} and constitutes an alternative formulation of the underlying logic (e.g., as discussed by Hayes \cite{hayes80}).
   
What was missing in \cite{schwartz97} was the notion of a controller.  There a reasoning system was presented and shown to provide intuitively plausible solutions to numerous `puzzles' that had previously appeared in the literature on nonmonotonic reasoning (e.g., Opus the Penguin \cite{touretsky84}, Nixon Diamond \cite{touretsky87}, and Clyde the Elephant \cite{touretsky87}).  But there was nothing to guide the reasoning processes---no means for providing a sense of purpose for the reasoning agent.  The present work fills this gap by adding a controller similar to the one described for the Document Management Assistant.  Moreover, it deals with a simpler system based on first-order logic and remands further exploitation of the logic of qualified syllogisms to a later work.  The kind of DRS developed in this section will be termed a {\it multiple inheritance system} (MIS).

For this application the language $\cal L$  given in Section 3.1 is expanded by including some {\it typed predicate symbols}, namely, some unary predicate symbols ${\bf A}^{(k)}_1,{\bf A}^{(k)}_2,\ldots$  representing {\it kinds} of things (any objects), and some unary predicate symbols ${\bf A}^{(p)}_1,{\bf A}^{(p)}_2,\ldots$ representing {\it properties} of things.  The superscripts $k$ and $p$ are applied also to generic denotations.  Thus an expression of the form $(\forall x)(\alpha^{(k)}(x)\to\beta^{(p)}(x))$ represents the proposition that all $\alpha$s have property $\beta$.  These new predicate symbols are used here purely as syntactical items for purposes of defining an extralogical `specificity principle' and some associated extralogical graphical structures and algorithms.  Semantically they are treated exactly the same as other predicate symbols.
  
A {\it multiple-inheritance hierarchy} $H$ will be a directed graph consisting of a set of {\it nodes} together with a set of {\it links} represented as ordered pairs of nodes.  Nodes may be either {\it object} nodes, {\it kind} nodes, or {\it property} nodes.  A link of the form (object node, kind node) will be an {\it object-kind\/} link, one of the form (kind node, kind node) will be a {\it subkind-kind\/} link, and one of the form (kind node, property node) will be a {\it has-property\/} link. There will be no other types of links.  Object nodes will be labeled with (represent) individual constant symbols, kind nodes will be labeled with (represent) kind-type unary predicate symbols, and property nodes will be labeled with (represent) property-type unary predicate symbols or negations of such symbols.  In addition, each property type predicate symbol with bear a numerical subscript, called an {\it occurrence index}, indicating an occurrence of that symbol in a given hierarchy $H$.  These indexes are used to distinguish different occurrences of the same property-type symbol in $H$.   An object-kind link between an individual constant symbol $a$ and a  predicate symbol $\alpha^{(k)}$ will represent the formula $\alpha^{(k)}(a)$, a subkind-kind link between  a predicate symbol $\alpha^{(k)}$ and a predicate symbol $\beta^{(k)}$ will represent the formula $(\forall x)(\alpha^{(k)}(x)\to\beta^{(k)}(x))$, and a has-property link between a predicate symbol  $\alpha^{(k)}$ and a predicate symbol $\beta^{(p)}_1$ will represent the formula $(\forall x)(\alpha^{(k)}(x)\to\beta^{(p)}_1(x))$.  

Given such an $H$, there is defined on the object nodes and the  kind nodes a {\it specificity relation} $>_s$ (read `more specific than') according to:  (i) if $({\rm node}_1,{\rm node}_2)$ is either an object-kind link or a kind-kind link, then ${\rm node}_1>_s{\rm node}_2$, and (ii) if ${\rm node}_1>_s{\rm node}_2$ and ${\rm node}_2>_s{\rm node}_3$, then ${\rm node}_1>_s{\rm node}_3$.  We shall also have a dual {\it generality relation} $>_g$ (read `more general than') defined by ${\rm node}_1>_g{\rm node}_2$ iff ${\rm node}_1<_s{\rm node}_2$.  It follows that object nodes are maximally specific and minimally general.  It also follows that $H$ may have any number of maximally general nodes, and in fact that it need not be connected.  A maximally general node is a {\it root} node.  A {\it path} in a hierarchy $H$ (not to be confused with the path in a path logic) will be a sequence ${\rm node}_1,\ldots,{\rm node}_n$ wherein, ${\rm node}_1$ is a root node and, for each $i=1,\ldots,n-2$, the pair $({\rm node}_{i+1},{\rm node}_i)$ is a subkind-kind link, and, the pair $({\rm node}_n,{\rm node}_{n-1})$ is either a subkind-kind link or an object-kind link. Note that property nodes do not participate in paths as here defined.  

Similarly as with the graphs used in the Document Management Assistant, it is desired to organize a multiple inheritance hierarchy as a directed acyclic graph (DAG) without redundant links with respect to the object-kind and subkind-kind links (i.e., here ignoring has-property links), where, as before, by a redundant link is meant a direct link from some node to an ancestor of that node other than the node's immediate ancestors (i.e., other than its parents).  More exactly, two distinct paths will form a {\it redundant pair} if they have some node in common beyond the first place where they differ.  This means that they comprise two distinct paths to the common node(s).  A path will be simply {\it redundant} (or {\it redundant in} $H$) if it is a member of a redundant pair.  As before, a path contains a {\it loop} if it has more than one occurrence of the same node.  Provisions are made in the following to ensure that hierarchies with loops or redundant paths are not allowed.  As is customary, the hierarchies will be drawn with the upward direction being from more specific to less (less general to more), so that roots appear at the top and objects appear at the bottom.  Kind-property links will extend horizontally from their associated kind nodes.

In terms of the above specificity relation on $H$, we can assign an {\it address} to each object and kind node in the following manner.  Let the addresses of the root nodes, in any order, be $(1),(2),(3),\ldots$.  Then for the node with address (1), say, let the next most specific nodes in any order have the addresses $(1,1),(1,2),(1,3),\ldots$; let the nodes next most specific to the one with address $(1,1)$ have addresses $(1,1,1),(1,1,2),(1,1,3),\ldots$; and so on.  Thus an address indicates the node's position in the hierarchy relative to some root node.  Inasmuch as an object or kind node may be more specific than several different root nodes, the same node may have more than one such address.  Note that the successive initial segments of an address are the addresses of the nodes appearing in the path from the related root node to the node having that initial segment as its address.  Let $>$ denote the usual lexicographic order on addresses.  We shall apply $>$ also to the nodes having those addresses.  It is easily verified that ${\rm node}_1>{\rm node}_2$ iff ${\rm node}_1>_s{\rm node}_2$.  For object and kind nodes, we shall use the term {\it specificity rank} (or just {\it rank}) synonymously with `address'. 

Since, as mentioned, it is possible for any given object or kind node to have more than one address, it thus can have more than one rank.  Two nodes are comparable with respect to the specificity relation $>_s$, however, only if they appear on the same path, i.e., only if one node is an ancestor of the other, in which case only the rank each has acquired due to its being on that path will apply.  Thus, if two nodes are comparable with respect to their ranks by the relation $>$, there is no ambiguity regarding the ranks being compared.

Having thus defined specificity ranks for object and kind nodes, let us agree that each property node inherits the rank of the kind node to which it is linked.  Thus for property nodes the rank is not an address.

\begin{figure}[htp]
\centerline{\includegraphics[height=1.65in]{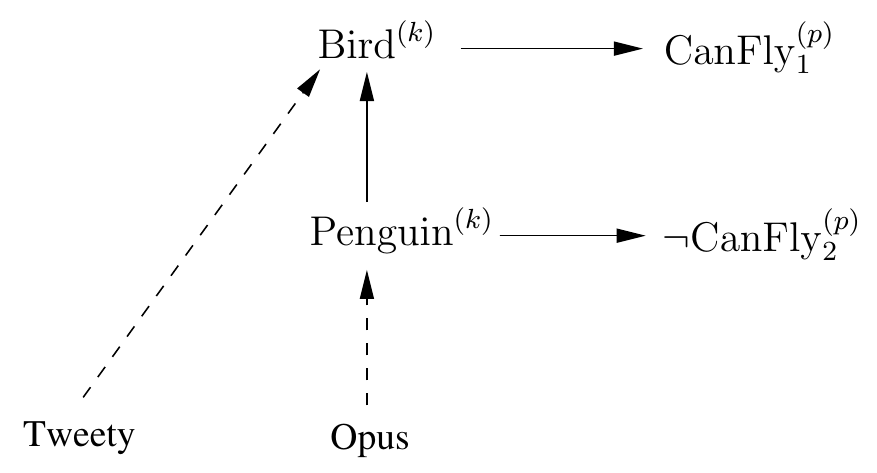}}
\medskip
\centerline{Figure 5.1. Tweety the Bird and Opus the Penguin as an MIS.}
\end{figure}  

An example of such a hierarchy is shown in Figure 5.1.  Here `Tweety' and `Opus' may be taken as names for the individual constants ${\bf a}_1$ and ${\bf a}_2$, and `${\rm Bird}^{(k)}$', `${\rm Penguin}^{(k)}$', and `${\rm CanFly}^{(p)}$' can be taken as names, respectively, for the unary predicate symbols ${\bf A}^{(k)}_1$,  ${\bf A}^{(k)}_2$, and ${\bf A}^{(p)}_1$.  [Note: The superscripts are retained on the names only to visually identify the types of the predicate symbols, and could be dropped without altering the meanings.] The links represent the formulas \medskip

\indent\indent $(\forall x)({\rm Penguin}^{(k)}(x)\to{\rm Bird}^{(k)}(x))$

\indent\indent $(\forall x)({\rm Bird}^{(k)}(x)\to{\rm CanFly}^{(p)}_1(x))$

\indent\indent $(\forall x)({\rm Penguin}^{(k)}(x)\to\lnot{\rm CanFly}^{(p)}_2(x))$

\indent\indent ${\rm Bird}^{(k)}({\rm Tweety})$

\indent\indent ${\rm Penquin}^{(k)}({\rm Opus})$  \medskip

\noindent The subscripts 1 and 2 on the predicate symbol ${\rm CanFly}^{(p)}$ in the graph distinguish the different occurrences of this symbol in the graph, and the same subcripts on the symbol occurrences in the formulas serve to correlate these with their occurrences in the graph.  Note that these are just separate occurrences of the same symbol, however, and therefore have identical semantic interpretations.  Formally,  ${\rm CanFly}^{(p)}_1$ and ${\rm CanFly}^{(p)}_2$ can be taken as standing for ${\bf A}^{(p)}_{1_1}$ and ${\bf A}^{(p)}_{1_2}$ with the lower subscripts being regarded as extralogical notations indicating different occurrences of ${\bf A}^{(p)}_1$.

This figure reveals the rationale for the present notion of multiple-inheritance hierarchy.  The intended interpretation of the graph is that element nodes and kind nodes inherit the properties of their parents, with the exception that more specific property nodes take priorty and block inheritances from those that are less specific.  Let us refer to this as the {\it specificity principle}. In accordance with this principle, in Figure 5.1 Tweety inherits the property CanFly from Bird, but Opus does not inherit this property because the inheritance is blocked by the more specific information that Opus is a Penguin and Penguins cannot fly.  

\begin{figure}[htp]
\centerline{\includegraphics[height=2in]{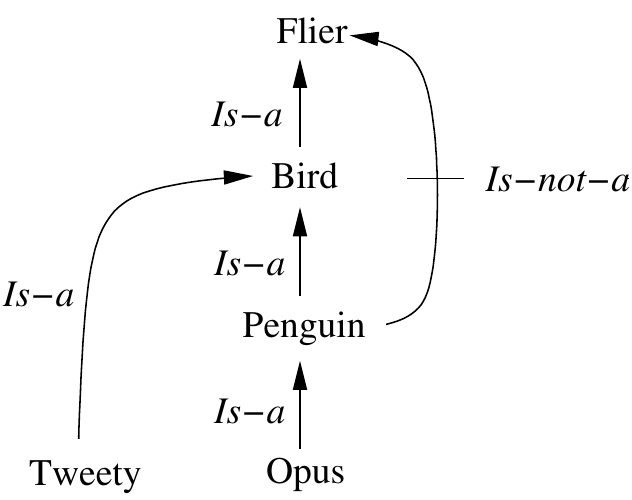}}
\medskip
\centerline{Figure 5.2. Tweety the Bird and Opus the Penguin, original version.}
\end{figure}  

Figure 5.1 constitutes a rethinking of the well-known example of Opus the penguin depicted in Figure 5.2 (adapted from \cite{touretsky84}).  The latter is problematic in that, by one reasoning path one can conclude that Opus is a flier, and by another reasoning path that he is not. This same contradiction is implicit in the formulas introduced above, since if one were to apply the axioms and rules of first-order logic as given in Section 3, one could derive both ${\rm CanFly}^{(p)}({\rm Opus})$ and $\lnot{\rm CanFly}^{(p)}({\rm Opus})$, in which case the system would be inconsistent.  

\subsection{Formal Specification of an Arbitrary MIS}

We are now in a position to define the desired kind of DRS.  For the path logic, let the language be the one described above, obtained from the $\cal L$ of Section 3.1 by adjoining the additional unary kind-type and property-type predicate symbols, let the axiom schemas be $(S 1)$ through $(S 6)$, and let the inference rules be $(R 1)$ through $(R 5)$ together with Aristotelian Syllogism and Contradiction Detection.  Thus the path logic component of an MIS is essentially identical to that of a DMA, the only exception being that we here use Contradiction Detection instead of Conflict Detection.  Also, the subsumption rule is omitted here, since as with the DMA as described in Section 4, this rule is not required.  It follows that all the results of Section 3 will apply.  Similarly with the DMA, derivation paths will consist of triples $(L_t,B_t,H_t)$, where these components respectively are the (sub)language (of $\cal L$), belief set, and multiple inheritance hierarchy at time $t$.  In accordance with Section 2.1, let $L_0$ be the minimal sublanguage of $\cal L$ consisting of all formulas that can be built up from the atomic formula $\bot$, and let $B_0=\emptyset$.  In addition, let $H_0=\emptyset$.

The controller for an MIS is defined analogously with that for the DMA, but differs in several important respects. In particular, the MIS controller is designed to enforce the above specificity principle. It may also be noted that this rendition of the MIS does not allow one to express that two kind categories are disjoint (although this could be included if desired), so that the source of contradictory formulas that was inherent in the DMA is not present here.  Nonetheless contradictions can arise in an MIS that has inherently contradictory root nodes in its multiple inheritance hierarchy.  An example of this (the famous Nixon Diamond \cite{touretsky87}) will be discussed. The purpose of the MIS controller will be essentially the same as for the DMA controller, namely (i) to derive and enter into the belief set all object classification implicit in the multiple inheritance hierarchy, i.e., all formulas of the form $\alpha^{(k)}(a)$ that can be derived from formulas describing the hierarchy (while observing the specificity principle), and (ii) to ensure that the belief set remains consistent. Item (i) thus defines what will be considered the {\it salient information} for an MIS.  Also similarly with the DMA, the MIS controller is intended to maintain the multiple inheritance hierarchy as a DAG without redundant paths, only here with respect to just the object and kind nodes.  Formulas that can be input by the users may have one of the forms (i) $\alpha^{(k)}(a)$, (ii) $(\forall x)(\alpha^{(k)}(x)\to\beta^{(k)}(x))$, (iii)  $(\forall x)(\alpha^{(k)}(x)\to\beta^{(p)}(x))$, and (iv)  $(\forall x)(\alpha^{(k)}(x)\to\lnot\beta^{(p)}(x))$.   Again it will be agreed that the epistemic entrenchment value for all input formulas is $0.5$.

We may now define some algorithms that are to be executed in response to each type of user input.  There will be eight types of events.  Event Types 1, 6, 7 and 8 correspond to user inputs, and the others occur as the result of rule applications.  As before, in all such events it is assumed that, if the formula provided to the controller already exists and is active in the current belief set, its input is immediately rejected.  In each event, assume that the most recent entry into the derivation path is $(L_t,B_t,H_t)$. \medskip

{\bf Event Type 1:}  A formula of the form $\alpha^{(k)}(a)$ is provided to the controller by a human user.  If either $\alpha^{(k)}$ or $a$ is not in the symbol set for $L_t$, form $L_{t+1}$ by adding the missing ones to the symbol set for $L_t$; otherwise set $L_{t+1}=L_t$.  Form $B_{t+1}$ from $B_t$ by adding the labeled formula $(\alpha^{(k)}(a), \{t+1,\{{\it hu}\}, \emptyset, {\it bel}, 0.5, {\it a\ posteriori}\})$. If there are no nodes representing either $a$ or $\alpha^{(k)}$ in $H_t$, form $H_{t+1}$ by adding such nodes together with an object-kind link from the $a$ node to the $\alpha^{(k)}$ node.  If one of $a$ and $\alpha^{(k)}$ is represented by a node in $H_t$ but the other is not, form $H_{t+1}$ by adding a node representing the one that is missing together with an object-kind link from the $a$ node to the $\alpha^{(k)}$ node.  Note that if a kind node is being added, this will become a root node.  If both $a$ and $\alpha^{(k)}$ are represented by nodes in $G_t$, form $G_{t+1}$ by adding an object-kind link from the $a$ node to the $\alpha^{(k)}$ node, unless this would create a redundant path.

Search $B_{t+1}$ for any active formulas having the form $(\forall x)(\alpha^{(k)}(x)\to\beta^{(k)}(x))$, where $\alpha$ is the predicate symbol of the input formula.  For each such formula, apply Aristotelian Syllogism to this and the formula $\alpha^{(k)}(a)$ to infer $\beta^{(k)}(a)$, and provide this formula to the controller. This is an event of Type 2.

Let $B_{t^*}$ be the most recent belief set, i.e., either it is $B_{t+1}$ or it is the belief set that has resulted from the processes associated with the indicated events of Type 2, if any occurred. If the input formula $\alpha^{(k)}(a)$ is still active, search $B_{t^*}$ for any active formulas having the form $(\forall x)(\alpha^{(k)}(x)\to\beta^{(p)}(x))$, where $\alpha$ is the predicate symbol of the input formula.  For each such formula, apply Aristotelian Syllogism to this and the formula $\alpha^{(k)}(a)$ to infer $\beta^{(p)}(a)$, and provide this formula to the controller. This is an event of Type 3.

Let $B_{t^*}$ be the most recent belief set, i.e., either it is $B_{t+1}$ or it is the belief set that has resulted from the processes associated with the indicated events of Type 2 and/or Type 3, if any occurred. If the input formula $\alpha^{(k)}(a)$ is still active, search $B_{t^*}$ for any active formulas having the form $(\forall x)(\alpha^{(k)}(x)\to\lnot\beta^{(p)}(x))$, where $\alpha$ is the predicate symbol of the input formula.  For each such formula, apply Aristotelian Syllogism to this and the formula $\alpha^{(k)}(a)$ to infer $\lnot\beta^{(p)}(a)$, and provide this formula to the controller. This is an event of Type 4. \medskip

{\bf Event Type 2:} A formula of the form $\alpha^{(k)}(a)$ is provided to the controller as a result of an inference rule application (Aristotelian Syllogism).  In this case, both $\alpha^{(k)}$ and $a$ are already in $L_t$, so let $L_{t+1}=L_t$.  Form $B_{t+1}$ from $B_t$ by (i) adding the labeled formula $(\alpha^{(k)}(a), \{t+1,F, \emptyset, {\it bel}, 0.5, {\it a\ posteriori}\})$, where the from-list $F$ contains the name of the inference rule that was used to infer $\alpha^{(k)}(a)$ (Aristotelian Syllogism), together with the indexes of the formulas that served as premises in the rule application, and (ii) updating the to-lists of all formulas that thus served as premises by including the index $t+1$. Let $H_{t+1}=H_t$.  Note that no modification of the hierarchy is warranted, since the membership of the object associated with $a$ in the category associated with $\alpha^{(k)}$ is already implicit in the hierarchy, so that a link between the respective nodes would form a redundant path.

Search $B_{t+1}$ for any active formulas having the form $(\forall x)(\alpha^{(k)}(x)\to\beta^{(k)}(x))$, where $\alpha^{(k)}$ is the predicate symbol of the above derived formula.  For each such formula, apply Aristotelian Syllogism to this and the formula $\alpha^{(k)}(a)$ to infer $\beta^{(k)}(a)$, and provide this formula to the controller. This is a recursive occurrence of an event of Type 2.

Let $B_{t^*}$ be the most recent belief set, i.e., either it is $B_{t+1}$ or it is the belief set that has resulted from the processes associated with the indicated events of Type 2, if any occurred. If the input formula $\alpha^{(k)}(a)$ is still active, search $B_{t^*}$ for any active formulas having the form $(\forall x)(\alpha^{(k)}(x)\to\beta^{(p)}(x))$, where $\alpha$ is the predicate symbol of the input formula.  For each such formula, apply Aristotelian Syllogism to this and the formula $\alpha^{(k)}(a)$ to infer $\beta^{(p)}(a)$, and provide this formula to the controller. This is an event of Type 3.

Let $B_{t^*}$ be the most recent belief set, i.e., either it is $B_{t+1}$ or it is the belief set that has resulted from the processes associated with the indicated events of Type 3 and/or Type 4, if any occurred. If the input formula $\alpha^{(k)}(a)$ is still active, search $B_{t^*}$ for any active formulas having the form $(\forall x)(\alpha^{(k)}(x)\to\lnot\beta^{(p)}(x))$, where $\alpha$ is the predicate symbol of the input formula.  For each such formula, apply Aristotelian Syllogism to this and the formula $\alpha^{(k)}(a)$ to infer $\lnot\beta^{(p)}(a)$, and provide this formula to the controller. This is an event of Type 4. \medskip

{\bf Event Type 3:} A formula of the form $\alpha^{(p)}(a)$ is provided to the controller as a result of an inference rule application (Aristotelian Syllogism).  Search the current hierarchy $H_t$ for the most specific occurrence (relative to the given occurrence of $\alpha^{(p)}(a)$) in a formula containing $\alpha^{(p)}$ (i.e., either it is an occurrence of $\alpha^{(p)}$ or of $\lnot\alpha^{(p)}$).  If this most specific occurrence is an occurrence of $\lnot\alpha^{(p)}$, let $L_{t+1}=L_t$, $B_{t+1}=B_t$, and $H_{t+1}=H_t$. This amounts to invoking the specificity principle to block an inheritance.  If this most specific occurrence is an occurrence of $\alpha^{(p)}(a)$, the formula being input already resides in the current belief set, and again one can let $L_{t+1}=L_t$, $B_{t+1}=B_t$, and $H_{t+1}=H_t$. Otherwise, do the following.  Let $L_{t+1}=L_t$.   Form $B_{t+1}$ from $B_t$ by (i) adding the labelled formula $(\alpha^{(p)}(a), \{t+1,F, \emptyset, {\it bel}, 0.5, {\it a\ posteriori}\})$, where the from-list $F$ contains the name of the inference rule that was used to infer $\alpha^{(p)}(a)$ (Aristotelian Syllogism), together with the indexes of the formulas that served as premises in the rule application, and (ii) updating the to-lists of all formulas that thus served as premises by including the index $t+1$. Let $H_{t+1}=H_t$. 

Search $B_{t+1}$ for an active occurrence of the formula $\lnot\alpha^{(p)}(a)$, where $\alpha^{(p)}(a)$ is the input formula, and, if found, apply Contradiction Detection to infer $\bot$ and provide this formula to the controller.  This is an event of Type 5.  If  $\alpha^{(p)}(a)$ is still active, repeat the foregoing.  Keep doing this until either $\alpha^{(p)}(a)$ becomes inactive (disbelieved) or no further occurrences of $\lnot\alpha^{(p)}(a)$ are found.  \medskip

{\bf Event Type 4:} A formula of the form $\lnot\alpha^{(p)}(a)$ is provided to the controller as a result of an inference rule application (Aristotelian Syllogism).  Search the current hierarchy $H_t$ for the most specific occurrence (relative to the given occurrence of $\lnot\alpha^{(p)}(a)$) of a formula containing $\alpha^{(p)}$ (i.e., either it is an occurrence of $\alpha^{(p)}$ or of $\lnot\alpha^{(p)}$ ).  If this most specific occurrence is an occurrence of $\alpha^{(p)}$, let $L_{t+1}=L_t$, $B_{t+1}=B_t$, and $H_{t+1}=H_t$. This amounts to invoking the specificity principle to block an inheritance.  If this most specific occurrence is an occurrence of $\lnot\alpha^{(p)}$, the formula being input already resides in the current belief set, and again one can let $L_{t+1}=L_t$, $B_{t+1}=B_t$, and $H_{t+1}=H_t$. Otherwise, do the following.  Let $L_{t+1}=L_t$.   Form $B_{t+1}$ from $B_t$ by (i) adding the labeled formula $(\lnot\alpha^{(p)}(a), \{t+1,F, \emptyset, {\it bel}, 0.5, {\it a\ posteriori}\})$, where the from-list $F$ contains the name of the inference rule that was used to infer $\lnot\alpha^{(p)}(a)$ (Aristotelian Syllogism), together with the indexes of the formulas that served as premises in the rule application, and (ii) updating the to-lists of all formulas that thus served as premises by including the index $t+1$. Let $H_{t+1}=H_t$. 

Search $B_{t+1}$ for an active occurrence of the formula $\alpha^{(p)}(a)$, where $\lnot\alpha^{(p)}(a)$ is the input formula, and, if found, apply Contradiction Detection to infer $\bot$ and provide this formula to the controller.  This is an event of Type 5.  If $\lnot\alpha^{(p)}(a)$ is still active, repeat the foregoing.  Keep doing this until either $\lnot\alpha^{(p)}(a)$ becomes inactive (disbelieved) or no further occurrences of $\alpha^{(p)}(a)$ are found.  \medskip

{\bf Event Type 5:} The formula $\bot$ is provided to the controller as the result of an application of Contradiction Detection.  Let $L_{t+1}=L_t$.  Form $B_{t+1}$ from $B_t$ by (i) adding the labeled formula $(\bot, \{t+1,F, \emptyset, {\it bel}, 0.5,$ ${\it a\ posteriori}\})$, where the from-list $F$ contains the name of the inference rule (Contradiction Detection) that was used to conclude this occurrence of $\bot$, together with the indexes of the formulas that served as premises in the rule application, and (ii) updating the to-lists of all formulas that thus served as premises by including the index $t+1$. Let $H_{t+1}=H_t$. 

Now invoke the Dialectical Belief Revision algorithm on $B_{t+1}$ as described in Section 2.2, starting with the formula $\bot$ just added to the belief set.  As a result of this process, some formulas in the current belief set will have their status changed from {\it bel} to {\it disbel}.  Let $B_{t+2}$ be the belief set obtained from $B_{t+1}$ by making these changes in the relevant formulas' labels.  Let $L_{t+2}=L_{t+1}$.  Obtain $H_{t+2}$ from $H_{t+1}$ by removing any elements representing formulas whose statuses have thus been changed to {\it disbel}.  Specifically, (i) if a formula of the form $\alpha^{(k)}(a)$ is disbelieved, remove the object-kind link connecting the node representing $a$ to the node representing $\alpha^{(k)}$, and remove the node representing $a$, unless it is connected to some node other than the one representing $\alpha^{(k)}$, and (ii) if a formula of the form $(\forall x)(\alpha^{(k)}(x)\to\beta^{(k)}(x))$ is disbelieved, remove the subkind-kind link connecting the node representing $\alpha^{(k)}$ to the node representing $\beta^{(k)}$, and remove the node representing $\alpha^{(k)}$, unless it is connected to some node other than the one representing $\beta^{(k)}$.  \medskip

{\bf Event Type 6:}  A formula of the form $(\forall x)(\alpha^{(k)}(x)\to\beta^{(k)}(x))$ is provided to the controller by a human user.  If both $\alpha^{(k)}$ and $\beta^{(k)}$ are already in $L_t$, begin by performing as many as possible of the following two actions.  First, explore all ancestors of $\alpha^{(k)}$ to see if they include $\beta^{(k)}$, and, if so, reject the input and inform the user that the input is disallowed inasmuch as it would create a redundant path in the subsumption hierarchy. Second, explore all ancestors of $\beta^{(k)}$ as expressed by formulas in $B_t$ to determine whether these include $\alpha^{(k)}$, and, if so, reject the input and inform the user that the input is disallowed inasmuch as it would create a loop in the subsumption hierarchy.   If the input is not rejected for either of these reasons, do the following.
  
If either $\alpha^{(k)}$ or $\beta^{(k)}$ is not in the symbol set for $L_t$, form $L_{t+1}$ by adding the ones that are missing, otherwise let $L_{t+1}=L_t$.  Form $B_{t+1}$ from $B_t$ by adding the labeled formula $((\forall x)(\alpha^{(k)}(x)\to\beta^{(k)}(x)),\{t+1,\{{\it hu}\}, \emptyset, {\it bel}, 0.5, {\it a\ posteriori}\})$. If there are no nodes representing either $\alpha^{(k)}$ or $\beta^{(k)}$ in $H_t$, form $H_{t+1}$ by adding such nodes together with a subkind-kind link from the $\alpha^{(k)}$ node to the $\beta^{(k)}$ node.  If one of $\alpha^{(k)}$ and $\beta^{(k)}$ is represented by a node in $H_t$ but the other is not, form $H_{t+1}$ by adding a node representing the one that is missing together with a subkind-kind link from the $\alpha^{(k)}$ node to the $\beta^{(k)}$ node. If both $\alpha^{(k)}$ and $\beta^{(k)}$ are represented by nodes in $H_t$, form $H_{t+1}$ by adding a subkind-kind link from the $\alpha^{(k)}$ node to the $\beta^{(k)}$ node.

Now search $B_{t+1}$ for any active formulas of the form $\alpha^{(k)}(a)$ where $\alpha^{(k)}$ is the predicate symbol in the input formula, and, for each such formula, apply Aristotelian Syllogism to infer $\beta^{(k)}(a)$, and provide this to the controller. This is an event of Type 2. \medskip

{\bf Event Type 7:}   A formula of the form $(\forall x)(\alpha^{(k)}(x)\to\beta^{(p)}(x))$ is provided to the controller by a human user.  If either $\alpha^{(k)}$ or $\beta^{(p)}$ is not in the symbol set for $L_t$, form $L_{t+1}$ by adding the ones that are missing, otherwise let $L_{t+1}=L_t$. Form $B_{t+1}$ from $B_t$ by adding the labeled formula $((\forall x)(\alpha^{(k)}(x)\to\beta^{(p)}(x)),\{t+1,\{{\it hu}\}, \emptyset, {\it bel}, 0.5, {\it a\ posteriori}\})$.  If the occurrence of $\beta^{(p)}$ in this formula is the $n$-th occurrence of this predicate symbol in the current derivation path, affix this occurrence with the extralogical subscript $n$ to serve as an occurrence index.  If there are no nodes representing either $\alpha^{(k)}$ or $\beta^{(p)}$ in $H_t$, form $H_{t+1}$ by adding such nodes (including the occurrence index on $\beta^{(p)}$) together with a has-property link from the $\alpha^{(k)}$ node to the $\beta^{(p)}$ node.  If one of $\alpha^{(k)}$ and $\beta^{(p)}$ is represented by a node in $H_t$ but the other is not, form $H_{t+1}$ by adding a node representing the one that is missing together with a has-property link from the $\alpha^{(k)}$ node to the $\beta^{(p)}$ node. If both $\alpha^{(k)}$ and $\beta^{(p)}$ are represented by nodes in $H_t$, form $H_{t+1}$ by adding a has-property link from the $\alpha^{(k)}$ node to the $\beta^{(p)}$ node.

Now search $B_{t+1}$ for any active formulas of the form $\alpha^{(k)}(a)$ where $\alpha^{(k)}$ is the predicate symbol in the input formula, and, for each such formula, apply Aristotelian Syllogism to infer $\beta^{(p)}(a)$, and provide this to the controller. This is an event of Type 3. \medskip

{\bf Event Type 8:}   A formula of the form $(\forall x)(\alpha^{(k)}(x)\to\lnot\beta^{(p)}(x))$ is provided to the controller by a human user.   If either $\alpha^{(k)}$ or $\beta^{(p)}$ is not in the symbol set for $L_t$, form $L_{t+1}$ by adding the ones that are missing, otherwise let $L_{t+1}=L_t$.  Form $B_{t+1}$ from $B_t$ by adding the labeled formula $((\forall x)(\alpha^{(k)}(x)\to\lnot\beta^{(p)}(x)),\{t+1,\{{\it hu}\}, \emptyset, {\it bel}, 0.5, {\it a\ posteriori}\})$.  If the occurrence of $\beta^{(p)}$ in this formula is the $n$-th occurrence of this predicate symbol in the current derivation path, affix this occurrence with the extralogical subscript $n$ to serve as an occurrence index.  If there are no nodes representing either $\alpha^{(k)}$ or $\lnot\beta^{(p)}$ in $H_t$, form $H_{t+1}$ by adding such nodes (including the occurrence index on $\beta^{(p)}$) together with a has-property link from the $\alpha^{(k)}$ node to the $\lnot\beta^{(p)}$ node.  If one of $\alpha^{(k)}$ and $\lnot\beta^{(p)}$ is represented by a node in $H_t$ but the other is not, form $H_{t+1}$ by adding a node representing the one that is missing together with a has-property link from the $\alpha^{(k)}$ node to the $\lnot\beta^{(p)}$ node. If both $\alpha^{(k)}$ and $\lnot\beta^{(p)}$ are represented by nodes in $H_t$, form $H_{t+1}$ by adding a has-property link from the $\alpha^{(k)}$ node to the $\lnot\beta^{(p)}$ node.

Now search $B_{t+1}$ for any active formulas of the form $\alpha^{(k)}(a)$ where $\alpha^{(k)}$ is the predicate symbol in the input formula, and, for each such formula, apply Aristotelian Syllogism to infer $\lnot\beta^{(p)}(a)$, and provide this to the controller. This is an event of Type 4. \medskip

We now turn to the task of establishing saliency and correctness for an arbitrary MIS.  This entails proving appropriate analogs of Proposition 14 and Theorems 5, 6, and 7.  \medskip

{\bf Proposition 5.1.} Every MIS is a normal DRS.  \medskip

\subsection {Illustration 1}

Some of the algorithms associated with the foregoing events can be illustrated by considering the inputs needed to create the inheritance hierarchy shown in Figure 5.1.  This focuses on the process of property inheritance with exceptions.  Let us abbreviate `Bird', `Penguin', and `CanFly', respectively, by `B', `P', and `CF'.   In accordance with the definition of derivation path in Section 2.1, the language $L_0$ will be the language generated by the logical symbols given in Section 3.1, i.e., by $\sigma_0=\{{\bf x}_1,{\bf x}_2,\ldots, \hbox{\rm `,'}, \hbox{\rm `('}, \hbox{\rm `)'}, \lnot, \lor, \forall, \bot\}$. This means that the only formula in $L_0$ is $\bot$.  Also in accordance with the Section 2.1, belief set $B_0=\emptyset$.  In accordance with the definition of an MIS, set $H_0=\emptyset$. 

Consider an input of the first formula in the foregoing list, namely,  $(\forall x)({\rm P}^{(k)}(x)\to{\rm B}^{(k)}(x))$. This is an event of Type 6.  The language $L_1$ is formed from $L_0$ by adding the symbols ${\rm P}^{(k)}$ and ${\rm B}^{(k)}$ (or, more exactly, the predicate letters ${\bf A}^{(k)}_2$ and  ${\bf A}^{(k)}_1$), i.e., $\sigma_1=\sigma_0\cup\{{\rm P}^{(k)}, {\rm B}^{(k)}\}$.  The belief set $B_1$ is formed from $B_0$ by adding the labeled formula $((\forall x)({\rm P}^{(k)}(x)\to{\rm B}^{(k)}(x)),\{1,\{{\it hu}\}, \emptyset, {\it bel}, 0.5,$ ${\it a\ posteriori}\})$.  The hierarchy $H_1$ is formed from $H_0$ by adding the nodes ${\rm P}^{(k)}$, ${\rm B}^{(k)}$ and the subkind-kind link $({\rm P}^{(k)},{\rm B}^{(k)})$.

Consider an input of the second formula in the list, namely,  $(\forall x)({\rm B}^{(k)}(x)\to{\rm CF}^{(p)}_1(x))$. This is an event of Type 7.  The language $L_2$ is formed from $L_1$ by adding the symbol ${\rm CF}^{(p)}$ (or, more exactly, the predicate letter ${\bf A}^{(p)}_1$), i.e., $\sigma_2=\sigma_1\cup\{{\rm CF}^{(p)}\}$.  The belief set $B_2$ is formed from $B_1$ by adding the labeled formula $((\forall x)({\rm B}^{(k)}(x)\to{\rm CF}^{(p)}_1(x)),\{2,\{{\it hu}\}, \emptyset, {\it bel}, 0.5,$ ${\it a\ posteriori}\})$, here including the extralogical occurrence index on ${\rm CF}^{(p)}$.  The hierarchy $H_2$ is formed from $H_1$ by adding the node ${\rm CF}^{(p)}_1$ and the has-property link $({\rm B}^{(k)},{\rm CF}^{(p)}_1)$ (again adding the extralogical subscript).

Consider an input of $(\forall x)({\rm P}^{(k)}(x)\to\lnot{\rm CF}^{(p)}_2(x))$. This is an event of Type 8.  The language $L_3$ is set equal to $L_2$.  The belief set $B_3$ is formed from $B_2$ by adding the labeled formula $((\forall x)({\rm P}^{(k)}(x)\to\lnot{\rm CF}^{(p)}_2(x)),\{3,\{{\it hu}\}, \emptyset, {\it bel}, 0.5,$ ${\it a\ posteriori}\})$, including the extralogical occurrence on ${\rm CF}^{(p)}$ as before.  The hierarchy $H_3$ is formed from $H_2$ by adding the node $\lnot{\rm CF}^{(p)}_2$ and the has-property link $({\rm B}^{(k)},\lnot{\rm CF}^{(p)}_2)$.

Consider an input of ${\rm B}^{(k)}({\rm Tweety})$. This is an event of Type 1.  The language $L_4$ is formed from $L_3$ by adding the symbol Tweety (or, more exactly, the individual constant ${\bf a}_1$).  The belief set $B_4$ is formed from $B_3$ by adding the labeled formula $({\rm B}^{(k)}({\rm Tweety}),\{4,\{{\it hu}\}, \emptyset, {\it bel}, 0.5,$ ${\it a\ posteriori}\})$.  The hierarchy $H_4$ is formed from $H_3$ by adding the node  Tweety and the object-kind link $({\rm Tweety}, {\rm B}^{(k)})$.  The algorithm for Event Type 1 then proceeds to apply Aristotelian Syllogism to  ${\rm B}^{(k)}({\rm Tweety})$ and the previously input formula $(\forall x)({\rm B}^{(k)}(x)\to{\rm CF}^{(p)}_1(x))$ to infer ${\rm CF}^{(p)}_1({\rm Tweety}))$ and provide this to the controller.  This is an event of Type 3.  The algorithm for Event Type 3 (i) sets $L_5=L_4$, (ii) forms $B_5$ from $B_4$ by adding the labeled formula $({\rm CF}^{(p)}_1({\rm Tweety}), \{5, F, \emptyset, {\it bel}, 0.5, {\it a\ posteriori}\})$, where the from-list $F=\{{\rm Aristotelian Syllogism}, 4, 2\}$, and changes the to-lists of formulas 4 and 2 from $\emptyset$ to $\{5\}$, and (iii) sets $H_5=H_4$.  

Consider an input of ${\rm P}^{(k)}({\rm Opus})$. This is an event of Type 1.  The language $L_6$ is formed from $L_5$ by adding the symbol Opus (or, more exactly, the individual constant ${\bf a}_2$).  The belief set $B_6$ is formed from $B_5$ by adding the labeled formula $({\rm P}^{(k)}({\rm Opus}),\{6,\{{\it hu}\}, \emptyset, {\it bel}, 0.5,$ ${\it a\ posteriori}\})$.  The hierarchy $H_6$ is formed from $H_5$ by adding the node  Opus and the object-kind link $({\rm Opus}, {\rm P}^{(k)})$.  The algorithm for Event Type 1 then proceeds to apply Aristotelian Syllogism to  ${\rm P}^{(k)}({\rm Opus})$ and the previously input formula $(\forall x)({\rm P}^{(k)}(x)\to{\rm B}^{(k)}(x))$ to infer ${\rm B}^{(k)}({\rm Opus}))$ and provide this to the controller.  This is an event of Type 2.  The algorithm for Event Type 2 (i) sets $L_7=L_6$, (ii) forms $B_7$ from $B_6$ by adding the labeled formula $({\rm B}^{(k)}({\rm Opus}), \{7, F, \emptyset, {\it bel}, 0.5, {\it a\ posteriori}\})$, where the from-list $F=\{{\rm Aristotelian Syllogism}, 6, 1\}$, and changes the to-lists of formulas 6 and 1 from $\emptyset$ to $\{7\}$, and (iii) sets $H_7=H_6$.  

The algorithm for Event Type 2 then continues by applying Aristotelian Syllogism to the inferred formula ${\rm B}^{(k)}({\rm Opus})$ and the previously input formula $(\forall x)({\rm B}^{(k)}(x)\to{\rm CF}^{(p)}_1(x))$ to infer ${\rm CF}^{(p)}_1({\rm Opus})$ and provides this formula to the controller. This is an event of Type 3. Since there is a more specific occurrence of CF in the formula $\lnot{\rm CF}^{(p)}_2(x)$, namely, the newly derived formula is not entered into the belief set, giving $L_8=L_7$, $B_8=B_7$, and $H_8=H_7$.

So far the process has invoked Event Type 1, which led to an invocation of Event Type 2, which led to an invocation of Event Type 3.  The Event Type 3 algorithm has now terminated, which effectively has also terminated the Event Type 2 algorithm, sending the flow of control back to where it left off in the Event Type 1 algorithm.  This algorithm then proceeds to apply Aristotelian Syllogism to the input formula ${\rm P}^{(k)}({\rm Opus})$ and the previously input formula $(\forall x)({\rm B}^{(k)}(x)\to\lnot{\rm CF}^{(p)}_2(x))$ to infer $\lnot{\rm CF}^{(p)}_2({\rm Opus})$ and provide this formula to the controller. This is an event of Type 4.  Because this formula contains the most specific occurrence of CF, the algorithm (i) sets $L_9=L_8$, (ii) forms $B_9$ from $B_8$ by adding the labeled formula $(\lnot{\rm CF}^{(p)}_2({\rm Opus}), \{9, F, \emptyset, {\it bel}, 0.5, {\it a\ posteriori}\})$, where the from-list $F=\{{\rm Aristotelian Syllogism}, 6, 3\}$, changes the to-list of formula 6 from $\{7\}$  to $\{7,9\}$ and to-list of formula 3 from $\emptyset$ to $\{9\}$, and (iii) sets $H_7=H_6$.   

Thus is is seen that, in this example, the algorithms serve to derive all salient information, i.e., all formulas of the forms $\alpha^{(k)(a)}$, $\alpha^{(p)(a)}$, and $\alpha^{(p)(a)}$ that are implicit in the graph, while at the same time correctly enforcing the specificity principle.

\subsection {Illustration 2}

This considers the application of contradiction detection.  The classic Nixon Diamond puzzle (cf. \cite{touretsky87}) is shown in Figure 5.3.  Here a contradiction arises because, by the reasoning portrayed on the left side, Nixon is a pacifist, whereas, by the reasoning portrayed on the right, he is not.  The resolution of this puzzle in the context of an MIS can be described in terms of the multiple inheritance hierarchy shown in Figure 5.4.    

\begin{figure}[htp]
\centerline{\includegraphics[height=2.15in]{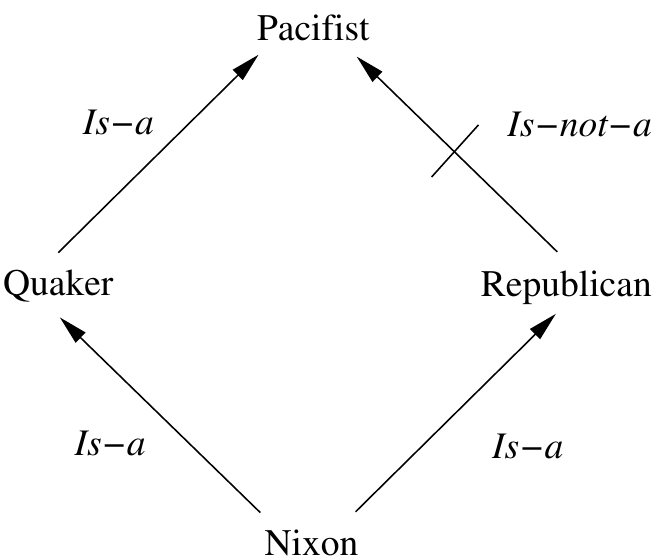}}
\medskip
\centerline{Figure 5.3. Nixon Diamond, original version.}
\end{figure}  

\begin{figure}[htp]
\centerline{\includegraphics[height=1.4in]{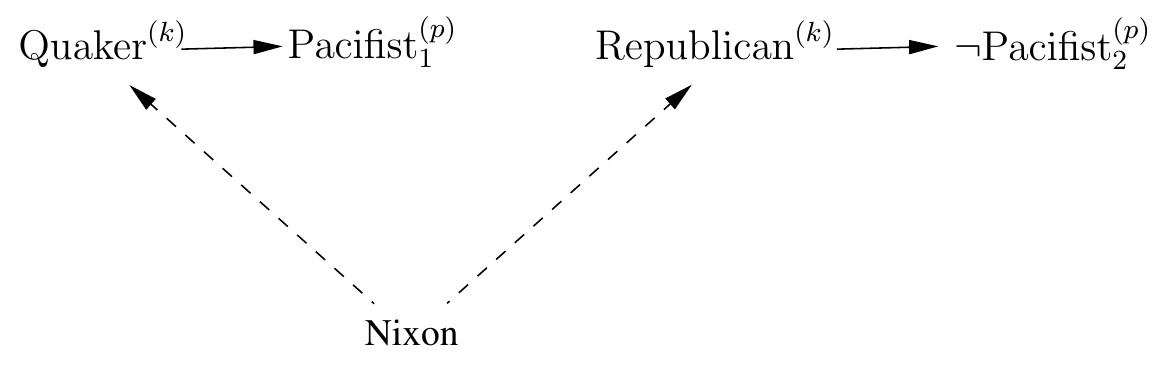}}
\medskip
\centerline{Figure 5.4. Nixon Diamond as an MIS.}
\end{figure}  

The links in Figure 5.4 represent the formulas \medskip

\indent\indent $(\forall x)({\rm Quaker}^{(k)}(x)\to{\rm Pacifist}^{(p)}_1(x))$

\indent\indent $(\forall x)({\rm Republican}^{(k)}(x)\to\lnot{\rm Pacifist}^{(p)}_2(x))$

\indent\indent ${\rm Quaker}^{(k)}({\rm Nixon})$

\indent\indent ${\rm Republican}^{(k)}({\rm Nixon})$ \medskip

The action of the algorithms may be traced similarly as in Illustration 1.  Let `Quaker',  `Republican' and `Pacifist' denote the predicate symbols ${\bf A}^{(k)}_1$, ${\bf A}^{(k)}_2$ and ${\bf A}^{(p)}_1$, and abbreviate these by `Q', `R' and `P'.  Let `Nixon' denote the individual constant ${\bf a}_1$. In accordance with the definition of derivation path in Section 2.1, the language $L_0$ will be the language generated by the logical symbols given in Section 3.1, i.e., by $\sigma_0=\{{\bf x}_1,{\bf x}_2,\ldots, \hbox{\rm `,'}, \hbox{\rm `('}, \hbox{`)'}, \lnot, \lor, \forall, \bot\}$. This means that the only formula in $L_0$ is $\bot$.  Also in accordance with the Section 2.1, belief set $B_0=\emptyset$.  In accordance with the definition of an MIS, set $H_0=\emptyset$. 

Consider an input of $(\forall x)({\rm Q}^{(k)}(x)\to{\rm P}^{(p)}_1(x))$.  This is an event of Type 7.  The language $L_1$ is formed from $L_0$ by adding the symbols ${\rm Q}^{(k)}$ and ${\rm P}^{(p)}$ (or, more exactly, the predicate letters ${\bf A}^{(k)}_1$ and ${\bf A}^{(p)}_1$), i.e., $\sigma_1=\sigma_0\cup\{{\rm Q}^{(k)},{\rm P}^{(p)}\}$.  The belief set $B_1$ is formed from $B_0$ by adding the labeled formula $((\forall x)({\rm Q}^{(k)}(x)\to{\rm P}^{(p)}_1(x)),\{1,\{{\it hu}\}, \emptyset, {\it bel}, 0.5,$ ${\it a\ posteriori}\})$, here including the extralogical occurrence index on ${\rm P}^{(p)}$.  The hierarchy $H_1$ is formed from $H_0$ by adding the nodes ${\rm Q}^{(k)}_1$ and ${\rm P}^{(p)}_1$ and the has-property link $({\rm Q}^{(k)},{\rm P}^{(p)}_1)$.

Consider an input of $(\forall x)({\rm R}^{(k)}(x)\to\lnot{\rm P}^{(p)}_1(x))$.  This is an event of Type 8.  The language $L_2$ is formed from $L_1$ by adding the symbol ${\rm R}^{(k)}$ (or, more exactly, the predicate letter ${\bf A}^{(k)}_2$), i.e., $\sigma_1=\sigma_0\cup\{{\rm R}^{(k)}\}$.  The belief set $B_2$ is formed from $B_1$ by adding the labeled formula $((\forall x)({\rm R}^{(k)}(x)\to\lnot{\rm P}^{(p)}_2(x)),\{2,\{{\it hu}\}, \emptyset, {\it bel}, 0.5,$ ${\it a\ posteriori}\})$, here again including the extralogical occurrence index on ${\rm P}^{(p)}$.  The hierarchy $H_2$ is formed from $H_1$ by adding the node $\lnot{\rm P}^{(p)}_2$ and the has-property link $({\rm R}^{(k)},{\rm P}^{(p)}_2)$.

Consider an input of ${\rm Q}^{(k)}({\rm Nixon})$. This is an event of Type 1.  The language $L_3$ is formed from $L_2$ by adding the symbol Nixon (or, more exactly, the individual constant ${\bf a}_1$).  The belief set $B_3$ is formed from $B_2$ by adding the labeled formula $({\rm Q}^{(k)}({\rm Nixon}),\{3,\{{\it hu}\}, \emptyset, {\it bel}, 0.5,$ ${\it a\ posteriori}\})$.  The hierarchy $H_3$ is formed from $H_2$ by adding the node Nixon and the object-kind link $({\rm Nixon}, {\rm Q}^{(k)})$.  The algorithm for Event Type 1 then proceeds to apply Aristotelian Syllogism to  ${\rm Q}^{(k)}({\rm Nixon})$ and the previously input formula $(\forall x)({\rm Q}^{(k)}(x)\to{\rm P}^{(p)}_1(x))$ to infer ${\rm P}^{(p)}_1({\rm Nixon})$ and provide this to the controller.  This is an event of Type 3.  The algorithm for Event Type 3 (i) sets $L_4=L_3$, (ii) forms $B_4$ from $B_3$ by adding the labeled formula $({\rm P}^{(p)}_1({\rm Nixon}), \{4, F, \emptyset, {\it bel}, 0.5, {\it a\ posteriori}\})$, where the from-list $F=\{{\rm Aristotelian Syllogism}, 3, 1\}$, and changes the to-lists of formulas 3 and 1 from $\emptyset$ to $\{4\}$, and (iii) sets $H_4=H_3$.  The algorithm for Event Type 3 then proceeds to scan the current belief set $B_4$ for any active occurrences of $\lnot{\rm P}^{(p)}({\rm Nixon})$ (ignoring the extralogical occurrence indexes) and doesn't find any.

Consider an input of ${\rm R}^{(k)}({\rm Nixon})$. This is an event of Type 1.  This sets $L_5=L_4$ and $H_5=H_4$. The belief set $B_5$ is formed from $B_4$ by adding the labeled formula $({\rm R}^{(k)}({\rm Nixon}),\{5,\{{\it hu}\}, \emptyset, {\it bel}, 0.5,$ ${\it a\ posteriori}\})$.  The algorithm for Event Type 1 then proceeds to apply Aristotelian Syllogism to  ${\rm R}^{(k)}({\rm Nixon})$ and the previously input formula $(\forall x)({\rm R}^{(k)}(x)\to\lnot{\rm P}^{(p)}_1(x))$ to infer $\lnot{\rm P}^{(p)}_2({\rm Nixon})$ and provide this to the controller.  This is an event of Type 4.  The algorithm for Event Type 4 (i) sets $L_6=L_5$, (ii) forms $B_6$ from $B_5$ by adding the labeled formula $(\lnot{\rm P}^{(p)}_2({\rm Nixon}), \{6, F, \emptyset, {\it bel}, 0.5, {\it a\ pos}$- ${\it teriori}\})$, where the from-list $F=\{{\rm Aristotelian Syllogism}, 5, 2\}$, and changes the to-lists of formulas 5 and 2 from $\emptyset$ to $\{6\}$, and (iii) sets $H_6=H_5$.   The algorithm for Event Type 4 then proceeds to scan the current belief set $B_6$ for any active occurrences of ${\rm P}^{(p)}({\rm Nixon})$ and finds the formula with index 4.  This triggers an application of Contradiction Detection, which derives $\bot$ and provides this to the controller. This is an event of Type 5.   This sets $L_7=L_6$ and $H_7=H_6$.  $B_7$ is formed from $B_6$ by adding the labeled formula $(\bot, \{7, F, \emptyset, {\it bel}, 0.5,$ ${\it a\ posteriori}\})$, where the from-list $F=\{{\rm Contradiction Detection}, 6, 4\}$, and changes the to-lists of formulas 6 and 4 from $\emptyset$ to $\{7\}$.  

Then Dialectical Belief Revision is invoked.  All the formulas that were input by the user are candidates for belief change.  Suppose that the formula with index 2, namely, $(\forall x)({\rm R}^{(k)}(x)\to\lnot{\rm P}^{(p)}_2(x))$, is chosen.  Then the procedure forward chains through to lists, starting with this formula, and changes to {\it disbel} the status first of formula 6, $\lnot{\rm P}^{(p)}_2({\rm Nixon})$, and then of formula 7, $\bot$. This results in $L_8=L_7$, $B_8$ is the belief set obtained from $B_7$ by making the three indicated belief status changes,  and $H_8$ is formed from $H_7$ by removing the links $({\rm Nixon}, {\rm R}^{(k)})$ and $({\rm R}^{(k)}, \lnot{\rm P}^{(p)}_2)$, leaving only the left side of the hierarchy in Figure 5.4.    

Further well-known puzzles that can be resolved similarly within an MIS are the others discussed in \cite{schwartz97}, namely, Bosco the Blue Whale \cite{stein92}, Suzie the Platypus\cite{stein92}, Clyde the Royal Elephant \cite{touretsky87}, and Expanded Nixon Diamond  \cite{touretsky87},

\subsection {Saliency}

That an MIS controller produces all relevant salient information as prescribed above can be summarized as a pair of theorems.  \medskip

{\bf Theorem 5.1.} The foregoing algorithms serve to maintain the hierarchy with respect to the object and kind nodes as a directed acyclic graph without redundant links. \medskip

Thus leaf nodes will be either object nodes or kind nodes that have no assigned objects.  If an object is linked to some kind node, it is not directly linked to any ancestors of that kind node; and no kind node is directly linked to any of its ancestors other than its parents.  This makes the hierarchy useful for visualization and browsing.  

In addition, the algorithms ensure that the belief set will contain explicit representation of all and only the object-kind classifications implicit in the hierarchy.  This amounts to the following.  \medskip

{\bf Theorem 5.2.} After any process initiated by a user input terminates, the resulting belief set will contain a formula of the form $\alpha^{(k)}(a)$ or $\alpha^{(p)}(a)$ or $\lnot\alpha^{(p)}(a)$  iff the formula is derivable from the formulas corresponding to links in the inheritance hierarchy, observing the specificity principle.  \medskip 

 \subsection {Correctness}

It is desired to establish the analog of Theorem 7 for the MIS DRS.  Here the concern is with maintaining consistency with respect to all formulas except those expressing that the elements of a kind have some property, since inconsistencies among these are allowed.  \medskip

{\bf Theorem 5.3.} For any derivation path in an MIS, the belief set that results at the conclusion of a process initiated by a user input will be consistent with respect to the formulas of the forms $\alpha^{(k)}(a)$, $(\forall x)(\alpha^{(k)}(x)\to\beta^{(p)}(x))$, and $\alpha^{(p)}(a)$.  \medskip

\section{Concluding Remarks}

This work has introduced a computational framework for nonmonotonic belief revision. The treatment has employed classical first-order predicate calculus as the underlying reasoning system, but the DRS framework can accommodate virtually any well-defined logic.  In each of the examples given, an application-specific controller was provided and proven to be adequate and correct for the intended application.  These proofs required that the logic have a well-defined semantics and be sound with respect to that semantics.  While it is difficult to speculate about further applications, it seems likely that such typically will be necessary and sufficient.  In any case, semantic completeness of the logic does not seem to be required.

There are several issues regarding this work that remain to be explored.  First, the formula labels considered herein have included a place for an epistemic entrenchment value, but these not been employed in the present examples, although it certainly would be possible to do so.  These could be used to remove the human from the loop when undergoing Dialectical Belief Revision, by using such values to automatically choose what extralogical axioms to disbelieve.  At issue, however, is how to determine exactly what should be the epistemic entrenchment value for any given formula.  No systematic methods for doing so currently exist.

Second, computational complexity of the controller algorithms is another issue.  Clearly, this must be studied for each different controller; there can be no general statements regarding this.  For the examples considered in this paper, however, the complexity seems manageable.      
     
Third, the previous work \cite{schwartz97} developed a `logic for qualified syllogisms' and applied this to resolve the same nonmonotonic puzzles discussed in the foregoing.  That work did not consider a controller, however, and it is an open question how a suitable one might be devised.  Doing so would require further developing the logic and providing a soundness result.

Last, the present work has provided all the precise mathematical details necessary for a software implementation.  Developing such an implementation would be a substantial undertaking, however, and before doing so it would be warranted to demonstrate that the DRS can be employed in practical real-world applications.   The present work has provided toy applications to illustrate the core ideas, but the prospect for substantive real-world applications has yet to be explored.    

\appendixhead{SCHWARTZ}

\bibliographystyle{plain}
\bibliography{p-bibfile}

\received{July 2013}{?}{?}

\elecappendix

\bigskip 

This appendix contains the proofs of all proposition and theorems.  \medskip

{\bf Proposition 3.1.} For any DRS, Dialectical Belief Revision is normal.  \medskip

{\bf Proof.} Suppose that $B_{t'}$ is obtained from $B_t$ by an application of Dialectical Belief Revision.  This involves changing the status of some formulas in $B_t$ from {\it bel} to {\it disbel}.  The issue is whether this might create `orphan' formulas whose status remain {\it bel} even though some formulas used in their derivations are now changed to {\it disbel}, i.e., so that their derivations are no longer valid, and they are no longer theorems.  Inspection of the Dialectical Belief Revision algorithm as defined in Section 2.2 makes it evident that this situation cannot occur.  In this process, the first formulas whose statuses are changed to {\it disbel} are extralogical axioms, and the process proceeds by following to-lists starting with these formulas and changing to {\it disbel} the status of all (and only) the formulas whose derivations depended on these.  Thus no such `orphan' formulas can arise.  $\square$ \medskip

{\bf Proposition 3.2.}  In a normal DRS, for each theory $T_t$ determined by a pair $(L_t,B_t)$ in a derivation path for the DRS, the active formulas in $B_t$ will be theorems of $T_t$.  \medskip 

{\bf Proof.} This is established by induction on the length $t$ of derivation paths.  For the base step with $t=0$, $B_t=\emptyset$ by definition of derivation path.  So all formulas in $B_t$ are theorems of $T_t$ by default.  For the induction step with $t\ge 0$, the induction hypothesis is that the formulas in $B_t$ are theorems of $T_t$, and it is required to show that, given this hypothesis, the formulas in $B_{t+1}$ must be theorems of $T_{t+1}$.  

Consider the five ways of forming $B_{t+1}$ from $B_t$ prescribed in the definition of derivation path.  Item 1: A logical axiom is added.  Logical axioms are theorems of $T_{t+1}$, by definition of theorem. Item 2: A formula is added as the derived conclusion of an inference rule application, where the premises of the application are in $B_t$ and are therefore theorems by the induction hypothesis.  Then the derived formula is a theorem of $T_{t+1}$, by definition of theorem. Item 3: An extralogical axiom is added.  An extralogical axiom of $T_{t+1}$ is a theorem of $T_{t+1}$, by definition of theorem.  Item 4: This also is addition of an extralogical axiom; so the argument for Item 3 applies.  Item 5: $B_{t+1}$ is obtained from $B_t$ by an application of belief revision algorithm.  Then the active formulas in $B_{t+1}$ are theorems of $T_{t+1}$, by definition of normal DRS.  This completes the induction step and concludes the proof by induction on $t$.  $\square$ \medskip

{\bf Proposition 3.3.} Let $T_t$ be the theory determined by an entry $(L_t,B_t)$ in the derivation path for a normal DRS, let $T$ be the theory with language $L_t$ and no extralogical axioms, and let $\Gamma$ be the set of active formulas in $B_t$.  Then, for any formula $P$ of $L_t$, $T_t\vdash P$ iff $T(\Gamma)\vdash P$.  \medskip

{\bf Proof.} Given the assumptions, suppose that $T_t\vdash P$.  By definition of $T_t$, the extralogical axioms of $T_t$ comprise a subset of $\Gamma$, so it follows by the definition of theorem that $T(\Gamma)\vdash P$. \medskip 

{\bf Theorem 3.1.} ({\it Soundness Theorem for PC\/}) For any theory $T$, if $T_{\rm PC}\vdash P$, then $P$ is a tautology.  \medskip

{\bf Proof.} This is Proposition 2.14 in \cite{hamilton88} and Proposition 1.11 in \cite{mendelson87}. The theorem is established by induction on the length of proofs $P_1,\ldots,P_n$ in $T_{\rm PC}$.  The base step, for $n=1$, amounts to verifying that all instances of axioms schemas $(S 1)$, $(S 2)$, and $(S 3)$ are tautologies.  The induction step, for $n>1$, amounts to showing that Modus Ponens preserves tautologousness, i.e., if $P$ and $P\to Q$ are tautologies, then $Q$ is a tautology. $\square$ \medskip

{\bf Theorem 3.2.} ({\it Adequacy Theorem for PC\/}) For any theory $T$, if $P$ is a tautology, then $T_{\rm PC}\vdash P$. \medskip

{\bf Proof.} This is Proposition 2.23 in \cite{hamilton88} and Proposition 1.13 in \cite{mendelson87}. $\square$ \medskip

These theorems can be used to show that axiom schema $(S4)$ serves merely as a defining axiom for $\bot$ and does not enable proving any additional formulas not involving $\bot$. Let $T_{{\rm PC(}\bot{\rm )}}\vdash P$ indicate that $P$ has a proof in $T$ using only $(S 1)$, $(S 2)$, $(S 3)$, and $(S 4)$, Schema Instantiation 1, and Modus Ponens.  \medskip

{\bf Proposition 3.4.} If $P$ does not contain any occurrences of $\bot$ and $T_{{\rm PC(}\bot{\rm )}}\vdash P$, then $T_{\rm PC}\vdash P$. \medskip

{\bf Proof.} This can be established by showing that the proof of Theorem 3.1 can be modified to show that, if $T_{{\rm PC(}\bot{\rm )}}\vdash P$, then $P$ is a tautology.  For then, if $P$ does not contain any occurrences of $\bot$, we have $T_{\rm PC}\vdash P$, by Theorem 3.2. The necessary modification of Theorem 1 amounts to augmenting the base step by showing that all instances of axiom schema $(S 4)$ are tautologies. $\square$ \medskip

{\bf Proposition 3.5.}  The above seven inference rules are valid in any theory $T$. \medskip

{\bf Proof.} Hypothetical Syllogism is Corollary 2.10 of \cite{hamilton88} and Corollary 1.9.(a) of \cite{mendelson87}. In the following, $T\vdash$ is shortened to $\vdash$.  

Aristotelian Syllogism can be derived as follows: \medskip 

\begin{tabular*}{6.2in}{ll@{\extracolsep\fill}l}
$\vdash (\forall x)(P\to Q)$ & (assumption) & (1) \\
\noalign{\medskip}
$\vdash P(a/x)$ & (assumption) & (2) \\
\noalign{\medskip}
$\vdash (\forall x)(P\to Q)\to(P\to Q)(a/x)$ & (Schema Instantiation 2) & (3) \\
\noalign{\medskip}
$\vdash (P\to Q)(a/x)$ & (1, 3, and Modus Ponens) & (4) \\
\noalign{\medskip}
$\vdash P(a/x)\to Q(a/x)$ & (4 and definition of notation $(a/x)$) & (5) \\
\noalign{\medskip}
$\vdash Q(a/x)$ & (2, 5, and Modus Ponens) & {}
\end{tabular*} \medskip

A derivation of Subsumption is: \medskip

\begin{tabular*}{6.2in}{ll@{\extracolsep\fill}l}
$\vdash (\forall x)(\alpha(x)\to\beta(x))$ & (assumption) & (1) \\
\noalign{\medskip}
$\vdash (\forall x)(\beta(x)\to\gamma(x))$ & (assumption) & (2) \\
\noalign{\medskip}
$\vdash (\forall x)(\alpha(x)\to\beta(x))\to(\alpha(x)\to\beta(x))$ & (Schema Instantiation 2 with $x$ for $t$) & (3) \\
\noalign{\medskip}
$\vdash (\forall x)(\beta(x)\to\gamma(x))\to(\beta(x)\to\gamma(x))$ & (Schema Instantiation 2 with $x$ for $t$) & (4) \\
\noalign{\medskip}
$\vdash \alpha(x)\to\beta(x)$ & (1, 3, and Modus Ponens) & (5) \\
\noalign{\medskip}
$\vdash \beta(x)\to\gamma(x)$ & (2, 4, and Modus Ponens) & (6) \\
\noalign{\medskip}
$\vdash \alpha(x)\to\gamma(x)$ & (4, 6, and Hypothetical Syllogism) & (7) \\
\noalign{\medskip}
$\vdash (\forall x)(\alpha(x)\to\gamma(x))$ & (7 and Generalization) & {} 
\end{tabular*} \medskip

And-Introduction can be established by: \medskip

\begin{tabular*}{6.2in}{ll@{\extracolsep\fill}l}
$\vdash P$ & (assumption) & (1) \\
\noalign{\medskip}
$\vdash Q$ & (assumption) & (2) \\
\noalign{\medskip}
$\vdash P\to (Q\to P\land Q)$ & (Exercise 1.46(h) of \cite{mendelson87}) & (3) \\ 
\noalign{\medskip}
$\vdash Q\to P\land Q$ & (1, 3, and Modus Ponens) & (4) \\
\noalign{\medskip}
$\vdash P\land Q$ & (2, 4, and Modus Ponens) & {}
\end{tabular*} \medskip

And-Elimination can be derived as follows: \medskip

\begin{tabular*}{6.2in}{ll@{\extracolsep\fill}l}
$\vdash P\land Q$ & (assumption) & (1) \\
\noalign{\medskip}
$\vdash \lnot(P\to\lnot Q)$ & (1 and definition of $\land$) & (2) \\
\noalign{\medskip}
$\vdash (\lnot P\to(P\to\lnot Q))\to(\lnot(P\to\lnot Q)\to\lnot\lnot P)$ & (Lemma 1.10.e of \cite{mendelson87}) & (3) \\
\noalign{\medskip}
$\vdash \lnot P\to(P\to\lnot Q)$ & (Lemma 1.10.c of \cite{mendelson87}) & (4) \\
\noalign{\medskip}
$\vdash \lnot(P\to\lnot Q)\to\lnot\lnot P$ & (3, 4, and Modus Ponens) & (5) \\
\noalign{\medskip}
$\vdash \lnot\lnot P\to P$ & (Lemma 1.10.a of \cite{mendelson87}) & (6) \\
\noalign{\medskip}
$\vdash \lnot(P\to\lnot Q)\to P$ & (5, 6, and Hypothetical Syllogism) & (7) \\
\noalign{\medskip}
$\vdash P$ & (2, 7, and Modus Ponens) & (8) \\
\noalign{\medskip}
$\vdash (\lnot Q\to(P\to\lnot Q))\to(\lnot(P\to\lnot Q)\to\lnot\lnot Q)$ & (Lemma 1.10.e of \cite{mendelson87}) & (9) \\
\noalign{\medskip}
$\vdash (\lnot Q\to(P\to\lnot Q))$ & ($(S 1)$ and Schema Instantiation 1) & (10) \\
\noalign{\medskip}
$\vdash \lnot(P\to\lnot Q)\to\lnot\lnot Q$ & (9, 10, and Modus Ponens) & (11) \\
\noalign{\medskip}
$\vdash \lnot\lnot Q\to Q$ & (Lemma 1.10.a of \cite{mendelson87}) & (12) \\
\noalign{\medskip}
$\vdash \lnot(P\to\lnot Q)\to Q$ & (11, 12, and Hypothetical Syllogism) & (13) \\
\noalign{\medskip}
$\vdash Q$ & (2, 13, and Modus Ponens) & {} 
\end{tabular*} \medskip

\noindent Items 8 and 13 together give the And-Elimination rule. \medskip

A derivation of Conflict Detection is: \medskip

\begin{tabular*}{6.2in}{ll@{\extracolsep\fill}l}
$\vdash (\forall x)\lnot(P\land Q)$ & (assumption) & (1) \\
\noalign{\medskip} 
$\vdash P(a/x)$ & (assumption) & (2) \\
\noalign{\medskip}
$\vdash Q(a/x)$ & (assumption) & (3) \\
\noalign{\medskip}
$\vdash (\forall x)\lnot(P\land Q)\to(\lnot(P\land Q))(a/x)$ & (Schema Instantiation 2) & (4) \\
\noalign{\medskip}
$\vdash (\lnot(P\land Q))(a/x)$ & (1, 4, and Modus Ponens) & (5) \\
\noalign{\medskip}
$\vdash P(a/x)\land Q(a/x)$ & (2, 3, and And-Introduction) & (6) \\
\noalign{\medskip}
$\vdash (P\land Q)(a/x)$ & (6 and definition of $(t/x)$) & (7) \\
\noalign{\medskip}
$\vdash (P\land Q)(a/x)\land(\lnot(P\land Q))(a/x)$ & (5, 7, and And-Introduction) & (8) \\
\noalign{\medskip}
$\vdash \lnot(P(a/x)\land Q(a/x))$ & (5 and definition of $(t/x)$) & (9) \\
\noalign{\medskip}
$\vdash \bot\leftrightarrow(P(a/x)\land Q(a/x))\land\lnot(P(a/x)\land Q(a/x))$ & ($(S 4)$ and Sch. Instantiation 1) & (10) \\
\noalign{\medskip}
$\vdash (\bot\to(P(a/x)\land Q(a/x))\land\lnot(P(a/x)\land Q(a/x)))\land $ & {} & {} \\
\kern 2em $((P(a/x)\land Q(a/x))\land\lnot(P(a/x)\land Q(a/x))\to\bot)$ & (10 and definition of $\leftrightarrow$) & (11) \\
\noalign{\medskip}
$\vdash (P(a/x)\land Q(a/x))\land\lnot(P(a/x)\land Q(a/x))\to\bot$ & (11 and And-Elimination) & (12) \\
\noalign{\medskip}
$\vdash (P\land Q)(a/x)\land(\lnot(P\land Q))(a/x)\to\bot$ & (12 and definition of $(t/x)$) & (13) \\
\noalign{\medskip}
$\vdash\bot$ & (8, 13, and Modus Ponens) & {}
\end{tabular*} \medskip

Contradiction Detection can be derived as follows: \medskip

\begin{tabular*}{6.2in}{ll@{\extracolsep\fill}l}
$\vdash P$ & (assumption) & (1) \\
\noalign{\medskip}
$\vdash \lnot P$ & (assumption) & (2) \\
\noalign{\medskip}
$\vdash P\land\lnot P$ & (1, 2, and And Introduction) & (3) \\ 
\noalign{\medskip}
$\vdash \bot\leftrightarrow P\land\lnot P$ & ($S 4$ and Schema Instantiation 1) & (4) \\
\noalign{\medskip}
$\vdash (\bot\to P\land\lnot P)\land (P\land\lnot P\to\bot)$ & (definition of $\leftrightarrow$) & (5) \\
\noalign{\medskip}
$\vdash P\land\lnot P\to\bot$ & (5 and And Elimination) & (6) \\
\noalign{\medskip}
$\vdash \bot$ & (2, 5, and Modus Ponens) & {} \\
\end{tabular*} \medskip

This completes the proof. $\square$\medskip

{\bf Proposition 3.6.} A theory $T$ is consistent iff there is a formula $P$ of $L_T$ such that $T{\not}\vdash P$. \medskip

{\bf Proof.} This is Proposition 2.18 in \cite{hamilton88}. $\square$ \medskip

{\bf Proposition 3.7.} For an entry $(L_t,B_t)$ in the derivation path of a normal DRS, $B_t$ is consistent if and only if the theory $T_t$ determined by $(L_t,B_t)$ is consistent. \medskip

{\bf Proof.} Let $T$ be the theory with language $L_t$ and no extralogical axioms, and let $\Gamma$ be the active formulas in $B_t$.  Note that $B_t$ is consistent if and only if $T(\Gamma)$ is consistent, by definition of consistent for $B_t$. By Proposition 3.3, $T(\Gamma)$ and $T_t$ have the same theorems, in which case $T(\Gamma)$ is consistent if and only if $T_t$ is consistent.  Thus $B_t$ is consistent if and only if  $T_t$ is consistent. $\square$ \medskip

{\bf Proposition 3.8.} Let $P$ be a formula of $L_T$ in which no individual variable other than $x$ occurs free, let $a$ be an individual constant of $L_T$, let $I$ be an interpretation of $L_T$, and let $d=I(a)$. Then $I\models P(a/x)$ iff $I\models P(\hat d/x)$. \medskip

{\bf Proof.} This is proven by induction on the length of formulas $P$ of $L_T$. Suppose $P$ is atomic having the form $\alpha(t_1,\ldots,t_n)$. If $x$ is not among $t_1,\ldots,t_n$, then, by definition of the notation $(t/x)$, both $P(a/x)$ and $P(\hat d/x)$ reduce to $P$, and the proposition is obviously true.  Suppose $x$ is among $t_1,\ldots,t_n$; say $x$ is $t_i$. Since $x$ is the only variable occurring free in $P$, the $t_k$ for $k\neq i$ are individual constants. Observe that $I(a)=I(\hat d)$. We have \medskip

\begin{tabular}{lll}
$I\models P(a/x)$ & iff\quad $v_I((\alpha(t_1,\ldots,t_{i-1},x,t_{i+1},\dots,t_n))(a/x))={\rm T}$ & (notation $\models$) \\
\noalign{\medskip}
{} & iff\quad $v_I(\alpha(t_1,\ldots,t_{i-1},a,t_{i+1},\dots,t_n))={\rm T}$ & (def. $(t/x)$) \\
\noalign{\medskip}
{} & iff\quad $v_I(\alpha)$ holds for $(I(t_1),\ldots,I(t_{i-1}),I(a),I(t_{i+1}),\dots,I(t_n))$ & (def. $v_I$) \\
\noalign{\medskip}
{} & iff\quad $v_I(\alpha)$ holds for $(I(t_1),\ldots,I(t_{i-1}),I(\hat d),I(t_{i+1}),\dots,I(t_n))$ & (above note) \\
\noalign{\medskip}
{} & iff\quad $v_I(\alpha(t_1,\ldots,t_{i-1},\hat d,t_{i+1},\dots,t_n))={\rm T}$ & (def. $v_I$) \\
\noalign{\medskip}
{} & iff\quad $v_I(\alpha(t_1,\ldots,t_{i-1},x,t_{i+1},\dots,t_n)(\hat d/x))={\rm T}$ & (def. $(t/x)$) \\
\noalign{\medskip}
{} & iff\quad $I\models P(\hat d/x)$ & (notation $\models$) 
\end{tabular} \medskip

Suppose $P$ is of the form $\lnot Q$. The induction hypothesis is that $I\models Q(a/x)$ iff $I\models Q(\hat d/x)$. By definition of the notation $\models$, this amounts to $v_I(Q(a/x))={\rm T}$ iff $v_I(Q(\hat d/x))={\rm T}$, which in turn amounts  to  $v_I(Q(a/x))={\rm F}$ iff $v_I(Q(\hat d/x))={\rm F}$. This gives \medskip

\begin{tabular}{lll}
$I\models P(a/x)$ & iff\quad $v_I(P(a/x))={\rm T}$ & (notation $\models$) \\
\noalign{\medskip}
{} & iff\quad $v_I(\lnot Q(a/x))={\rm T}$ & (assumption) \\
\noalign{\medskip}
{} & iff\quad $v_I(Q(a/x))={\rm F}$ & (def $v_I$) \\
\noalign{\medskip}
{} & iff\quad $v_I(Q(\hat d/x))={\rm F}$ & (induction hypothesis) \\
\noalign{\medskip}
{} & iff\quad $v_I(\lnot Q(\hat d/x))={\rm T}$ & (def $v_I$) \\
\noalign{\medskip}
{} & iff\quad $v_I(P(\hat d/x))={\rm T}$ & (assumption) \\
\noalign{\medskip}
{} & iff\quad $I\models P(\hat d/x)$ & (notation $\models$)
\end{tabular} \medskip

Suppose $P$ is of the form $Q\to R$. The induction hypothesis is that $I\models Q(a/x)$ iff $I\models Q(\hat d/x)$ and $I\models R(a/x)$ iff $I\models R(\hat d/x)$.  By definition of the notation $\models$, this amounts to $v_I(Q(a/x))={\rm T}$ iff $v_I(Q(\hat d/x))={\rm T}$ and $v_I(R(a/x))={\rm T}$ iff $v_I(R(\hat d/x))={\rm T}$, and the former in turn amounts to  $v_I(Q(a/x))={\rm F}$ iff $v_I(Q(\hat d/x))={\rm F}$. This gives \medskip

\begin{tabular}{lll}
$I\models P(a/x)$ & iff\quad $v_I(P(a/x))={\rm T}$ & (notation $\models$) \\
\noalign{\medskip}
{} & iff\quad $v_I((Q\to R)(a/x))={\rm T}$ & (assumption) \\
\noalign{\medskip}
{} & iff\quad $v_I(Q(a/x))={\rm F}$ or $v_I(R(a/x))={\rm T}$ & (def $v_I$) \\
\noalign{\medskip}
{} & iff\quad $v_I(Q(\hat d/x))={\rm F}$ or $v_I(R(\hat d/x))={\rm T}$ & (induction hypothesis) \\
\noalign{\medskip}
{} & iff\quad $v_I((Q\to R)(\hat d/x))={\rm T}$ & (def $v_I$) \\
\noalign{\medskip}
{} & iff\quad $v_I(P(\hat d/x))={\rm T}$ & (assumption) \\
\noalign{\medskip}
{} & iff\quad $I\models P(\hat d/x)$ & (notation $\models$)
\end{tabular} \medskip

This completes the proof by induction. $\square$\medskip 

{\bf Proposition 3.9.} For any language $L$, the logical axioms of $L$, i.e., all formulas derivable by means of the schema instantiation rules $(R 1)$ through $(R 3)$, are logically valid. \medskip

{\bf Proof.} Let $I$ be an interpretation of $L$. For Schema Instantiation 1, one can use a proof by contradiction for each axiom schema. To illustrate the method, consider an instance $P\to(Q\to P)$ of axiom schema $(S 1)$. Let $(P\to(Q\to P))(\hat d_1,\ldots,\hat d_n/x_1,\ldots,x_n)$ be an $I$-instance and suppose that $v_I((P\to(Q\to P))(\hat d_1,\ldots,\hat d_n/x_1,\ldots,x_n))={\bf F}$. Then, by the definition of $v_I$ (and definition of the notation for substitution), this means that $v_I(P(\hat d_1,\ldots,\hat d_n/x_1,\ldots,x_n))={\bf T}$ and $v_I((Q\to P)(\hat d_1,\ldots,\hat d_n/x_1,\ldots,x_n))={\bf F}$. But then, for the same reasons, the latter means that $v_I(Q(\hat d_1,\ldots,\hat d_n/x_1,\ldots,x_n))={\bf T}$ and $v_I(P(\hat d_1,\ldots,\hat d_n/x_1,\ldots,x_n))={\bf F}$, the latter of which is a contradiction. Thus the $I$-instance must be true in $I$. Since the $I$-instance was chosen arbitrarily, this means that the formula $P\to(Q\to P)$ is valid in $I$.  Since $I$ was chosen arbitrarily, the latter means that the formula is logically valid. Logical axioms derived as instances of axiom schemas $(S 2)$, $(S 3)$, and $(S 4)$ may be handled similarly.

For Schema Instantiation 2, consider a schema instance $(\forall x)P\to P(t/x)$ and an $I$-instance \medskip

\indent\indent $((\forall x)P\to P(t/x))(\hat d_1,\ldots,\hat d_n/x_1,\ldots,x_n)$.  \medskip

\noindent By definition of the notation $(t/x)$, the latter can be written \medskip

\indent\indent $((\forall x)P)(\hat d_1,\ldots,\hat d_n/x_1,\ldots,x_n)\to P(t/x)(\hat d_1,\ldots,\hat d_n/x_1,\ldots,x_n)$.  \medskip

\noindent Suppose that \medskip

\indent\indent$v_I(((\forall x)P)(\hat d_1,\ldots,\hat d_n/x_1,\ldots,x_n))={\bf T}$. \hfill (1)\kern 1em\medskip

\noindent It is desired to show that this implies \medskip

\indent\indent $v_I(P(t/x)(\hat d_1,\ldots,\hat d_n/x_1,\ldots,x_n))={\bf T}$ \hfill (2)\kern 1em\medskip

\noindent from which it will follow, by the definition of $v_I$, that \medskip

\indent\indent $v_i(((\forall x)P\to P(t/x))(\hat d_1,\ldots,\hat d_n/x_1,\ldots,x_n))={\bf T}$. \medskip

{\it Case 1\/}: $x$ is not free in $P$. Then, by the definition of $I$-instance, $x$ is not among $x_1,\ldots,x_n$, so that \medskip

\indent\indent $((\forall x)P)(\hat d_1,\ldots,\hat d_n/x_1,\ldots,x_n)$ \medskip

\noindent can be written \medskip

\indent\indent $(\forall x)(P(\hat d_1,\ldots,\hat d_n/x_1,\ldots,x_n))$ \medskip

\noindent Then it follows from (1) by the definition of $v_I$ that \medskip

\indent\indent $v_I(P(\hat d_1,\ldots,\hat d_n/x_1,\ldots,x_n)(\hat d/x))={\bf T}$, for all $d\in D_I$. \medskip 

\noindent Since $x$ does not occur in $P$, this reduces to \medskip

\indent\indent $v_I(P(\hat d_1,\ldots,\hat d_n/x_1,\ldots,x_n))={\bf T}$. \hfill (3)\kern 1em\medskip

\noindent Observe that, since $x$ does not occur in $P$, $P(t/x)$ is just $P$, by definition of the notation $(t/x)$. Thus (3) amounts to (2).

{\it Case 2\/}: $x$ is free in $P$. {\it Subcase 2.a\/}: $t$ is an individual constant, say $a$.  Then $x$ does not occur free in $(\forall x)P\to P(t/x)$, so by the definition of $I$-instance, $x$ is not among $x_1,\ldots,x_n$.  Thus, as above, \medskip

\indent\indent $((\forall x)P)(\hat d_1,\ldots,\hat d_n/x_1,\ldots,x_n)$ \medskip

\noindent can be written \medskip

\indent\indent $(\forall x)(P(\hat d_1,\ldots,\hat d_n/x_1,\ldots,x_n))$ \medskip

\noindent and it follows from (1) by the definition of $v_I$ that \medskip

\indent\indent $v_I(P(\hat d_1,\ldots,\hat d_n/x_1,\ldots,x_n)(\hat d/x))={\bf T}$, for all $d\in D_I$. \hfill (4)\kern 1em\medskip 

\noindent Note that, since $x$ is not among $x_1,\ldots,x_n$, $P(\hat d_1,\ldots,\hat d_n/x_1,\ldots,x_n)(\hat d/x)$ is the same as \break $P(\hat d/x)(\hat d_1,\ldots,\hat d_n/x_1,\ldots,x_n)$, and (4) becomes \medskip

\indent\indent $v_I(P(\hat d/x)(\hat d_1,\ldots,\hat d_n/x_1,\ldots,x_n))={\bf T}$, for all $d\in D_I$. \medskip

\noindent Thus, in particular, for $d=I(a)$, we have \medskip

\indent\indent $v_I(P(\hat d/x)(\hat d_1,\ldots,\hat d_n/x_1,\ldots,x_n))={\bf T}$ \medskip

\noindent from which it follows, by Proposition 8, that \medskip

\indent\indent $v_I(P(a/x)(\hat d_1,\ldots,\hat d_n/x_1,\ldots,x_n))={\bf T}$ \medskip

\noindent This is (2) with $a$ for $t$.

{\it Subcase 2.b\/}: $t$ is the individual variable $x$. Then $x$ occurs free in $(\forall x)P\to P(t/x)$, so by the definition of $I$-instance, $x$ is among $x_1,\ldots,x_n$.  Say $x$ is $x_i$. Then $(\forall x)P$ is $(\forall x_i)P$. Since $x_i$ is bound in $(\forall x_i)P$, one has by definition of the notation $(t/x)$ that \medskip

\indent\indent $((\forall x_i)P)(\hat d_1,\ldots,\hat d_n/x_1,\ldots,x_n)$ \medskip

\noindent can be written \medskip

\indent\indent $(\forall x_i)(P(\hat d_1,\ldots,\hat d_{i-1}, \hat d_{i+1},\ldots,\hat d_n/x_1,\ldots,x_{i-1},x_{i+1},\ldots,x_n))$ \medskip

\noindent and it follows from (1) by the definition of $v_I$ that \medskip

\indent\indent $v_I(P(\hat d_1,\ldots,\hat d_{i-1}, \hat d_{i+1},\ldots,\hat d_n/x_1,\ldots,x_{i-1},x_{i+1},\ldots,x_n)(\hat d/x_i))={\bf T}$, for all $d\in D_I$ \medskip

\noindent Then, in particular, \medskip

\indent\indent $v_I(P(\hat d_1,\ldots,\hat d_{i-1}, \hat d_{i+1},\ldots,\hat d_n/x_1,\ldots,x_{i-1},x_{i+1},\ldots,x_n)(\hat d_i/x_i))={\bf T}$ \medskip

\noindent which can be written as \medskip

\indent\indent $v_I(P(\hat d_1,\ldots,\hat d_n/x_1,\ldots,x_n))={\bf T}$ \hfill (5)\kern 1em\medskip

\noindent Observe that, since $t$ is $x$, $P(t/x)$ is just $P$, so (5) is (2) with $x$ for $t$.

{\it Subcase 2.c\/}: $t$ is an individual variable different from $x$, say $y$. Then $(\forall x)P\to P(t/x))$ is $(\forall x)P\to P(y/x)$, in which $y$ is free and $x$ is not. Consider an $I$-instance \medskip

\indent\indent $((\forall x)P\to P(y/x))(\hat d_1,\ldots,\hat d_n/x_1,\ldots,x_n)$ \medskip

\noindent Then, by the definition of $I$-instance, $y$ is among $x_1,\ldots,x_n$ and $x$ is not.  Suppose $y$ is $x_i$. Since $x$ is not among $x_1,\ldots,x_n$, \medskip

\indent\indent $((\forall x)P)(\hat d_1,\ldots,\hat d_n/x_1,\ldots,x_n)$ \medskip

\noindent can be written \medskip

\indent\indent $(\forall x)(P(\hat d_1,\ldots,\hat d_n/x_1,\ldots,x_n))$ \medskip

\noindent and it follows from (1) by the definition of $v_I$ that \medskip

\indent\indent $v_I(P(\hat d_1,\ldots,\hat d_n/x_1,\ldots,x_n)(\hat d/x))={\bf T}$, for all $d\in D_I$ \medskip

\noindent In particular, \medskip

\indent\indent $v_I(P(\hat d_1,\ldots,\hat d_n/x_1,\ldots,x_n)(\hat d_i/x))={\bf T}$ \medskip

\noindent Since $y$ is $x_i$, this is \medskip

\indent\indent $v_I(P(\hat d_1,\ldots,\hat d_n/x_1,\ldots,x_{i-1},y,x_{i+1},\ldots,x_n)(\hat d_i/x))={\bf T}$ \medskip

\noindent Since $(\hat d_i/x)$ is equivalent with $(y/x)(\hat d_i/y)$, this can be written \medskip

\indent\indent $v_I(P(y/x)(\hat d_1,\ldots,\hat d_n/x_1,\ldots,x_{i-1},y,x_{i+1},\ldots,x_n))={\bf T}$ \medskip

\noindent which, since $y$ is $x_i$, is \medskip

\indent\indent $v_I(P(y/x)(\hat d_1,\ldots,\hat d_n/x_1,\ldots,x_{i-1},x_i,x_{i+1},\ldots,x_n))={\bf T}$ \medskip

\noindent or simply \medskip

\indent\indent $v_I(P(y/x)(\hat d_1,\ldots,\hat d_n/x_1,\ldots,x_n))={\bf T}$ \medskip

\noindent This is (2) with $y$ for $t$.

For Schema Instatiation 3, consider a schema instance $(\forall x)(P\to Q)\to(P\to(\forall x)Q)$, where $x$ is not free in $P$, and an $I$-instance \medskip

\indent\indent $((\forall x)(P\to Q)\to(P\to(\forall x)Q))(\hat d_1,\ldots,\hat d_n/x_1,\ldots,x_n)$. \medskip

\noindent For the general case, assume that $x$ occurs free in $Q$. Then, by the definition of $I$-instance, $x$ is among $x_1,\ldots,x_n$.  Say $x$ is $x_i$. By the definition of the notation $(t/x)$, the above can be written \medskip

\indent\indent $((\forall x)(P\to Q))(\hat d_1,\ldots,\hat d_n/x_1,\ldots,x_n)\to(P\to(\forall x)Q)(\hat d_1,\ldots,\hat d_n/x_1,\ldots,x_n)$. \medskip 

\noindent For a proof by contradiction, suppose that \medskip

\indent\indent $v_I(((\forall x)(P\to Q)\to(P\to(\forall x)Q))(\hat d_1,\ldots,\hat d_n/x_1,\ldots,x_n))={\bf F}$. \medskip

\noindent Then from the above, by definition of $v_I$, we have \medskip

\indent\indent $v_I(((\forall x)(P\to Q))(\hat d_1,\ldots,\hat d_n/x_1,\ldots,x_n))={\bf T}$. \hfill (6)\kern 1em\medskip 

\noindent and \medskip

\indent\indent $v_I((P\to(\forall x)Q)(\hat d_1,\ldots,\hat d_n/x_1,\ldots,x_n))={\bf F}$. \hfill (7)\kern 1em\medskip 

\noindent Since $x$ is $x_i$, by definition of the notation $(t/x)$ and the fact that $x$ is bound in $(\forall x)(P\to Q)$, (6) can be written \medskip

\indent\indent $v_I(((\forall x)(P\to Q))(\hat d_1,\ldots,\hat d_{i-1},\hat d_{i+1},\dots,\hat d_n/x_1,\ldots,x_{i-1},x_{i+1},\ldots,x_n))={\bf T}$ \medskip 

\noindent which, since $x$ is not among $x_1,\ldots,x_{i-1},x_{i+1},\ldots,x_n$, can be written \medskip

\indent\indent $v_I((\forall x)((P\to Q)(\hat d_1,\ldots,\hat d_{i-1},\hat d_{i+1},\dots,\hat d_n/x_1,\ldots,x_{i-1},x_{i+1},\ldots,x_n)))={\bf T}$ \medskip

\noindent Then, by the definition of $v_I$, we have that, for all $d\in D_I$, \medskip

\indent\indent $v_I(((P\to Q)(\hat d_1,\ldots,\hat d_{i-1},\hat d_{i+1},\dots,\hat d_n/x_1,\ldots,x_{i-1},x_{i+1},\ldots,x_n))(\hat d/x))={\bf T}$. \hfill (8)\kern 1em\medskip

\noindent Since $x$ is $x_i$ and $x$ is not free in $P$, (7) can be written \medskip

\indent\indent $v_I((P\to(\forall x)Q)(\hat d_1,\ldots,\hat d_{i-1},\hat d_{i+1},\dots,\hat d_n/x_1,\ldots,x_{i-1},x_{i+1},\ldots,x_n))={\bf F}$ \medskip

\noindent which, by definition of the notation $(t/x)$, can be written \medskip

\indent\indent $v_I(P(\hat d_1,\ldots,\hat d_{i-1},\hat d_{i+1},\dots,\hat d_n/x_1,\ldots,x_{i-1},x_{i+1},\ldots,x_n)\to$

\indent\indent\qquad $((\forall x)Q)(\hat d_1,\ldots,\hat d_{i-1},\hat d_{i+1},\dots,\hat d_n/x_1,\ldots,x_{i-1},x_{i+1},\ldots,x_n))={\bf F}$ \medskip

\noindent which, since $x$ is not among $x_1,\ldots,x_{i-1},x_{i+1},\ldots,x_n$, can be written \medskip

\indent\indent $v_I(P(\hat d_1,\ldots,\hat d_{i-1},\hat d_{i+1},\dots,\hat d_n/x_1,\ldots,x_{i-1},x_{i+1},\ldots,x_n)\to$

\indent\indent\qquad $(\forall x)(Q(\hat d_1,\ldots,\hat d_{i-1},\hat d_{i+1},\dots,\hat d_n/x_1,\ldots,x_{i-1},x_{i+1},\ldots,x_n)))={\bf F}$ \medskip

\noindent By definition of $v_I$, this gives \medskip

\indent\indent $v_I(P(\hat d_1,\ldots,\hat d_{i-1},\hat d_{i+1},\dots,\hat d_n/x_1,\ldots,x_{i-1},x_{i+1},\ldots,x_n))={\bf T}$ \hfill (9)\kern 1em\medskip

\noindent and \medskip

\indent\indent $v_I((\forall x)(Q(\hat d_1,\ldots,\hat d_{i-1},\hat d_{i+1},\dots,\hat d_n/x_1,\ldots,x_{i-1},x_{i+1},\ldots,x_n)))={\bf F}$.\medskip

\noindent From the latter, by definition of $v_I$, we have that there exist $d\in D_I$ such that \medskip

\indent\indent $v_I(Q(\hat d_1,\ldots,\hat d_{i-1},\hat d_{i+1},\dots,\hat d_n/x_1,\ldots,x_{i-1},x_{i+1},\ldots,x_n)(\hat d/x))={\bf F}$. \hfill (10)\kern 1em\medskip

\noindent Since $x$ is not free in $P$, by definition of the notation $(t/x)$, (9) can be written \medskip

\indent\indent $v_I(P(\hat d_1,\ldots,\hat d_{i-1},\hat d_{i+1},\dots,\hat d_n/x_1,\ldots,x_{i-1},x_{i+1},\ldots,x_n)(\hat d/x))={\bf T}$.\medskip 

\noindent By the definition of $v_I$, this together with (10) gives that there exist $d\in D_I$ such that \medskip

\indent\indent $v_I(P(\hat d_1,\ldots,\hat d_{i-1},\hat d_{i+1},\dots,\hat d_n/x_1,\ldots,x_{i-1},x_{i+1},\ldots,x_n)(\hat d/x)\to$

\indent\indent \qquad $Q(\hat d_1,\ldots,\hat d_{i-1},\hat d_{i+1},\dots,\hat d_n/x_1,\ldots,x_{i-1},x_{i+1},\ldots,x_n)(\hat d/x))={\bf F}$ \medskip

\noindent which, by definition of the notation $(t/x)$, can be written \medskip

\indent\indent $v_I(((Q\to P)(\hat d_1,\ldots,\hat d_{i-1},\hat d_{i+1},\dots,\hat d_n/x_1,\ldots,x_{i-1},x_{i+1},\ldots,x_n))(\hat d/x))={\bf F}$. \medskip

\noindent This contradicts (8). $\square$ \medskip

{\bf Proposition 3.10.} Modus Ponens and Generalization are validity preserving. \medskip 

{\bf Proof.} Let $T$ be a theory and let $I$ be an interpretation for $L_T$. For Modus Ponens, consider an application $(P,P\to Q,Q)$ in $L_T$ and suppose that both $P$ and $P\to Q$ are valid in $I$.  Where $x_1,\ldots,x_n$ are the distinct individual variables in $P$ and $Q$, consider the $I$-instances \medskip

\indent\indent $P(\hat d_1,\ldots,\hat d_n/x_1,\ldots,x_n)$ \medskip

\noindent and \medskip

\indent\indent $(P\to Q)(\hat d_1,\ldots,\hat d_n/x_1,\ldots,x_n)$\medskip  

\noindent Then, by assumption, \medskip

\indent\indent $v_I(P(\hat d_1,\ldots,\hat d_n/x_1,\ldots,x_n))={\rm T}$ \hfill (1)\kern 1em\medskip

\noindent and \medskip

\indent\indent $v_I((P\to Q)(\hat d_1,\ldots,\hat d_n/x_1,\ldots,x_n))={\rm T}$ \medskip 

\noindent By definition of the notation $(t/x)$, the latter can be written \medskip

\indent\indent $v_I(P(\hat d_1,\ldots,\hat d_n/x_1,\ldots,x_n)\to Q(\hat d_1,\ldots,\hat d_n/x_1,\ldots,x_n))={\rm T}$ \hfill (2)\kern 1em\medskip

\noindent Then, by the definition of $v_I$, (1) amd (2) imply \medskip

\indent\indent $v_I(Q(\hat d_1,\ldots,\hat d_n/x_1,\ldots,x_n))={\rm T}$ \medskip

\noindent Since the choice of the $\hat d_1,\ldots,\hat d_n/x_1,\ldots,x_n$ was arbitrary, it follows that every $I$-instance of $Q$ is true, so that $Q$ is valid in $I$. 

For Generalization, consider an application $(P,(\forall x)P)$ in $L_T$ and suppose that $P$ is valid in $I$.  Let $x_1,\ldots,x_n$ be the distinct individual variables occurring in $P$.  Then, by the definition of valid, \medskip

\indent\indent $v_I(P(\hat d_1,\ldots,\hat d_n/x_1,\ldots,x_n))={\rm T}$ \hfill (3)\kern 1em\medskip

\noindent for every choice of $d_1,\ldots,d_n\in D_I$.  First suppose that $x$ is not among the $x_1,\ldots,x_n$.  Then, for any $d\in D_I$, \medskip

\indent\indent $P(\hat d_1,\ldots,\hat d_n/x_1,\ldots,x_n)(\hat d/x)$ \medskip

\noindent is just another way of writing \medskip

\indent\indent $P(\hat d_1,\ldots,\hat d_n/x_1,\ldots,x_n)$ \medskip

\noindent so that (3) can be written \medskip

\indent\indent $v_I(P(\hat d_1,\ldots,\hat d_n/x_1,\ldots,x_n)(\hat d/x))={\rm T}$ \medskip

\noindent from which it follows by definition of $v_I$ that \medskip

\indent\indent $v_I((\forall x)(P(\hat d_1,\ldots,\hat d_n/x_1,\ldots,x_n)))={\rm T}$ \medskip

\noindent Since the $\hat d_1,\ldots,\hat d_n/x_1,\ldots,x_n$ were chosen arbitrarily, this means that $(\forall x)P$ is valid in $I$.

Next suppose that $x$ is among $x_1,\ldots,x_n$, say $x$ is $x_i$. Then, by definition of the notation $(t/x)$, (3) can be written \medskip

\indent\indent $v_I((P(\hat d_1,\ldots,\hat d_{i-1},\hat d_{i+1},\ldots,\hat d_n/x_1,\ldots,n_{n-1},x_{n+1},\ldots,x_n))(\hat d_i/x))={\rm T}$. \medskip

\noindent Since $d_i$ was chosen arbitrarily from $D_I$, by definition of $v_I$, \medskip

\indent\indent $v_I((\forall x)P(\hat d_1,\ldots,\hat d_{i-1},\hat d_{i+1},\ldots,\hat d_n/x_1,\ldots,n_{n-1},x_{n+1},\ldots,x_n))={\rm T}$. \medskip

\noindent Since $d_1,\ldots,d_{i-1},d_{i+1},\ldots,d_n$ were chosen arbitrarily from $D_I$, this means that $(\forall x)P$ is valid in $I$. $\square$ \medskip 

{\bf Theorem 3.3.} ({\it Soundness Theorem for FOL\/}) Let $T$ be a theory with no extralogical axioms and let $P$ be a formula of $L_T$.  If $T\vdash P$, then $P$ is logically valid. \medskip

{\bf Proof.} This is established by induction on the length $n$ of proofs $P_1,\ldots,P_n$ in $T$. For the base step, with $n=1$, note that $P_1$ can only be a logical axiom, and hence must be logically valid by Proposition 3.9. 

For the induction step, with $n>1$, suppose first that $P_n$ is inferred from some $P_i$ and $P_j$ where $i,j<n$ by Modus Ponens. The induction hypothesis is that $P_i$ and $P_j$ are logically valid. Consider an interpretation $I$ for $L_T$. By definition of logical validity, $I\models P_i$ and $I\models P_j$, which, by Proposition 3.10, means that $I\models P_n$. Since $I$ was chosen arbitrarily, this means that $P_n$ is logically valid.  

Next suppose that $P_n$ is inferred from some $P_i$ by Generalization.  The induction hypothesis is that $P_i$ is logically valid. Consider an interpretation $I$ for $L_T$.  Then $I\models P_i$, so Proposition 3.10 gives $I\models P_n$. Again, since $I$ was arbitrary, $P_n$ is logically valid.  This completes the induction step and concludes the proof by induction on $n$.  $\square$ \medskip

{\bf Theorem 3.4.} ({\it Consistency Theorem\/}) If a theory $T$ has a model, then $T$ is consistent. \medskip

{\bf Proof.} For a proof by contradiction, suppose that $T$ has a model $I$ but is inconsistent. Since $T$ is inconsistent, there is a formula $P$ of $L_T$ such that $T\vdash P$ and $T\vdash\lnot P$. Let $x_1,\ldots,x_n$ be the variables occurring free in $P$, and let $d_1,\ldots,d_n\in D_I$.  Consider the $I$-instances $P(\hat d_1,\ldots,\hat d_n/x_1,\ldots,x_n$ and $\lnot P(\hat d_1,\ldots,\hat d_n/x_1,\ldots,x_n)$. By the definition of $v_I$, these cannot both be true in $I$.  Thus $P$ and $\lnot P$ cannot both be valid in $I$.  So $I$ cannot be a model of $T$.  This contradicts the assumption. $\square$ \medskip

{\bf Proposition 3.11.} If $T$ is a theory with no extralogical axioms, then $T$ is consistent.  \medskip

{\bf Proof.} Let $T$ be a theory with no extralogical axioms and let $I$ be an interpretation for $L_T$.  Then the theorems of $T$ are valid in $I$, by Theorem 3.3 and the definition of logically valid.  Then $I\models T$, by definition of model for first-order theories.  Whence $T$ is consistent, by Theorem 3.4.  $\square$ \medskip

{\bf Proposition 3.12.} Let $\Gamma$ be the extralogical axioms of a theory $T$ and let $I$ be an interpretation for $L_T$. If $I\models\Gamma$, then $I\models T$. \medskip

{\bf Proof.} This is established by induction on the length $n$ of proofs $P_1,\ldots,P_n$ in $T$.  For the base step, with $n=1$, $P_n$ must be an axiom. If $P_n$ is a logical axiom, then $I\models P_n$ because $P_n$ is logically valid by Proposition 3.9. If $P_n$ is an extralogical axiom, then $I\models P_n$ by assumption.  For the induction step, with $n>1$, the proof is similar to the corresponding case in the proof of Theorem 3.3. To wit, if the premises in an application of either Modus Ponens or Generalization are valid in $I$, then, since the rules are validity preserving, by Proposition 3.10, the conclusions are valid in $I$.  This completes the induction step and concludes the proof by induction on $n$.  $\square$ \medskip

{\bf Proposition 3.13.} Let $(L_t,B_t)$ be an entry in a derivation path for a normal DRS. If there is an interpretation $I$ of $L_t$ that is a model of $B_t$, then $B_t$ is consistent. \medskip

{\bf Proof.} Suppose that $I$ is an interpretation for $L_t$ such that $I\models B_t$. Then $I$ is a model of the set of active formulas in $B_t$, by definition of model for belief sets. Let $T_t$ be the theory determined by $(L_t,B_t)$. By definition of $T_t$, the extralogical axioms of $T_t$ are the active formulas in $B_t$. It follows that $I\models T_t$, by Proposition 3.12.  Hence $T_t$ is consistent, by Theorem 3.4.  

Let $T$ be the theory with language $L_t$ and no extralogical axioms.  Let $\Gamma$ be the set of active formulas in $B_t$. Then $T(\Gamma)$ has the same theorems as $T_t$, by Proposition 3.3.  So $T(\Gamma)$ is consistent.  Thus $\Gamma$ is consistent, by definition of consistent for sets $\Gamma$.  Hence $B_t$ is consistent, by definition of consistent for the belief sets $B_t$.  $\square$ \medskip

{\bf Proposition 4.1.} The DMA is a normal DRS.  \medskip

{\bf Proof.} The DMA employs no belief revision algorithm other than Dialectical Belief Revision, so this proposition is established by Proposition 3.1 and the definition of normal for an arbitrary DRS.  $\square$ \medskip

{\bf Theorem 4.1.} The foregoing algorithms serve to maintain the graph, ignoring are-disjoint links, as a directed acyclic graph without redundant links. \medskip

{\bf Proof.} This uses induction on the length $t$ of derivation paths to show that all graphs $G_t$ have the desired property.  For the base step, where $t=0$, we have $G_t=\emptyset$, and the graph has the desired property by default.  For the induction step, where $t\ge 0$, it is necessary to consider each of the five event types and establish that, if the graph has the desired property before an event of that type, it will continue to have the property after completion of the event up to any point where it invokes an occurrence of another event.

Event Type 1: This enters a document-category link into the graph, unless this would create a redundant path.  Document-category links do not participate in loops.  Thus the desired property is preserved.

Event Type 2: Nothing is added to the graph, but links may be removed.  Removing links cannot affect the desired property.

Event Type 3: No changes are made to the graph.

Event Type 4: An event of this type enters a category subsumption link into the graph, unless this would be a redundant link or would create a loop.  Thus in this case the desired property is preserved.  

Event Type 5: This adds an are-disjoint link to the graph, which has no effect with regard to desired property.

This concludes the induction step and completes the proof by induction on $t$.  $\square$ \medskip

{\bf Theorem 4.2.} After any process initiated by a user input terminates, the resulting belief set will contain a formula of the form $\alpha(a)$ iff the formula is derivable from the formulas corresponding to links in the graph.  \medskip 

{\bf Proof.} This uses induction on the length $t$ of derivation paths to show that every belief set $B_t$ has the desired property.  Note from the descriptions of the five event types that the formulas corresponding to links in the graph are the extralogical axioms that have been input by the users. Recall that these inputs occur during Event Types 1, 4, and 5.  All other formulas in the belief set are derived from these using Aristotelian Syllogism and do not correspond to links in the graph.
 
For the base step, where $t=0$, we have $B_t=\emptyset$, and the belief set has the desired property by default.  For the induction step, where $t\ge 0$, it is necessary to consider each of the three types of user input events and establish that, if the belief set has the desired property before an event of that type, it will continue to have the property after completion of the process initiated by that event.  To simplify the discussion, the following identifies the predicate symbol representing a classification category with that category, so that, for example, an expression such as `the classification category represented by $\alpha$' will be shortened to `the category $\alpha$', unless the more detailed clarification is required by the context.  Similarly, the individual constant representing a document will be associated with that document. 

Consider an input of the form $\alpha(a)$.  This is an event of Type 1.  The associated algorithm adds this formula to the belief set and adds a corresponding is-an-element-of link into the graph. It then proceeds to determine whether this introduces a contradiction into the belief set, and if so, invokes an event of Type 2. That the process initiated by an event of Type 2 does not alter the desired property is established separately below. Next the algorithm proceeds to search for all explicitly expressed subsumptions of the category $\alpha$ by another category $\beta$, and, via Aristotelian Syllogism, derives the formula $\beta(a)$ and provides this formula to the controller. The latter is an event of Type 3, wherein the derived formula is entered into the belief set. The algorithm for Event Type 3 then proceeds recursively through formulas expressing child-parent (subclass-of) relations between categories to find all ancestors $\gamma$ of $\beta$, and, for each, adds the formula $\gamma(a)$ into the belief set (and each time checking as above whether the added formula might lead to an event of Type 2). Thus this process ensures that the resulting belief set will contain all, and only, formulas of the form $\alpha(a)$ that are implicit in the new belief set. This will be the set implicit in the graph, as long as the belief set contains all child-parent category expressions, i.e., formulas  of the form $(\forall x)(\alpha(x)\to\beta(x))$, implicit in the graph. That this is the case is established below. Thus the resulting belief set will have the desired property.

Consider an input of the form $(\forall x)(\alpha(x)\to\beta(x))$.  This is an event of Type 4. The associated algorithm first checks to see if there is information in the belief set indicating that $\alpha$ and $\beta$ are disjoint, and, if so, rejects the input. Otherwise it checks to see if the formula would create a redundant path in the subsumption hierarchy, and if so, rejects the input. Otherwise it checks to see if the formula would create a loop in the subsumption hierarchy, and if so, rejects the input.  Otherwise, the algorithm adds this formula to the belief set and adds a corresponding is-a-subclass-of link into the graph. It then proceeds to search the belief set for any formulas of the form $\alpha(a)$ where $\alpha$ is the predicate symbol in the input formula, and, for each, invokes the process of Event Type 3 discussed above to ensure that the belief set contains all, and only, formulas expressing membership of $a$ in all ancestors $\gamma$ of $\beta$, insofar as the relevant child-parent relations are expressed by formulas in the belief set. Now note that it is a feature of Event Type 4 that an is-a-subclass-of link is entered into the graph if and only if the corresponding formula is entered into the belief set. This establishes the fact promised above; more exactly, it shows that the child-parent relations expressed in the belief set are exactly those that are expressed in the graph. It also shows that the belief set resulting from the present event of Type 4 will have the desired property. 

Last, consider an input of the form $(\forall x)\lnot(\alpha(x)\land\beta(x))$, an event of Type 5. This does not lead to the introduction of any new formulas of the forms $\alpha(a)$ or $(\forall x)(\alpha(x)\to\beta(x))$ into the belief set.  Thus both the collection of formulas of the form $\alpha(a)$ and the collection of formulas of the form $(\forall x)(\alpha(x)\to\beta(x))$ that are derivable from the extralogical axioms remain unchanged.  As with the events triggered by the other two forms of user input, however, this can conclude with one or more instances of Event Type 2. This is considered next. 

An event of Type 2, wherein the formula $\bot$ is provided to the controller, initiates an application of Dialectical Belief Revision. The objective of this process is to identify extralogical axioms that were previously input by the user and that led to this derivation of $\bot$ and to change the status of one or more of these to {\it disbel} so that the derivation is removed.  It is possible, depending on the preference of the human user, that this will be accomplished by changing the status of the extralogical axiom of the form $(\forall x)\lnot(\alpha(x)\land\beta(x))$ that was used as a premise in the associated application of Conflict Detection.  In this case, the resulting belief set obviously has the desired property.  Otherwise, the premise of the form $\alpha(a)$ in the application of Conflict Detection is explored.  Say that the formula of this form is in fact $\alpha(a)$, for some predicate symbol $\alpha$ and individual constant $a$. There are two cases.

Case 1: $\alpha(a)$ is an extralogical axiom. Then this formula is the only remaining candidate for status change in order to remove the conflict.  Changing the status of this formula to {\it disbel} requires removing the is-an-element-of link connecting $a$ to $\alpha$ from the graph. After doing this, the to-list of this formula is used to determine if it has served as a premise in any other rule applications. If so, then these must have been applications of Aristotelian Syllogism leading to the addition of formulas of the form $\beta(a)$ into the belief set for categories $\beta$ that subsume $\alpha$.  Moreover, by the description of the Event Types 1 and 3, this includes all categories that subsume $\alpha$, and no others.  The Dialectical Belief Revision process forward chains through all such deductions, changing the status of all such formulas $\beta(a)$ to {\it disbel}.  This has the effect of removing all assignments of $a$ to categories that subsume $\alpha$, and no others.  Also, no new document-category assignments are added.  Thus the belief set will have the desired property.   

Case 2: $\alpha(a)$ was entered into the belief set as the conclusion of a rule application.  Then $\alpha(a)$ was derived from some formulas $\beta(a)$ and $(\forall x)(\beta(x)\to\alpha(x))$ by Aristotelian Syllogism.  If the former is not an extralogical axiom, then it was derived as the conclusion in an application of Aristotelian Syllogism.  By proceeding backwards in this manner through the derivations using from-lists, one eventually arrives at one or more extralogical axioms of the form $\beta(a)$ and/or one or more extralogical axioms of the form $(\forall x)(\beta(x)\to\gamma(x))$.  This gives two subcases.

Subcase 2.a: An extralogical axiom $\beta(a)$ has its status changed to {\it disbel}.  The process here is identical to the above Case 1, with $\beta$ here playing the role of $\alpha$. 

Subcase 2.b: An extralogical axiom $(\forall x)(\beta(x)\to\gamma(x))$ has its status changed to {\it disbel}.  This requires removing the is-a-subclass-of link connecting $\beta$ to $\gamma$ from the graph.  Thus $\gamma$ no longer subsumes $\beta$.  The algorithm uses to-lists to forward chain through derivations originating with this formula, changing the status of all derived formulas to {\it disbel}.  The rule applications in these derivations must all be of Aristotelian Syllogism.  These will have had the effect of assigning any documents in $\beta$ to $\gamma$ and all ancestors of $\gamma$. The Dialectical Belief Revision process forward chains through all such deductions, changing the status of all such formulas to {\it disbel}.  This has the effect of removing all such document assignments, and no others.  Also, no new document-category assignments are added.  Thus, with respect to formulas expressing document-category relations, the belief set will have the desired property.  

This concludes the induction step and completes the proof by induction on $t$.  $\square$ \medskip

{\bf Theorem 4.3.} For any derivation path in the DMA, the belief set that results at the conclusion of a process initiated by a user input will be consistent.  \medskip

{\bf Proof.} Let $B_0,B_1,\ldots$ and $L_0,L_1,\ldots$  be the belief sets and languages appearing in a derivation path for the DMA. Define subsequences $\hat B_0,\hat B_1,\ldots$ and $\hat L_0,\hat L_1,\ldots$ by (i) $\hat B_0=B_0$ and  $\hat L_0=L_0$, and (ii) for $u\ge 0$, where $t$ is the time stamp for which $\hat B_u=B_t$, $\hat B_{u+1}$ and $\hat L_{u+1}$ are the belief set and language that result from application of the algorithm initiated by the user input that produced $B_{t+1}$ from $B_t$. The theorem amounts to the statement that all the $\hat B_u$ are consistent.  By Proposition 4.1, the DMA is a normal DRS, so, by Proposition 3.13, it is sufficient to show that every $\hat B_u$ has a model, or more exactly, that there is an interpretation $I$ of $\hat L_u$ such that $I\models\hat B_u$. This can be established by induction on $u$.

For the base step, with $u=0$, we have $\hat B_u = B_u=\emptyset$. Thus the set $\Gamma$ of active formulas in $\hat B_u$ is $\emptyset$. Any interpretation $I$ of $\hat L_0=L_0$ is a model of $\emptyset$, by definition of model for sets of formulas; i.e., all formulas in $\emptyset$ are valid in $I$ by default. Thus any such $I$ is a model of $\Gamma$, which makes it a model of $\hat B_u$ by definition of model for belief sets. (Note: There is an alternate, more direct, proof that $\hat B_0$ is consistent. The theory determined by $(L_0,B_0)$ is consistent by Proposition 11, whence $\hat B_0=B_0$ is consistent by definition of consistent for belief sets.)  

For the induction step, with $u\ge 0$, it is required to show for each of the three types of user input, that, if there is an interpretation $I$ of $\hat L_u$ such that $I\models\hat B_u$, then $I$ can be transformed into an interpretation $I^*$ for $\hat L_{u+1}$ such that $I^*\models\hat B_{u+1}$. In all cases, $I^*$ will agree with $I$ on all predicate symbols and individual constant symbols of $\hat L_u$.

Consider an input $\alpha(a)$. This is an event of Type 1, which has three phases: (i) add the input formula into the current belief set, (ii) scan the belief set for any opportunities to apply Conflict Detection, and (iii) scan the belief set for any opportunities to apply Aristotelian Syllogism.  The latter two phases lead to further events, respectively, of Types 2 and 3. We consider these three phases in order.  Let $t$ be such that $\hat B_u=B_t$. Then $\hat L_u=L_t$. 

Phase (i): The objective is to show that $I$ can be expanded to an interpretation $I'$ of $L_{t+1}$ such that $I'\models\alpha(a)$. If either $\alpha$ or $a$ is missing from $L_t$, then $L_{t+1}$ will be the language obtained from $L_t$ by adding the ones that are missing; otherwise $L_{t+1}=L_t$.  Case 1: Neither are added.  Let $D_{I'}=D_I$ and extend $I$ to $I'$ by setting $I'(\alpha)=I(\alpha)\cup\{I(a)\}$, i.e., the relation designated by $\alpha$ is expanded to include the document designated by $a$.  Case 2: $\alpha$ is added, but $a$ is not.  Let $D_{I'}=D_I$ and extend $I$ to $I'$ by assigning $I'(\alpha)=\{I(a)\}$, i.e., $I'(\alpha)$ is a new unary relation on $D_{I'}$ that holds only for $I(a)$. Case 3: $a$ is added, but $\alpha$ is not.  Subcase 3.a: $a$ is taken as representing a new document $d$ not in the domain $D_I$ (perhaps the most typical case).  Let $D_{I'}=D_I\cup\{d\}$, assign $I'(a)=d$, and let $I'(\alpha)=I(\alpha)\cup\{d\}$.  Subcase 3.b: $a$ is taken as representing a document $d\in D_I$ (this is atypical inasmuch as in the DMA one should not need more than one representation for the same document).  Let $D_{I'}=D_I$ and let $I'(\alpha)=I(\alpha)\cup\{d\}$.  Case 4: Both $\alpha$ and $a$ are added.  Subcase 4.a: $a$ is taken as representing a new document $d$ not in the domain $D_I$ (typical). Let $D_{I'}=D_I\cup\{d\}$, assign $I'(a)=d$, and set $I'(\alpha)=\{d\}$.  Subcase 4.b: $a$ is taken as representing a document $d\in D_I$ (atypical).  Let $D_{I'}=D_I$, assign $I'(a)=d$, and set $I'(\alpha)=\{d\}$.  It follows in every case that $I'\models\alpha(a)$.  

Phase (ii): Note that $B_{t+1}=B_t\cup\{(\alpha(a),\lambda)\}$, where $\lambda$ is the appropriate label. Since $I$ is a model of $B_t$, the foregoing makes $I'$ a model of $B_{t+1}$, unless the inclusion of $I(a)$ in the relation $I(\alpha)$ conflicts with some active formulas in $B_t$ expressing that $I(a)\in I(\beta)$ and that $I(\alpha)$ and $I(\beta)$ are disjoint.  Note that this can only occur in the above Case 1, since only it implies that both $a$ and $\alpha$ are already in $L_t$.  Whether there is such a conflict is determined, as described for Event Type 1, by scanning the set $B_{t+1}$ for any active formulas of the form $(\forall x)\lnot(\alpha(x)\land\beta(x))$ or $(\forall x)\lnot(\beta (x)\land\alpha(x))$, where $\alpha$ is the predicate symbol of the input formula and $\beta$ is any other predicate symbol, and for each such formula scanning for any active occurrence of the formula $\beta(a)$, where $a$ is the individual constant of the input formula.  All such formulas $\beta(a)$ implicit in $B_{t+1}$ are ensured to be present in $B_{t+1}$ by Theorem 6.  If no such conflicting formulas are found, then, as just asserted, $I'\models B_{t+1}$.  Otherwise, the found formulas will have triggered an event of Type 2 leading to an application of Dialectical Belief Revision, in which case it remains to show that further modifications can be made to $I'$ and $B_{t+1}$ that transforms $I'$ into a model of the belief set, call it  $B_{t^*}$, that results from this process.  This fact regarding Event Type 2 is established below.

For phase (iii), let $B_{t^{**}}$ be the most recent belief set; i.e., it is either the above $B_{t+1}$ or the above $B_{t^*}$, depending on whether any conflicts were detected.  Let $I^{**}$ be the given model of $B_{t^{**}}$.  In this phase,  $B_{t^{**}}$ is searched for active formulas of the form $(\forall x)(\alpha(x)\to\beta(x))$, where $\alpha$ is the predicate symbol of the input formula, and, if found, Aristotelian Syllogism is applied to this and the input formula to infer $\beta(a)$, which is provided to the controller. This constitutes an event of Type 3.  Note that there can be any number of such $\beta$, each leading to an event of Type 3. Thus it remains to show that, if, at the beginning of an event of Type 3, the belief set has a model, then, at the conclusion of the process initiated by that event, the resulting belief set has a model.  This is established below. Observe that $\hat B_{u+1}$ is the belief set that results from this process. Note also that $\hat L_{u+1}=L_{t+1}$. Given the following verifications regarding events of Type 2 and 3, the foregoing shows that $\hat L_{u+1}$ has an interpretation that is a model of $\hat B_{u+1}$. 

Consider an event of Type 2, namely, providing $\bot$ to the controller. This only occurs via an application of Conflict Detection. The formula $\bot$ is entered into the belief set, and this initiates an application of Dialectical Belief Revision. Let $B_t$ be the belief set that is current prior to the occurrence of this event, and let $I$ be an interpretation of $L_t$ that models $B_t$. It is desired to show that $I$ can be transformed into a interpretation $I'$ of $L_t$ that is a model of the belief set $B_{t'}$ that results from the Dialectical Belief Revision process.

Let $\alpha(a)$, $\beta(a)$, and $(\forall x)\lnot(\alpha(x)\land\beta(x))$ be the premises in the application of Conflict Detection.  The first phase of the Dialectical Belief Revision process employs from-lists to backtrack through the derivations of these formulas, identifying all extralogical axioms involved in their proofs.  One of these will be the formula $(\forall x)\lnot(\alpha(x)\land\beta(x))$.  If the conflict is resolved by changing the status of this formula to {\it disbel}, then $B_{t'}$ is obtained from $B_t$ by making this change in this formula's label.  Then letting $I'=I$ gives $I'\models B_{t'}$.  

The formulas $\alpha(a)$ and $\beta(a)$ in the application of Conflict Detection could also be extralogical axioms.  If either is, then the conflict could be resolved by changing the status of that formula to {\it disbel}.  Suppose the formula whose status is changed is $\alpha(a)$.  (The considerations will be similar for $\beta(a)$.)  Let $B_{t'}$ be obtained from $B_t$ by making this change in this formula's label.  Let $I'$ be obtained from $I$ by  letting $I'(\alpha)=I(\alpha)-\{I(a)\}$, i.e., removing the document $I(a)$ from the category $I(\alpha)$.  Then $I'\models B_{t'}$.

The formulas $\alpha(a)$ and $\beta(a)$ in the application of Conflict Detection might not be extralogical axioms, however.  Suppose that this is true for $\alpha(a)$.  (The considerations will be similar for $\beta(a)$.)  Then this formula was derived by an application of Aristotelian Syllogism.  Suppose that the premises in this rule application were $\gamma(a)$ and $(\forall x)(\gamma(x)\to\alpha(x))$.  Consider the ways in which these formulas could have come to be input into the belief set.  If $\gamma(a)$ had been input as an extralogical axiom, then the conflict could be resolved by changing the status of this formula to {\it disbel}, after which the second phase of Dialectical Belief Revision forward-chains via the formula's to-list and changes the status of $\alpha(a)$ to {\it disbel}.   In this case, $B_{t'}$ is obtained from $B_t$ by making this change in the label for $\gamma(a)$.  Let $I'$ be obtained from $I$ by setting $I'(\gamma)=I(\gamma)-\{I(a)\}$ and $I'(\alpha)=I(\alpha)-\{I(a)\}$, i.e., removing the document $I(a)$ from the categories $I(\gamma)$ and $I(\alpha)$.  Then $I'\models B_{t'}$.  If $(\forall x)(\gamma(x)\to\alpha(x))$ had been input as an extralogical axiom, then the conflict could be resolved by changing the status of this formula to {\it disbel}, after which the second phase of Dialectical Belief Revision forward-chains via the formula's to-list and changes the status of $\alpha(a)$ to {\it disbel}.  In this case, let $B_{t'}$ is obtained from $B_t$ by making this change in the label for $(\forall x)(\gamma(x)\to\alpha(x))$.  Let $I'$ be obtained from $I$ by  setting $I'(\alpha)=I(\alpha)-\{I(a)\}$, i.e., removing the document $I(a)$ from the category $I(\alpha)$.  Then $I'\models B_{t'}$.

If $\gamma(a)$ had not been input as an extralogical axiom, then it was derived by an application of Aristotelian Syllogism.  By continuing to backtrack in this manner using from-lists, the process eventually finds all extralogical axioms involved in the derivation of the current occurrence of $\bot$.  Some will be of the form $\alpha'(a)$, where $a$ is the individual constant in the premise of the related application of Conflict Detection, and some will be of the form $(\forall x)(\beta'(x)\to\gamma'(x))$.  One or more of these will have their belief status changed to {\it disbel}, enough to ensure that the derivation of $\bot$ no longer applies.  Then $B_{t'}$ is obtained from $B_t$ by making these changes in these formulas' labels.  Note incidentally that the corresponding $B_{t'}$ will have been obtained by following the to-lists of all such formulas, similarly changing the labels of all their consequences up to and including the triggering occurrence of $\bot$.  Let $I'$ be obtained from $I$ by  setting $I'(\alpha'')=I(\alpha'')-\{I(a)\}$  for every formula $\alpha''(a)$ whose label is changed to {\it disbel} in this second phase of the Dialectical Belief Revision process.  Then $I'\models B_{t'}$.

The forgoing shows that, whichever of the available choices are made to resolve the detected conflict, there will be an interpretation $I'$ of $L_t$ that is a model of the resulting belief set $B_{t'}$. 

Consider an event of Type 3, where a formula of the form $\alpha(a)$ is provided to the controller as the result of an inference rule application. This only occurs via an application of Aristotelian Syllogism. The formula $\alpha(a)$ is entered into the belief set. Because no new symbols need to be added, $\hat L_{u+1}=\hat L_u$.  Let $I$ be an interpretation of $\hat L_u$ that models $\hat B_u$. It is desired to show that $I$ can be transformed into a model $I'$ of the belief set $B_{u+1}$ that results from the process associated with this event. This can be established by the same considerations as above for Event Type 1, Phases (i) and (ii), except that, in Phase (i), here only Case 1 applies.   

Consider an input $(\forall x)(\alpha(x)\to\beta(x))$.  This is an event of Type 4.  Let $t$ be such that $\hat B_u=B_t$. Then $\hat L_u=L_t$. Let $I$ be an interpretation of $\hat L_u$ that models $\hat B_u$. Then $I$ is an interpretation of $L_t$ that models $B_t$.  As prescribed by the event type's description, if $\alpha$ and $\beta$ are already in $L_t$, the input is rejected if there are active formulas asserting that the categories represented by $\alpha$ and $\beta$ are disjoint, or if the input formula would create either a redundant path or a loop.  Assuming that none of these is the case, then the process has two phases: (i) add the input formula into the belief set, (ii) scan the belief set for any opportunities to apply Aristotelian syllogism, leading to invocations of Event Type 3.

Phase (i): Here $B_{t+1}=B_t\cup\{((\forall x)(\alpha(x)\to\beta(x)),\lambda)\}$, where $\lambda$ is the appropriate label.  The objective is to show that $I$ can be expanded to an interpretation $I'$ of $L_{t+1}$ such that $I'\models B_{t+1}$. This amounts to showing that $I$ can be expanded to an interpretation $I'$ of $L_{t+1}$ such that $I'\models (\forall x)(\alpha(x)\to\beta(x))$.  If either $\alpha$ or $\beta$ is missing from $L_t$, then $L_{t+1}$ will be the language obtained from $L_t$ by adding the ones that are missing; otherwise $L_{t+1}=L_t$. There are four cases depending on which symbols are added. In all cases, set $D_{I'}=D_I$.  Case 1: Neither is added.  Then $\alpha$ and $\beta$ are already in $L_t$.  Form $I'$ from $I$ by setting $I'(\beta)=I(\beta)\cup(I(\alpha)-I(\beta))$, i.e., add to $I(\beta)$ the elements in $I(\alpha)$ that are not already in $I(\beta)$.  Note that in a conventional first-order system, such a redefinition of $I(\beta)$ would not be permissible, since it could potentially invalidate some theorems of the form $\lnot\beta(a)$  expressing that $I(a)$ is not a member of $I(\beta)$.  This potential problem is ruled out in the DMA, however, since the DMA controller does not enable formulas of this form to ever be input into the belief set, either by a human user or via derivations.  Case 2: $\alpha$ is added, but $\beta$ is not.  Set $I'(\alpha)=\emptyset$, i.e., introduce a new relation corresponding to $\alpha$ and let it be the empty subset of $I(\beta)$.  Case 3: $\beta$ is added, but $\alpha$ is not.  Set $I'(\beta)=I'(\alpha)$, i.e., introduce a new relation corresponding to $\beta$ and make it coextensive with $I(\alpha)$.  Case 4: Both $\alpha$ and $\beta$ are added.  Set $I'(\alpha)=I'(\beta)=\emptyset$.  It follows in every case that the formula $(\forall x)(\alpha(x)\to\beta(x))$ will be true in $I'$. Thus, in every case, $I'\models B_{t+1}$.

Phase (ii): The above $B_{t+1}$ is scanned for formulas of the form $\alpha(a)$, where $\alpha$ is the predicate symbol of the input formula, and, if found, Aristotelian Syllogism is applied, leading to an event of Type 3. That this results in a belief set with a model was established in the foregoing. 

Let $\hat L_{u+1}$ and$\hat B_{u+1}$ be the language and belief set that results from the above process. It follows that there is an interpretation $I'$ of $\hat L_{u+1}$ such that $I'\models B_{u+1}$. 

Consider an input $(\forall x)\lnot(\alpha(x)\land\beta(x))$.  This is an event of Type 5.   Let $t$ be such that $\hat B_u=B_t$.  Then $\hat L_u=L_t$.  Let $I$ be an interpretation of $\hat L_u$ that models $\hat B_u$.  Then $I$ is an interpretation of $L_t$ that models $B_t$.  Then $B_{t+1}=B_t\cup\{((\forall x)\lnot(\alpha(x)\land\beta(x)),\lambda)\}$, where $\lambda$ is the appropriate label. It is desired to show that $I$ can be expanded to an interpretation $I'$ of $L_{t+1}$ such that $I'\models B_{t+1}$. This amounts to showing that $I$ can be expanded to an interpretation $I'$ of $L_{t+1}$ such that $I'\models (\forall x)\lnot(\alpha(x)\land\beta(x))$.   If either $\alpha$ or $\beta$ is missing from $L_t$, then $L_{t+1}$ will be the language obtained from $L_t$ by adding the ones that are missing; otherwise $L_{t+1}=L_t$.   There are four cases depending on which symbols are added.  In all cases, set $D_{I'}=D_I$.  Case 1: Neither is added.  Then $\alpha$ and $\beta$ are already in $L_t$, and $I(\alpha)\cap I(\beta)=\emptyset$.  Set $I'=I$.  Case 2: $\alpha$ is added, but $\beta$ is not.  Set $I'(\alpha)=\emptyset$, i.e., introduce a new relation corresponding to $\alpha$ and let it be the empty set.  Case 3: $\beta$ is added, but $\alpha$ is not.  Set $I'(\beta)=\emptyset$, i.e., introduce a new relation corresponding to $\beta$ and let it be the empty set.  Case 4: Both $\alpha$ and $\beta$ are added.  Set $I'(\alpha)=I'(\beta)=\emptyset$.  Thus in every case the formula $(\forall x)\lnot(\alpha(x)\land\beta(x))$ will be true in $I'$.  

It follows that $I'\models B_{t+1}$, unless the provision that $I(\alpha)$ and $I(\beta)$ are disjoint conflicts with some formulas in $B_{t+1}$ expressing that $I(a)\in I(\alpha)$ and $I(a)\in I(\beta)$.  Note that this can only occur in the above Case 1, since it implies that both $\alpha$ and $\beta$ are already in $L_t$.  Whether there is a conflict of this type is determined by scanning the set $B_{t+1}$ for any formulas of the form $\alpha(a)$ and $\beta(a)$, where $\alpha$ and $\beta$ are the predicate symbols of the input formula and $a$ is any individual constant.  All such formulas $\alpha(a)$ and $\beta(a)$ implicit in $B_{t+1}$ are ensured to be present in $B_{t+1}$ by Theorem 4.2.  

If no such formulas are found, then $\hat B_{u+1}=B_{t+1}$ and we have that $I'\models\hat B_{u+1}$.  Otherwise, the found formulas will have triggered an event of Type 2 leading to an application of Dialectical Belief Revision. That this leads to a belief set with a model was established above.

This concludes the induction step and completes the proof by induction on $u$.  $\square$ \medskip

{\bf Proposition 5.1.} Every MIS is a normal DRS.  \medskip

{\bf Proof.} No MIS employs a belief revision algorithm other than Dialectical Belief Revision, so this proposition is established by Proposition 3.1 and the definition of normal for an arbitrary DRS.  $\square$ \medskip

{\bf Theorem 5.1.} The foregoing algorithms serve to maintain the hierarchy with respect to the object and kind nodes as a directed acyclic graph without redundant links. \medskip

{\bf Proof.} This uses induction on the length $t$ of derivation paths to show that all inheritance hierarchies $H_t$ have the desired property.  For the base step, where $t=0$, we have $H_t=\emptyset$, and the hierarchy has the desired property by default.  For the induction step, where $t\ge 0$, it is necessary to consider each of the eight event types and establish that, if the hierarchy has the desired property before an event of that type, it will continue to have the property after completion of the event up to any point where it invokes an occurrence of another event.

Event Type 1: This enters an object-kind link into the hierarchy, unless this would create a redundant path.  Object nodes do not participate in loops.  Thus the desired property is preserved.

Event Types 2, 3, and 4: No changes are made to the hierarchy.

Event Type 5: Nothing is added to the hierarchy, but links may be removed.  Removing links cannot affect the desired property.

Event Type 6: An event of this type enters a subkind-kind link into the hierarchy, unless this would create a loop or a redundant path.  Thus in this case the desired property is preserved.  

Event Types 7 and 8: These add has-property links to the hierarchy, which has no effect with regard to desired property.

This concludes the induction step and completes the proof by induction on $t$.  $\square$ \medskip

{\bf Theorem 5.2.} After any process initiated by a user input terminates, the resulting belief set will contain a formula of the form $\alpha^{(k)}(a)$ or $\alpha^{(p)}(a)$ or $\lnot\alpha^{(p)}(a)$  iff the formula is derivable from the formulas corresponding to links in the inheritance hierarchy, observing the specificity principle.  \medskip 

{\bf Proof.} This uses induction on the length $t$ of derivation paths to show that every belief set $B_t$ has the desired property.  Note from the descriptions of the eight event types that the formulas corresponding to links in the inheritance hierarchy are the extralogical axioms that have been input by the users. Recall that these inputs occur during Event Types 1, 6, 7 and 8.  All other formulas in the belief set are derived from these using Aristotelian Syllogism and do not correspond to links in the hierarchy.
 
For the base step, where $t=0$, we have $B_t=\emptyset$, and the belief set has the desired property by default.  For the induction step, where $t\ge 0$, it is necessary to consider each of the four types of user input events and establish that, if the belief set has the desired property before an event of that type, it will continue to have the property after completion of the process initiated by that event.  To simplify the discussion, the following identifies the predicate symbol representing a kind or property with that kind or property, so that, for example, an expression such as `the kind represented by $\alpha$' will be shortened to `the kind $\alpha$', unless the more detailed clarification is required by the context.  Similarly, the individual constant representing an object will be associated with that object. 

Consider an input of the form $\alpha^{(k)}(a)$.  This is an event of Type 1.  The associated algorithm adds this formula to the belief set and adds a corresponding object-kind link into the hierarchy. It then proceeds to search for all explicitly expressed subsumptions of the kind $\alpha^{(k)}$ by another kind $\beta^{(k)}$, and, via Aristotelian Syllogism, derives the formula $\beta^{(k)}(a)$ and provides this formula to the controller. The latter is an event of Type 2, wherein the derived formula is entered into the belief set. The algorithm for Event Type 2 then proceeds recursively through formulas expressing child-parent (subkind-kind) relations between kind nodes to find all ancestors $\gamma^{(k)}$ of $\beta^{(k)}$, and, for each, adds the formula $\gamma^{(k)}(a)$ into the belief set.  Thus this process ensures that the resulting belief set will contain all, and only, formulas of the form $\alpha^{(k)}(a)$ that are implicit in the new belief set. This will be the set implicit in the hierarchy, as long as the belief set contains all child-parent kind expressions, i.e., formulas  of the form $(\forall x)(\alpha^{(k)}(x)\to\beta^{(k)}(x))$, implicit in the hierarchy. That this is the case is established below.  Thus the resulting belief set will have the desired property regarding formulas of the form $\alpha^{(k)}(a)$ (as expressed in the Theorem statement and not to be confused with the input formula).

The algorithm for Event Type 1 next searches the belief set for formulas expressing that objects of the kind $\alpha^{(k)}$ have some property $\beta^{(p)}$, and for each such formula applies Aristotelian Syllogism to infer $\beta^{(p)}(a)$.  This invokes an event of Type 3, which adds this formula to the belief set unless this is overridden by a more specific occurrence of $\lnot\beta^{(p)}(a)$ in the hierarchy.  Note that this is the only way that a formula of this form can be entered into the belief set.  The Type 3 event algorithm then proceeds to search for a formula that contradicts $\beta^{(p)}(a)$, and, if found, invokes an event of Type 5.   The Type 3 event algorithm may do this zero or more times.  That a Type 5 event preserves the desired property for the belief set is established below.  

 Last it searches the belief set for formulas expressing that object of the kind $\alpha^{(k)}$ does not have some property $\beta^{(p)}$, and for each such formula applying Aristotelian Syllogism to infer $\lnot\beta^{(p)}(a)$.  This invokes an event of Type 4, which adds this formula to the belief set unless this is overridden by a more specific occurrence of $\beta^{(p)}(a)$ in the hierarchy.  Note that this is the only way that a formula of this form can be entered into the belief set.  The Type 4 event algorithm then proceeds to search for a formula that contradicts $\lnot\beta^{(p)}(a)$, and, if found, invokes an event of Type 5.   The Type 4 event algorithm may do this zero or more times.  That a Type 5 event preserves the desired property for the belief set is established below. 

It follows that the resulting belief set will have the desired property regarding formulas of the form $\alpha^{(p)}(a)$  and $\lnot\alpha^{(p)}(a)$ (as expressed in the Theorem statement).  This completes the consideration for an input to the form $\alpha^{(k)}(a)$.

Consider an input of the form $(\forall x)(\alpha^{(k)}(x)\to\beta^{(k)}(x))$.  This is an event of Type 6. The associated algorithm first checks to see if the formula would create a redundant path or a loop in the subsumption hierarchy, and if so, rejects the input.  Otherwise, the algorithm adds this formula to the belief set and adds a corresponding subkind-kind link into the hierarchy.  It then proceeds to search the belief set for any formulas of the form $\alpha^{(k)}(a)$ where $\alpha^{(k)}$ is the predicate symbol in the premise component of the input formula, and, for each, invokes the process of Event Type 3 discussed above to ensure that the belief set contains all, and only, formulas expressing membership of $a$ in all ancestors $\gamma^{(k)}$ of $\beta^{(k)}$, insofar as the relevant child-parent relations are expressed by formulas in the belief set.  Now note that it is a feature of Event Type 6 that a subkind-kind link is entered into the hierarchy if and only if the corresponding formula is entered into the belief set. This establishes the fact promised above, namely, it shows that the child-parent relations expressed in the belief set are exactly those that are expressed in the hierarchy.  It also shows that the belief set resulting from the present event of Type 6 will have the desired property. 

Consider an input of the form $(\forall x)(\alpha^{(k)}(x)\to\beta^{(p)}(x))$, an event of Type 7.  The algorithm for this event type adds this formula into the belief set and adds the corresponding has-property link into the inheritance hierarchy.  It then proceeds to search the belief set for formulas of the form $\alpha^{(k)}(a)$, and, for each, applies Aristotelian Syllogism to infer $\beta^{(p)}(a)$, and adds this to the belief set unless this is overridden by a more specific occurrence of $\lnot\beta^{(p)}(a)$ in the hierarchy (and checking as above whether the added formula might introduce a contradiction and lead to an event of Type 5).  As noted above, this is the only way that a formula of this form can be entered into the belief set.  It follows that the belief set resulting from the present event of Type 7 will have the desired property.  

Consider an input of the form $(\forall x)(\alpha^{(k)}(x)\to\lnot\beta^{(p)}(x))$, an event of Type 8.  The algorithm for this event type adds this formula into the belief set and adds the corresponding has-property link into the inheritance hierarchy.  It then proceeds to search the belief set for formulas of the form $\alpha^{(k)}(a)$, and, for each, applies Aristotelian Syllogism to infer $\lnot\beta^{(p)}(a)$, and adds this to the belief set unless this is overridden by a more specific occurrence of $\beta^{(p)}(a)$ in the hierarchy (and checking as above whether the added formula might introduce a contradiction and lead to an event of Type 5).  As noted above, this is the only way that a formula of this form can be entered into the belief set.  It follows that the belief set resulting from the present event of Type 8 will have the desired property.

An event of Type 5, wherein the formula $\bot$ is provided to the controller, initiates an application of Dialectical Belief Revision. The objective of this process is to identify extralogical axioms that were previously input by a user and that led to this derivation of $\bot$, and to change the status of one or more of these to {\it disbel} so that the derivation is removed.  There are two ways in which this process can be triggered: (1) a formula of the form $\alpha^{(p)}(a)$ is derived by means of a rule application and is found to contradict a formula of the form $\lnot\alpha^{(p)}(a)$ already active in the belief set, and (2) a formula of the form $\lnot\alpha^{(p)}(a)$ is derived by means of a rule application and is found to contradict a formula of the form $\alpha^{(p)}(a)$ already active in the belief set.  We here consider only Case 1, as Case 2 is analogous.

In this case, $\alpha^{(p)}(a)$ was derived from some formulas $\beta^{(k)}(a)$ and $(\forall x)(\beta^{(k)}(x)\to\alpha^{(p)}(x))$ by Aristotelian Syllogism.  By the definition of the controller, the latter formula must be an extralogical axiom input by a user, and the former must have been derived from some formulas of the forms $\gamma^{(k)}(a)$ and $(\forall x)(\gamma^{(k)}(x)\to\beta^{(k)}(x))$ by Aristotelian Syllogism.  This gives two subcases.

Subcase 1.a: The extralogical axiom $(\forall x)(\beta^{(k)}(x)\to\alpha^{(p)}(x))$ has its status changed to {\it disbel}.  This requires removing the has-property link connecting $\beta^{(k)}$ to $\alpha^{(p)}$ from the hierarchy.  The algorithm uses to-lists to forward chain through derivations originating with this formula, changing the status of all derived formulas to {\it disbel}.  The rule applications in these derivations must all be of Aristotelian Syllogism, and will have had the effect of assigning objects in $\beta^{(k)}$ the property $\alpha^{(p)}$.  Note that such assignments are not represented by links in the hierarchy. The Dialectical Belief Revision process has the effect of removing all such property assignments (changing their statuses to {\it disbel}), and no others.  Also, no new property assignments are added.  It follows that the belief set will have the desired property.  

Subcase 1.b: The derivation of $\beta^{(k)}(a)$ is explored. This requires working backwards through derivations using from-lists, eventually arriving at some extralogical axioms of the forms $\gamma^{(k)}(a)$ and $(\forall x)(\gamma^{(k)}(x)\to\delta^{(k)}(x))$.  If a formula of the former form has its status changed to {\it disbel}, then the corresponding object-kind link is removed from the hierarchy and Dialectical Belief Revision proceeds to forward chain through to-lists, changing the status of all derived formulas to {\it disbel}. These derived formulas will have been assignments of $a$ to kinds that subsume $\gamma$, leading up the hierarchy to $\beta^{(k)}$, and then to the formula $\alpha^{(p)}(a))$. If a formula of the latter form has its status changed to {\it disbel}, then the corresponding subkind-kind link is removed from the hierarchy and Dialectical Belief Revision proceeds to forward chain through to-lists, changing the status of all derived formulas to {\it disbel}.  Again, these derived formulas will have been assignments of $a$ to kinds that subsume $\gamma$, leading up the hierarchy to $\beta^{(k)}$, and then to the formula $\alpha^{(p)}(a))$. Thus regardless of which of the discovered extralogical axioms are selected for removal (status change) the desired property of the belief set will be preserved. 

This concludes the induction step and completes the proof by induction on $t$.  $\square$ \medskip

{\bf Theorem 5.3.} For any derivation path in an MIS, the belief set that results at the conclusion of a process initiated by a user input will be consistent with respect to the formulas of the forms $\alpha^{(k)}(a)$, $(\forall x)(\alpha^{(k)}(x)\to\beta^{(p)}(x))$, and $\alpha^{(p)}(a)$.  \medskip

{\bf Proof.} This adopts the approach presented in Theorem 7 for the DMA.  Let $B_0,B_1,\ldots$ and $L_0,L_1,\ldots$  be the belief sets and languages appearing in a derivation path for an MIS. Define subsequences $\hat B_0,\hat B_1,\ldots$ and $\hat L_0,\hat L_1,\ldots$ by (i) $\hat B_0=B_0$ and  $\hat L_0=L_0$, and (ii) for $u\ge 0$, where $t$ is the time stamp for which $\hat B_u=B_t$, $\hat B_{u+1}$ and $\hat L_{u+1}$ are the belief set and language that result from application of the algorithm initiated by the user input that produced $B_{t+1}$ from $B_t$. The theorem amounts to the statement that all the $\hat B_u$ are consistent.  By Proposition 15, every MIS is a normal DRS, so, by Proposition 13, it is sufficient to show that every $\hat B_u$ has a model, or more exactly, that there is an interpretation $I$ of $\hat L_u$ such that $I\models\hat B_u$. This can be established by induction on $u$.

For the base step, with $u=0$, we have $\hat B_u = B_u=\emptyset$. Thus the set $\Gamma$ of active formulas in $\hat B_u$ is $\emptyset$. Any interpretation $I$ of $\hat L_0=L_0$ is a model of $\emptyset$, by definition of model for sets of formulas; i.e., all formulas in $\emptyset$ are valid in $I$ by default. Thus any such $I$ is a model of $\Gamma$, which makes it a model of $\hat B_u$ by definition of model for belief sets.

For the induction step, with $u\ge 0$, it is required to show for each of the four types of user input, that, if there is an interpretation $I$ of $\hat L_u$ such that $I\models\hat B_u$, then $I$ can be transformed into an interpretation $I^*$ for $\hat L_{u+1}$ such that $I^*\models\hat B_{u+1}$. In all cases, $I^*$ will agree with $I$ on all predicate symbols and individual constant symbols of $\hat L_u$.

Consider an input $\alpha^{(k)}(a)$. This is an event of Type 1, which has four phases: (i) add the input formula into the current belief set, (ii) scan the belief set for any formulas of the form $(\forall x)(\alpha^{(k)}(x)\to\beta^{(k)}(x))$ and apply Aristotelian Syllogism, (iii) scan the belief set for any formulas of the form $(\forall x)(\alpha^{(k)}(x)\to\beta^{(p)}(x))$ and apply Aristotelian Syllogism, (iv) scan the belief set for any formulas of the form $(\forall x)(\alpha^{(k)}(x)\to\lnot\beta^{(p)}(x))$ and apply Aristotelian Syllogism.  The latter three phases lead to further events, respectively, of Types 2, 3 and 4. We consider these four phases in order.  Let $t$ be such that $\hat B_u=B_t$. Then $\hat L_u=L_t$. 

Phase (i): The objective is to show that $I$ can be expanded to an interpretation $I'$ of $L_{t+1}$ such that $I'\models\alpha^{(k)}(a)$. If either $\alpha^{(k)}$ or $a$ is missing from $L_t$, then $L_{t+1}$ will be the language obtained from $L_t$ by adding the ones that are missing; otherwise $L_{t+1}=L_t$.  Case 1: Neither are added.  Let $D_{I'}=D_I$ and extend $I$ to $I'$ by setting $I'(\alpha^{(k)})=I(\alpha^{(k)})\cup\{I(a)\}$, i.e., the relation designated by $\alpha^{(k)}$ is expanded to include the object designated by $a$.  Case 2: $\alpha^{(k)}$ is added, but $a$ is not.  Let $D_{I'}=D_I$ and extend $I$ to $I'$ by assigning $I'(\alpha^{(k)})=\{I(a)\}$, i.e., $I'(\alpha^{(k)})$ is a new unary relation on $D_{I'}$ that holds only for $I(a)$. Case 3: $a$ is added, but $\alpha^{(k)}$ is not.  Subcase 3.a: $a$ is taken as representing a new object $d$ not in the domain $D_I$ (perhaps the most typical case).  Let $D_{I'}=D_I\cup\{d\}$, assign $I'(a)=d$, and let $I'(\alpha^{(k)})=I(\alpha^{(k)})\cup\{d\}$.  Subcase 3.b: $a$ is taken as representing a object $d\in D_I$ (this is atypical inasmuch as in an MIS one should not need more than one representation for the same object).  Let $D_{I'}=D_I$ and let $I'(\alpha^{(k)})=I(\alpha^{(k)})\cup\{d\}$.  Case 4: Both $\alpha^{(k)}$ and $a$ are added.  Subcase 4.a: $a$ is taken as representing a new object $d$ not in the domain $D_I$ (typical). Let $D_{I'}=D_I\cup\{d\}$, assign $I'(a)=d$, and set $I'(\alpha^{(k)})=\{d\}$.  Subcase 4.b: $a$ is taken as representing an object $d\in D_I$ (atypical).  Let $D_{I'}=D_I$, assign $I'(a)=d$, and set $I'(\alpha^{(k)})=\{d\}$.  It follows in every case that $I'\models\alpha(a)$.  Note that $B_{t+1}=B_t\cup\{(\alpha^{(k)}(a),\lambda)\}$, where $\lambda$ is the appropriate label.  Since $I$ is a model of $B_t$, the foregoing makes $I'$ a model of $B_{t+1}$.

Phase (ii): In this phase,  $B_{t+1}$ is searched for active formulas of the form $(\forall x)(\alpha^{(k)}(x)\to\beta^{(k)}(x))$, where $\alpha^{(k)}$ is the predicate symbol of the input formula, and, if found, Aristotelian Syllogism is applied to this and the input formula to infer $\beta^{(k)}(a)$, which is provided to the controller. This constitutes an event of Type 2.  Note that there can be any number of such $\beta^{(k)}$, each leading to an event of Type 2. Thus it remains to show that, if, at the beginning of an event of Type 2, the belief set has a model, then, at the conclusion of the process initiated by that event, the resulting belief set has a model.  This is established below.

Phase (iii): Let $B^*$ be the most recent belief set, i.e., the one that resulted from the above instances of Event Type 2.  In this phase, $B^*$  is searched for active formulas of the form $(\forall x)(\alpha^{(k)}(x)\to\beta^{(p)}(x))$, where $\alpha^{(k)}$ is the predicate symbol of the input formula, and, if found, Aristotelian Syllogism is applied to this and the input formula to infer $\beta^{(p)}(a)$, which is provided to the controller. This constitutes an event of Type 3.  Note that there can be any number of such $\beta^{(p)}$, each leading to an event of Type 3. Thus it remains to show that, if, at the beginning of an event of Type 3, the belief set has a model, then, at the conclusion of the process initiated by that event, the resulting belief set has a model.  This is established below. 

Phase (iv): Let $B^**$ be the most recent belief set, i.e., the one that resulted from the above instances of Event Type 2 and/or Type 3.  In this phase, $B^**$  is searched for active formulas of the form $(\forall x)(\alpha^{(k)}(x)\to\lnot\beta^{(p)}(x))$, where $\alpha^{(k)}$ is the predicate symbol of the input formula, and, if found, Aristotelian Syllogism is applied to this and the input formula to infer $\lnot\beta^{(p)}(a)$, which is provided to the controller. This constitutes an event of Type 4.  Note that there can be any number of such $\lnot\beta^{(p)}$, each leading to an event of Type 4. Thus it remains to show that, if, at the beginning of an event of Type 4, the belief set has a model, then, at the conclusion of the process initiated by that event, the resulting belief set has a model.  This is established below. 

Now observe that $\hat B_{u+1}$ is the belief set that results from this process. Note also that $\hat L_{u+1}=L_{t+1}$. Given the following verifications regarding events of Type 2, 3 and 4, the foregoing shows that $\hat L_{u+1}$ has an interpretation that is a model of $\hat B_{u+1}$. 

Consider an event of Type 2, where a formula of the form $\alpha^{(k)}(a)$ is provided to the controller as the result of an inference rule application. This only occurs via an application of Aristotelian Syllogism. The formula $\alpha^{(k)}(a)$ is entered into the belief set. Because no new symbols need to be added, $\hat L_{u+1}=\hat L_u$.  Let $I$ be an interpretation of $\hat L_u$ that models $\hat B_u$. It is desired to show that $I$ can be transformed into a model $I'$ of the belief set $B_{u+1}$ that results from the process associated with this event. This can be established by the same considerations as above for Phase (i) of Event Type 1, except that here only Case 1 applies.   As with Event Type 1, an event of Type 2 can lead to events of Type 3 and/or Type 4.  These are considered next. 

Consider an event of Type 3, where a formula of the form $\alpha^{(p)}(a)$ is provided to the controller as the result of an inference rule application. This only occurs via an application of Aristotelian Syllogism. If the entry of this formula into the belief set is not blocked in accordance with the specificity principle, and if it does not already reside in the belief set, then the formula is entered into the belief set. Because no new symbols need to be added, $\hat L_{u+1}=\hat L_u$.  Let $I$ be an interpretation of $\hat L_u$ that models $\hat B_u$. It is desired to show that $I$ can be transformed into a model $I'$ of the belief set $B_{u+1}$ that results from the process associated with this event.  This can be established by the same considerations as above for Phase (i) of Event Type 1, Case 1, unless a scan of the current belief set discovers an occurrence of $\lnot\alpha^{(p)}(a)$, leading to an event of Type 5.  This event type is considered below.  

Consider an event of Type 4, where a formula of the form $\lnot\alpha^{(p)}(a)$ is provided to the controller as the result of an inference rule application. This only occurs via an application of Aristotelian Syllogism. If the entry of this formula into the belief set is not blocked in accordance with the specificity principle, and if it does not already reside in the belief set, then the formula is entered into the belief set. Because no new symbols need to be added, $\hat L_{u+1}=\hat L_u$.  Let $I$ be an interpretation of $\hat L_u$ that models $\hat B_u$. It is desired to show that $I$ can be transformed into a model $I'$ of the belief set $B_{u+1}$ that results from the process associated with this event.  This can be established by the same considerations as above for Phase (i) of Event Type 1, Case 1, unless a scan of the current belief set discovers an occurrence of $\alpha^{(p)}(a)$, leading to an event of Type 5.  This event type is considered below.  

Consider an event of Type 5, namely, providing $\bot$ to the controller. This only occurs via an application of Contradiction Detection. The formula $\bot$ is entered into the belief set, and this initiates an application of Dialectical Belief Revision. Let $B_t$ be the belief set that is current prior the occurrence of this event, and let $I$ be an interpretation of $L_t$ that models $B_t$. It is desired to show that $I$ can be transformed into a interpretation $I'$ of $L_t$ that is a model of the belief set $B_{t'}$ that results from the Dialectical Belief Revision process.

Let $\alpha^{(p)}(a)$ and $\lnot\alpha^{(p)}(a)$ be the premises in the application of Contradiction Detection.  The first phase of the Dialectical Belief Revision process employs from-lists to backtrack through the derivations of these formulas, identifying all extralogical axioms involved in their proofs.  Note that neither of $\alpha^{(p)}(a)$ and $\lnot\alpha^{(p)}(a)$ are extralogical axioms, but they will have been derived via Aristotelian Syllogism from some formulas of the form $\beta^{(k)}(a)$ and $(\forall x)(\beta^{(k)}(x)\to\alpha^{(p)}(x))$ (for the former) and $\gamma^{(k)}(a)$ and $(\forall x)(\gamma^{(k)}(x)\to\lnot\alpha^{(p)}(x))$ (for the latter), where $(\forall x)(\beta^{(k)}(x)\to\alpha^{(p)}(x))$ and $(\forall x)(\gamma^{(k)}(x)\to\lnot\alpha^{(p)}(x))$ are extralogical axioms.  Both of these are candidates for status change.

Suppose that the status of $(\forall x)(\beta^{(k)}(x)\to\alpha^{(p)}(x))$ is changed to {\it disbel}.  Then by forward chaining through to-lists Dialectical Belief Revision results in the status of the derived formula $\alpha^{(p)}(a)$ being changed to {\it disbel}, and then also the status of the derived $\bot$ being changed to {\it disbel}.  There are two subcases.  Subcase (i): The current instance of Event Type 5 arose as a result of this derivation, i.e., the formula $\lnot\alpha^{(p)}(a)$ was already in the belief set.  Note from the above that this belief set is $B_t$.  Thus the effect of Dialectical Belief Revision in this case is simply to disbelieve the formula that was just entered, in which case one can let $I'=I$, i.e., the interpretation $I$ is itself a model of $B_{t'}$.  Subcase (ii): The current instance of Event Type 5 arose as a result of the derivation of  $\lnot\alpha^{(p)}(a)$ from some formulas $\delta^{(k)}(a)$ and  $(\forall x)(\delta^{(k)}(x)\to\lnot\alpha^{(p)}(x))$. Then $B_t$ already contained the formula $\alpha^{(p)}(a)$, so that $I$ was such that $I(a)\in I(\alpha^{(p)})$.  Let $I'$ be obtained from $I$ by setting $I'(\alpha^{(p)})=I(\alpha^{(p)})-\{I(a)\}$, i.e., removing the object $I(a)$ from $I(\alpha^{(p)})$. Then $I'\models B_{t'}$.

Suppose that the status of $(\forall x)(\gamma^{(k)}(x)\to\lnot\alpha^{(p)}(x))$ is changed to {\it disbel}.  Then by forward chaining through to-lists Dialectical Belief Revision results in the status of the derived formula $\lnot\alpha^{(p)}(a)$ being changed to {\it disbel}, and then also the status of the derived $\bot$ being changed to {\it disbel}.  Subcase (i): The current instance of Event Type 5 arose as a result of this derivation, i.e., the formula $\alpha^{(p)}(a)$ was already in the belief set.  Note from the above that this belief set is $B_t$.  Thus the effect of Dialectical Belief Revision in this case is simply to disbelieve the formula that was just entered, in which case one can let $I'=I$, i.e., the interpretation $I$ itself is a model of $B_{t'}$.  Subcase (ii): The current instance of Event Type 5 arose as a result of the derivation of  $\alpha^{(p)}(a)$ from some formulas $\delta^{(k)}(a)$ and  $(\forall x)(\delta^{(k)}(x)\to\alpha^{(p)}(x))$. Then $B_t$ already contained the formula $\lnot\alpha^{(p)}(a)$, so that $I$ was such that $I(a)\notin I(\alpha^{(p)})$.  Let $I'$ be obtained from $I$ by setting $I'(\alpha^{(p)})=I(\alpha^{(p)})\cup\{I(a)\}$, i.e., adding the object $I(a)$ to $I(\alpha^{(p)})$. Then $I'\models B_{t'}$.

If the status of neither $(\forall x)(\beta^{(k)}(x)\to\alpha^{(p)}(x))$ nor $(\forall x)(\gamma^{(k)}(x)\to\lnot\alpha^{(p)}(x))$ is changed to {\it disbel}, then Dialectical Belief Revision proceeds to explore the derivations of $\beta^{(k)}(a)$ and $\gamma^{(k)}(a)$.  We need only consider one of these; let us choose the former.  This formula could be an extralogical axiom.  If so, then the contradiction could be removed by changing the status of $\beta^{(k)}(a)$ to {\it disbel} and then forward chaining via its to-list to change the status of  $\alpha^{(p)}(a)$ to {\it disbel} (and then on also to the input occurrence of $\bot$).  Form $I'$ from $I$ by setting $I'(\beta^{(k)})=I(\beta^{(k)})-\{I(a)\}$ and  $I'(\alpha^{(p)})=I(\alpha^{(p)})-\{I(a)\}$. Then $I'\models B_{t'}$. 

If $\beta^{(k)}(a)$ had not been input as an extralogical axiom, then it was derived by an application of Aristotelian Syllogism.  By continuing to backtrack in this manner using from-lists, the process eventually finds all extralogical axioms involved in the derivation of the current occurrence of $\bot$.  Some will be of the form $\alpha'^{(k)}(a)$, where $a$ is the individual constant in the given $\beta^{(k)}(a)$, and some will be of the form $(\forall x)(\beta'^{(k)}(x)\to\gamma'^{(k)}(x))$.  One or more of these will have their belief status changed to {\it disbel}, enough to ensure that the derivation of $\bot$ no longer applies.  Then $B_{t'}$ is obtained from $B_t$ by making these changes in these formulas' labels.  Note incidentally that the corresponding $B_{t'}$ will have been obtained by following the to-lists of all such formulas, similarly changing the labels of all their consequences up to and including the triggering occurrence of $\bot$.  Let $I'$ be obtained from $I$ by  setting $I(\alpha''^{(k)})=I(\alpha''^{(k)})-\{I(a)\}$  for every formula $\alpha''^{(k)}(a)$ whose label is changed to {\it disbel} in this second phase of the Dialectical Belief Revision process.  Then $I'\models B_{t'}$.

The forgoing shows that, whichever of the available choices are made to resolve the detected contradiction, there will be an interpretation $I'$ of $L_t$ that is a model of the resulting belief set $B_{t'}$. 

Consider an input $(\forall x)(\alpha^{(k)}(x)\to\beta^{(k)}(x))$.  This is an event of Type 6.  Let $t$ be such that $\hat B_u=B_t$. Then $\hat L_u=L_t$. Let $I$ be an interpretation of $\hat L_u$ that models $\hat B_u$. Then $I$ is an interpretation of $L_t$ that models $B_t$.  As prescribed by the event type's description, if $\alpha$ and $\beta$ are already in $L_t$, the input is rejected if the input formula would create either a redundant path or a loop.  Assuming that neither of these is the case, then the process has two phases: (i) add the input formula into the belief set, (ii) scan the belief set for any opportunities to apply Aristotelian syllogism, leading to invocations of Event Type 2.

Phase (i): Here $B_{t+1}=B_t\cup\{((\forall x)(\alpha^{(k)}(x)\to\beta^{(k)}(x)),\lambda)\}$, where $\lambda$ is the appropriate label.  The objective is to show that $I$ can be expanded to an interpretation $I'$ of $L_{t+1}$ such that $I'\models B_{t+1}$. This amounts to showing that $I$ can be expanded to an interpretation $I'$ of $L_{t+1}$ such that $I'\models (\forall x)(\alpha^{(k)}(x)\to\beta^{(k)}(x))$.  If either $\alpha^{(k)}$ or $\beta^{(k)}$ is missing from $L_t$, then $L_{t+1}$ will be the language obtained from $L_t$ by adding the ones that are missing; otherwise $L_{t+1}=L_t$. There are four cases depending on which symbols are added. In all cases, set $D_{I'}=D_I$.  Case 1: Neither is added.  Then $\alpha^{(k)}$ and $\beta^{(k)}$ are already in $L_t$.  Form $I'$ from $I$ by setting $I'(\beta^{(k)})=I(\beta^{(k)})\cup(I(\alpha^{(k)})-I(\beta^{(k)}))$, i.e., add to $I(\beta^{(k)})$ the elements in $I(\alpha^{(k)})$ that are not already in $I(\beta^{(k)})$.  Note that this cannot invalidate any expressions asserting that some of the newly added objects are not in $I(\beta^{(k)})$ since, by definition of the MIS controller, such assertions of nonmembership cannot be entered into the belief set either by the user or via derivation from other formulas in the belief set. Case 2: $\alpha^{(k)}$ is added, but $\beta^{(k)}$ is not.  Set $I'(\alpha^{(k)})=\emptyset$, i.e., introduce a new relation corresponding to $\alpha^{(k)}$ and let it be the empty subset of $I(\beta^{(k)})$.  Case 3: $\beta^{(k)}$ is added, but $\alpha^{(k)}$ is not.  Set $I'(\beta^{(k)})=I'(\alpha^{(k)})$, i.e., introduce a new relation corresponding to $\beta^{(k)}$ and make it coextensive with $I(\alpha^{(k)})$.  Case 4: Both $\alpha^{(k)}$ and $\beta^{(k)}$ are added.  Set $I'(\alpha^{(k)})=I'(\beta^{(k)})=\emptyset$.  It follows in every case that the formula $(\forall x)(\alpha^{(k)}(x)\to\beta^{(k)}(x))$ will be true in $I'$. Thus, in every case, $I'\models B_{t+1}$.

Phase (ii): The above $B_{t+1}$ is scanned for formulas of the form $\alpha^{(k)}(a)$, where $\alpha^{(k)}$ is the predicate symbol of the input formula, and, if found, Aristotelian Syllogism is applied, leading to an event of Type 2. That this results in a belief set with a model was established in the foregoing. 

Let $\hat L_{u+1}$ and$\hat B_{u+1}$ be the language and belief set that results from the above process. It follows that there is an interpretation $I'$ of $\hat L_{u+1}$ such that $I'\models B_{u+1}$. 

Consider an input $(\forall x)(\alpha^{(k)}(x)\to\beta^{(p)}(x))$.  This is an event of Type 7, which may lead to some occurrences of Event Type 3.  The latter is discussed in the foregoing.

Consider an input $(\forall x)(\alpha^{(k)}(x)\to\lnot\beta^{(p)}(x))$.   This is an event of Type 8, which may lead to some occurrences of Event Type 4.  The latter is discussed in the foregoing.

This concludes the induction step and completes the proof by induction on $u$.  $\square$

\end{document}